\documentclass[preprint,12pt,authoryear]{elsarticle}
\usepackage{amssymb}
\usepackage{hyperref} 
\usepackage{lineno}
\usepackage{soul}
\usepackage{setspace}
\usepackage{subfig}
\usepackage{bm}
\usepackage{amssymb}
\usepackage{amsmath}
\usepackage{upgreek}
\usepackage{bigints}
\usepackage{algorithm2e}
\usepackage{algpseudocode}
\usepackage{hyphenat}
\usepackage{xurl}
\RestyleAlgo{ruled}
\usepackage{float}
\usepackage{xcolor}
\usepackage{multirow}
\usepackage[left=1.7cm, right=1.7cm, top=3.0cm, bottom=3.0cm]{geometry}

\journal{Computer Methods in Applied Mechanics and Engineering}
\begin{document}
\begin{frontmatter}

\title{Multimodal Auto-regressive Transformer Surrogate for Modeling Variable Operations and Quantifying Uncertainty in Geological Carbon Storage}

\author[inst1]{Yifu Han}
\affiliation[inst1]{organization={Department of Energy Science and Engineering}, 
            addressline={Stanford University}, 
            city={Stanford},
            state={CA},
            postcode={94305}, 
            country={USA}}

\author[inst1]{Louis J.~Durlofsky}

\begin{abstract}
The use of variable well perforation and injection strategies can improve the efficiency of geological carbon storage operations. We develop a new multimodal auto-regressive transformer surrogate to model these operations under geological uncertainty. A modified SEAM CO$_2$ geomodel, which involves a faulted system with three stacked aquifers, is considered. The two injection wells are perforated in stages, from bottom to top, with the stage durations and individual well injection rates treated as control variables. The surrogate model processes three input modalities -- the 3D geomodel, scalar parameters characterizing relative permeability functions, and control variables -- through separate encoders. These are fused via self-attention in a transformer encoder, and a temporal decoder generates predictions auto-regressively through encoder-decoder cross-attention. The surrogate is trained, using 4000 GEOS flow simulations, to predict saturation and pressure at monitoring locations, total injected and mobile CO$_2$ mass, and saturation footprints. For a new test set, involving randomly sampled geomodels and control variables, the surrogate achieves a median saturation MAE of 0.028 and median relative errors of 0.2--5\% for the other quantities of interest. Importantly, it captures the switch from rate to bottom-hole-pressure control. The surrogate model is used within a hierarchical Markov chain Monte Carlo data assimilation procedure for a synthetic true model under three operational strategies. Substantial uncertainty reduction is achieved for key metaparameters, particularly the fault permeabilities. Posterior predictions for saturation footprints and total injected and mobile CO$_2$ mass are also shown to be generally consistent with true model results.

\end{abstract}


\end{frontmatter}

\section{Introduction}
\label{Introduction}
Geological carbon storage (GCS) entails the injection of large volumes of supercritical CO$_2$ into porous subsurface formations. The performance of these operations depends on the geological properties of the storage aquifer, which are often highly uncertain, as well as on operational settings. These settings could include the injection rates for each well and the portion of the formation over which the well is perforated, i.e., open to flow. By adjusting these operational variables through time, improved performance, in terms of the total amount of CO$_2$ injected, the size of the CO$_2$ footprint, and the fraction of stored CO$_2$ that is immobile (and thus unable to leak through overlying caprock), can be achieved. To enable these types of operational decisions, calibrated geomodels, meaning models that provide flow predictions in agreement with observations, are required. The associated data assimilation problem is computationally expensive, and for this reason many surrogate models have been developed. However, in most existing data assimilation setups, the operational variables (rates, perforation intervals) are assumed to be fixed, so those surrogates are not applicable for the flexible-operation situations considered in this work.

In this study, we develop a new surrogate model to predict key flow responses, in both the injection and post-injection periods, under variable perforation and injection strategies and geological uncertainty. A modified version of an existing community model, the Society of Exploration Geophysicists Advanced Modeling (SEAM) CO$_2$ geomodel, is considered. The modified SEAM geomodel, described in \citet{han2026recurrent}, represents a complicated Gulf of Mexico faulted aquifer system containing three stacked reservoirs and two large faults. In the operational strategies treated here, the injection wells in the target aquifer are perforated in three stages at a sequence of times. We develop a multimodal auto-regressive transformer surrogate model to predict key quantities of interest for this setup given a new geomodel along with a set of perforation intervals and injection rates. This surrogate is trained using results from 4000 high-fidelity GEOS~\citep{Settgast2024} flow simulations. The surrogate model is used within a Markov chain Monte Carlo (MCMC) data assimilation procedure to reduce uncertainty in geological parameters and in system-level quantities of interest.

The impact of the perforation strategy on flow performance has been widely recognized in hydrocarbon production settings. \citet{stiles1990investigating}, for example, adjusted the perforation strategy to balance the advance of water fronts and delay water breakthrough at production wells. \citet{zhang2020alternated} showed that alternating injection between separate layers over time acted to improve recovery efficiency. \citet{azamipour2023completion} optimized perforated intervals to control the injection profile and delay breakthrough. Perforation and injection strategies have also been studied for GCS applications. \citet{kumar2008optimizing} showed that, for a given injection rate, the fraction of immobile CO$_2$ can be increased by adjusting the injection interval. More generally, \citet{kumar2009semi} showed that highly nonuniform injection profiles can result when vertical wells are perforated over the full formation thickness. This nonuniform injection acts to reduce the volume of brine contacted by CO$_2$, thereby lowering the amount of dissolution and residual trapping and leaving a larger fraction of mobile CO$_2$. \citet{luo2013numerical} further investigated the impact of injection and production well perforation strategies on CO$_2$ storage with enhanced gas recovery. They showed that optimal perforation depends strongly on formation properties. \citet{oldenburg2021radial} introduced the concept of radial storage efficiency and used it to demonstrate that particular completion strategies can limit flow into high permeability layers and increase flow into low permeability layers. This more uniform injection improves sweep and trapping, and acts to reduce the CO$_2$ footprint and mobile CO$_2$ fraction. 

Many deep-learning-based surrogate models have been developed for GCS and related subsurface problems. Here we highlight a few of the models that are most relevant to our problem -- this discussion is not intended to be comprehensive. \citet{mo2019deep} introduced a convolutional encoder-decoder network to predict pressure and saturation for 2D GCS problems. \citet{wang2021efficient} developed a theory-guided neural network surrogate, which incorporates physical constraints during training, for Monte Carlo-based uncertainty quantification of subsurface flow problems. \citet{tang2022deep} extended the recurrent R-U-Net surrogate model~\citep{tang2021deep} to treat coupled flow and geomechanics. Their procedure provided 3D pressure and saturation fields along with vertical surface displacements. \citet{yan2022robust} developed a Fourier neural operator (FNO) surrogate to predict saturation and pressure during both injection and post-injection periods. \citet{wen2023real} developed a nested FNO to treat 3D problems involving heterogeneous geomodels and variable well configurations. \citet{lee2024efficient} proposed a nested Fourier-DeepONet that achieved improved extrapolation capability relative to the nested FNO. \citet{tang2025graph} developed a graph network surrogate model to predict 3D pressure and saturation fields for varying horizontal well configurations. \citet{ju2026adaptive} developed an adaptive physics transformer with fused global-local attention for surrogate modeling in subsurface flow problems. \citet{han2026recurrent} introduced a recurrent transformer U-Net surrogate model with an attention mechanism for 3D faulted aquifer systems (specifically the modified SEAM geomodel) in which qualitatively different fault leakage scenarios can occur. 

Most previous surrogate models used to perform data assimilation in subsurface flow settings were developed for cases with fixed operational settings. By this we mean the geomodel was the primary input, and the outputs were flow quantities for the new geomodel under the operational strategy used in training. Practical problems, however, may require the surrogate to incorporate additional types of inputs involving well control. Problems of this type can be treated within a multimodal learning framework. Multimodal learning constructs models that combine and relate multiple data modalities, where representation, alignment, fusion, and co-learning are the key issues~\citep{baltrusaitis2019multimodal}. Transformer-based architectures~\citep{vaswani2017attention} are often used for this purpose, as attention mechanisms can capture long-range dependencies and can be extended to treat cross-modal interactions~\citep{xu2023multimodal}. Here we mention a few of the many such models developed in the computer vision and natural language processing communities. \citet{radford2021clip} introduced a model that aligns image and text representations in a shared embedding space. \citet{lu2019vilbert} and \citet{tan2019lxmert} developed transformer-based models that fuse visual and textual inputs through cross-attention. Transformer-based architectures have also been used as neural operators for the solution of partial differential equations~\citep{shih2025transformers}. 

These ideas are now being applied for GCS problems. \citet{seabra2024ai} developed a transformer U-Net surrogate model for 2D problems in which the injection rate is incorporated at the latent level through cross-attention. \citet{feng2025hybrid} developed a CNN-transformer model for idealized 3D problems in which the injection rate sequence is fused with the encoded geomodel in the latent space. \citet{feng2026coswinnet} developed a conditional Swin transformer surrogate model with a U-Net structure for 2D problems, with the operational parameters incorporated at each stage through feature-wise linear modulation. The models discussed here, however, did not treat additional input modalities, such as the scalar parameters characterizing relative permeability curves. In addition, previous studies have not considered complicated realistic systems, e.g., 3D faulted geomodels with variable perforation and injection under well bottom-hole pressure (BHP) limits, which are considered in this work. 

Our goals in this study are to develop a multimodal surrogate that treats different input modalities and to apply it for modeling and data assimilation for the modified SEAM geomodel. To motivate the overall setup, we first demonstrate, using high-fidelity GEOS simulations, that the variable perforation and injection strategy is effective in improving the performance of CO$_2$ storage operations. We then introduce the multimodal surrogate. This model accepts as input the operational variables, the geomodel, and relative permeability parameters. Embedding and fusion treatments are introduced to accommodate these different input modalities, and modality-specific encoders map the geomodel, relative permeability parameters, and operational variables into a shared latent space. The resulting tokens are fused and processed by a transformer encoder via multihead self-attention, and a temporal decoder then generates the time-dependent predictions auto-regressively through encoder–decoder cross-attention. Comparisons between surrogate model predictions and high-fidelity simulation results are presented for key quantities including total CO$_2$ injected, the mass of mobile CO$_2$, and the saturation footprint associated with each injector. We then incorporate the surrogate model into a hierarchical MCMC-based data assimilation procedure to evaluate the degree of uncertainty reduction achieved for a particular ``true'' model under different perforation and injection specifications. The geological uncertainty considered here is hierarchical because both the high-level geological metaparameters and the cell-by-cell realizations are uncertain. To our knowledge, this is the first surrogate-based treatment of variable perforation and injection strategies, combined with uncertain hierarchical geomodels and uncertain relative permeability functions, in the context of geological carbon storage.

This paper proceeds as follows. In Section~\ref{Geomodel}, we describe the modified SEAM CO$_2$ geomodel and the uncertain parameters characterizing the geological scenarios and relative permeability curves. The potential advantages of the variable perforation and injection strategy are illustrated. The multimodal auto-regressive transformer surrogate model is presented in Section~\ref{Surrogate Model}. Surrogate model accuracy is assessed, and its predictions are compared with high-fidelity simulation results, in Section~\ref{Surrogate Evaluation}. The hierarchical MCMC data assimilation workflow and results, for three different perforation and injection strategies, are presented in Section~\ref{Data Assimilation}. We conclude in Section~\ref{Conclusions} with a summary and suggestions for future work.

\section{Geomodel and Simulation Setup}
\label{Geomodel}
In this section we present the modified SEAM CO$_2$ geomodel and the simulation setup used in this study. The structural model (fault locations and aquifer geometries) is identical to that used in \citet{han2026recurrent}, though the parameters describing the aquifer properties and the flow setup under the variable perforation and injection strategy are different.

\subsection{Geomodel and metaparameters}
\label{sec:geomodel}

\citet{fehler2024seam} and \citet{yoon2024assessing} developed the geomodels used for the SEAM CO$_2$ project. These models were constructed based on earlier SEAM Gulf of Mexico tertiary salt models~\citep{stefani2010seam, oristaglio2016seam}. Geologically, these models represent a turbidite system consisting of a shaly sandstone aquifer, an overlying caprock, two shallower aquifers, and intersecting faults. 

\begin{figure}[!ht]
\centering   
{\includegraphics[width = 175mm]{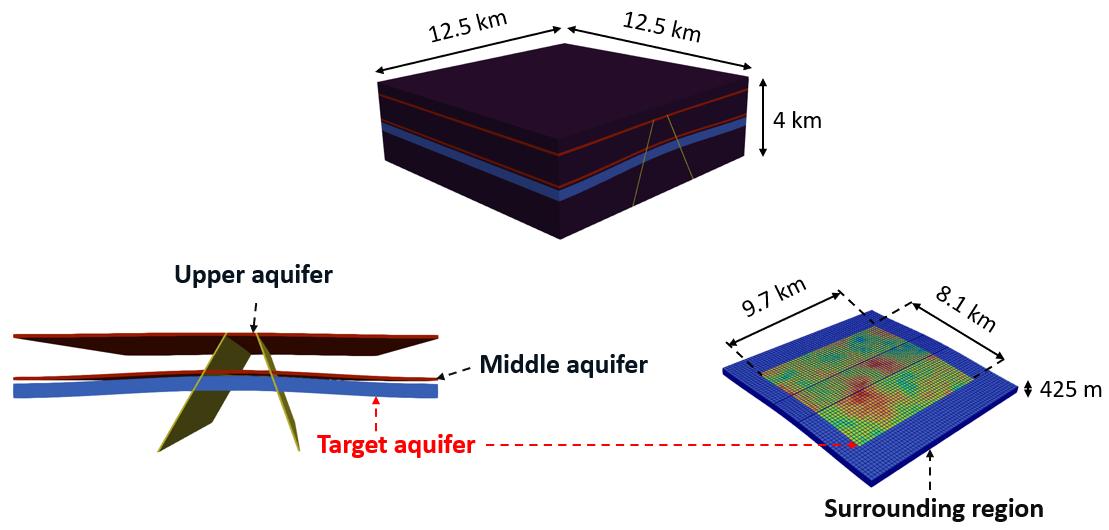}}
\caption{Overall domain of the modified SEAM CO$_2$ geomodel (top), cross-sectional view showing the three aquifers and two faults (lower left), and areal view of the target aquifer and surrounding region (lower right).}
\label{SEAM_Model}
\end{figure}

The overall domain is shown in Fig.~\ref{SEAM_Model}. The geomodel consists of three stacked aquifers, separated by low-permeability caprock and connected by two extensive faults. Low-permeability overburden and underburden rock lies above and below the three permeable aquifers. Supercritical CO$_2$ is injected into the lowermost aquifer, which is referred to as the target aquifer. The target aquifer spans 8.1~km $\times$ 9.7~km $\times$ 425~m and is discretized into 48 $\times$ 48 $\times$ 25 cells, corresponding to an average cell size of 168~m $\times$ 202~m $\times$ 17~m. The aquifer is folded and the depth to its top varies spatially, averaging about 1854~m. The middle aquifer (12.5~km $\times$ 12.5~km $\times$ 102.5~m, average depth 1717~m) is discretized into 69 $\times$ 62 $\times$ 6 cells. The upper aquifer (average depth 563~m) is of similar physical dimensions and grid resolution. The full model is represented on a grid containing 69 $\times$ 62 $\times$ 78~cells (total of 330,000~cells). Two north-south-striking faults, separated by roughly 2~km in the target aquifer, intersect all three aquifers. Each fault is 60~m wide and is represented by two grid cells. The permeabilities of the faults are uncertain -- CO$_2$ can migrate into the two shallower aquifers if the faults are permeable. For more details on the structure and gridding of the modified SEAM CO$_2$ geomodel, please see~\citet{han2026recurrent}.

The original SEAM CO$_2$ geomodel~\citep{fehler2024seam, yoon2024assessing} is deterministic. In this study, consistent with \citet{han2026recurrent}, the target aquifer is characterized by multi-Gaussian log-permeability and porosity fields with uncertain scenario parameters (also referred to as metaparameters or hyperparameters). Once the metaparameters are sampled or otherwise specified, cell-by-cell geomodel properties can be constructed using geological modeling software or principal component analysis (PCA). In addition to parameters characterizing the geostatistical property distributions, the metaparameters also define the permeabilities of the faults and relative permeability parameters. The full set of parameters, denoted $\boldsymbol{\uptheta}_{\mathrm{meta}} \in \mathbb{R}^{10}$ and collectively
referred to as metaparameters, are 
\begin{equation} \label{multi-modal_faulted_metaparameters}
\boldsymbol{\uptheta}_{\mathrm{meta}} = [\mu_{\log k}, \sigma_{\log k}, a_r, d, e, k_{f1}, k_{f2}, S_{wi}, S_{gr}, k_{rg}^0].
\end{equation}
\noindent Here $\mu_{\log k}$ and $\sigma_{\log k}$ are the mean and standard deviation of the multi-Gaussian log-permeability field in the target aquifer, $a_r$ is permeability anisotropy ratio, and $d$ and $e$ are coefficients relating porosity to log-permeability in the target aquifer (via $\phi = d \cdot \log k_x + e$). Here $k_x$ and $k_y$ denote the $x$ and $y$-components of permeability. We specify $k_x=k_y$ and $k_z=a_r k_x$. The parameters $k_{f1}$ and $k_{f2}$ represent the permeabilities for fault~1 and fault~2. Each fault is assigned a single permeability value in this study, which differs from our previous setup where fault permeability could vary with depth. The parameters characterizing the CO$_2$-brine relative permeability functions with hysteresis, \emph{S$_{wi}$}, \emph{S$_{gr}$}, and \emph{k$_{rg}^0$} , are the irreducible water saturation, the residual CO$_2$ saturation, and the end-point CO$_2$ relative permeability at the irreducible water saturation, respectively. These parameters apply to the full domain.

The procedures used to construct heterogeneous geomodel realizations of the overall domain closely follow those described in \citet{han2026recurrent}. The prior ranges for the metaparameters (note that uncertain parameters are taken to be uniformly distributed), along with the values specified for many of the fixed parameters, are provided in Tables~\ref{multi-modal_rock_physics} and~\ref{relative_permeability}. The prior ranges are generally similar to those in the previous study, though there are some differences. Specifically, here we fix the permeabilities in the middle and upper aquifers at 300~md rather than treat them as uncertain, and we consider smaller prior ranges for the fault permeabilities. In addition, the vertical correlation length is smaller here, and relative permeability parameters are now treated as uncertain. Once the metaparameters are specified, random realizations of the target aquifer are generated using the PCA representation. The PCA basis matrix is constructed from realizations generated using sequential Gaussian simulation within the geomodeling software SGeMS~\citep{remy2009applied}. This treatment is consistent with that applied in previous studies~\citep{han2023surrogate, han2025accelerated, han2026recurrent}.

\begin{figure}[!ht] 
\centering
\subfloat[Relative permeability curves]{\includegraphics[width = 80mm]{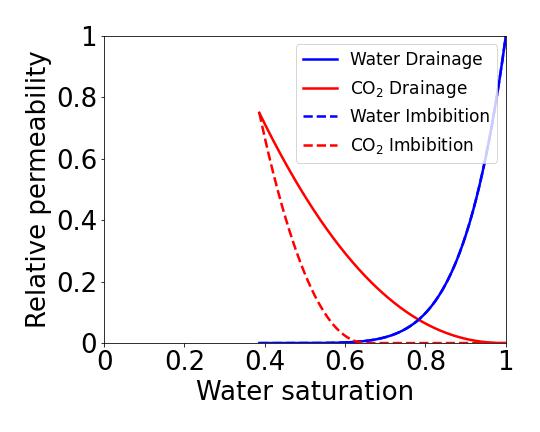}}
\hspace{5mm}
\subfloat[Capillary pressure curve]{\includegraphics[width = 80mm]{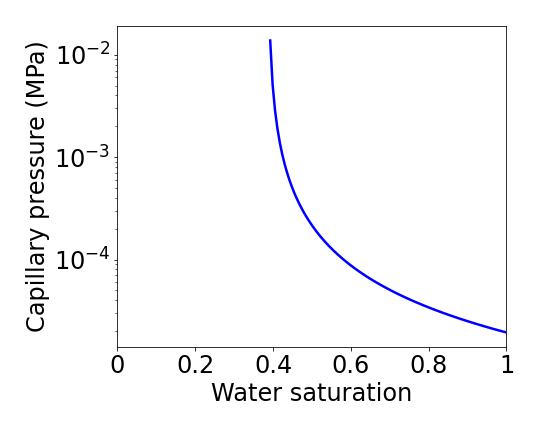}}
\caption{CO$_2$-brine relative permeability with hysteresis effects and capillary pressure curves for geomodel realization~1. Capillary pressure curve in (b) is for a cell with $\phi=0.19$ and $k=330$~md.}
\label{fig:multi_modal_relper_pc}
\end{figure}

The metaparameters characterizing three randomly generated geomodel realizations and the associated relative permeability curves are provided in Table~\ref{multi-modal_metaparameters}. Realization~1 will be considered in our detailed assessments later in this paper. The relative permeability curves with hysteresis effects for realization~1 are displayed in Fig.~\ref{fig:multi_modal_relper_pc}a. The capillary pressure curve for a grid cell with $\emph{S$_{wi}$} = 0.39$ (as in realization~1), $\phi=0.19$, and $k=330$~md, which correspond to the arithmetic mean of porosity and geometric mean of permeability in the target aquifer for this realization, is shown in Fig.~\ref{fig:multi_modal_relper_pc}b. 

\begin{table}[!ht]
\begin{center}
\footnotesize
\caption{Parameters used in the flow simulations}
\label{multi-modal_rock_physics}
\renewcommand{\arraystretch}{1.3}
\begin{tabular}{ c @{\hskip 3em} c } 
\hline
\textbf{Overburden/Underburden/Caprock} & \textbf{Value} \\ 
\hline
Permeability & 0.0001 md \\ 
Porosity & 0.01 \\ 
Permeability anisotropy ratio & 0.1 \\
\hline
\textbf{Upper/Middle aquifer} & \textbf{Value or range} \\ 
\hline
Permeability ($k_u$, $k_m$) & 300 md \\ 
Porosity & 0.3 \\ 
Permeability anisotropy ratio  & 0.1 \\
\hline
\textbf{Target aquifer} & \textbf{Value or range} \\ 
\hline
Correlation length in $x$ ($l_x$) & 10~cells (1.7~km) \\
Correlation length in $y$ ($l_y$) & 10~cells (2~km) \\
Correlation length in $z$ ($l_z$) & 1~cells (17~m) \\
Mean of log-permeability ($\mu_{\log k}$)  & $\mu_{\log k} \sim$ $U$(4, 6) $\Leftrightarrow$ (54.6, 403.4)~md \\ 
Standard deviation of log-permeability ($\sigma_{\log k}$) & $\sigma_{\log k} \sim$ $U$(1.0, 1.5) \\ 
Parameter $d$ in $\log k$--$\phi$ correlation & $d \sim U(0.02, 0.04)$ \\ 
Parameter $e$ in $\log k$--$\phi$ correlation & $e \sim U(0.06, 0.08)$ \\ 
Permeability anisotropy ratio ($a_r=k_z/k_x$)  & $\log_{10}(a_r)$ $\sim$ $U$(-1.3, -0.7) \\
\hline
\textbf{Surrounding region} & \textbf{Value} \\ 
\hline 
Permeability & 10 md \\ 
Porosity & 0.1 \\ 
Permeability anisotropy ratio & 0.1 \\
\hline
\textbf{Faults} & \textbf{Value or range} \\ 
\hline
Fault~1 permeability ($k_{f1}$) & $\log_{10}(k_{f1}) \sim$ $U$(-2, 2)\\ 
Fault~2 permeability ($k_{f2}$) & $\log_{10}(k_{f2}) \sim$ $U$(-2, 2) \\ 
Porosity & 0.2 \\ 
Permeability anisotropy ratio  & 1 \\
\hline
\end{tabular}
\end{center}
\end{table}

\begin{table}[!ht]
\begin{center}
\footnotesize
\caption{Parameters for relative permeability and capillary pressure curves}
\label{relative_permeability}
\renewcommand{\arraystretch}{1.3}
\begin{tabular}{ c @{\hskip 3em} c }  
\hline
\textbf{Parameter} & \textbf{Value or range} \\ 
\hline
Irreducible water saturation ($\emph{S$_{wi}$}$) & $U$(0.2, 0.4) \\
Residual CO$_2$ saturation ($\emph{S$_{gr}$}$) & $U$(0.2, 0.4) \\
Water exponent for Corey model ($\emph{n$_{w}$}$) & 5.9 \\
CO$_2$ exponent for Corey model ($\emph{n$_{g}$}$) & 2.2 \\
Relative permeability of CO$_2$ at $\emph{S$_{wi}$}$ ($\emph{k$_{rg}^0$}$) & $U$(0.4, 1) \\
Capillary pressure exponent ($\lambda$) & 0.7 \\ 
\hline
\end{tabular}
\end{center}
\end{table}

\begin{table}[!ht]
\begin{center}
\footnotesize
\caption{Metaparameters characterizing geomodel realizations~1,~2, and~3}
\label{multi-modal_metaparameters}
\renewcommand{\arraystretch}{1.3}
\begin{tabular}{ c @{\hskip 3em} c @{\hskip 1em} c @{\hskip 1em} c @{\hskip 1em}}
\hline
\textbf{Metaparameter} & \textbf{Realization~1} & \textbf{Realization~2} & \textbf{Realization~3} \\ 
\hline
Mean of log-permeability ($\mu_{\log k}$)  &  5.8 $\Leftrightarrow$ 330.3~md &  5.4 $\Leftrightarrow$ 221.4~md & 4.5 $\Leftrightarrow$ 90~md\\ 
Standard deviation of log-permeability ($\sigma_{\log k}$) & 1.2 & 1.2 & 1.4\\ 
Parameter $d$ in $\log k$--$\phi$ correlation & 0.02 & 0.03 & 0.03\\ 
Parameter $e$ in $\log k$--$\phi$ correlation & 0.07 & 0.07 & 0.07\\ 
Permeability anisotropy ratio ($a_r$) & 0.08 & 0.2 & 0.05\\
Fault~1 permeability ($k_{f1}$) & 0.02~md & 29.2~md & 0.06~md\\ 
Fault~2 permeability ($k_{f2}$) & 88.7~md & 74.2~md & 0.6~md\\ 
Irreducible water saturation ($\emph{S$_{wi}$}$) & 0.39 & 0.27 & 0.36\\
Residual CO$_2$ saturation ($\emph{S$_{gr}$}$) & 0.34 & 0.25 & 0.23\\
Relative permeability of CO$_2$ at $\emph{S$_{wi}$}$ ($k_{rg}^0$) & 0.75 & 0.87 & 0.88\\
\hline
\end{tabular}
\end{center}
\end{table}

\subsection{Simulation with variable perforation and injection}
\label{sec:simulation}

Supercritical CO$_2$ is injected into the target aquifer through two vertical wells. The combined target injection rate for the two wells is 2~Mt~CO$_2$ per year for a 50-year period, corresponding to a total cumulative injection of 100~Mt. Maximum bottom-hole pressure (BHP) values are specified for both wells, and if these values are reached at any time in the simulation, the system switches from rate control to BHP control. This results in less total CO$_2$ injection, i.e., the 100~Mt total will not be reached. The maximum BHP depends on rock properties and the in-situ stress. It is specified such that the injection pressure stays sufficiently below the rock fracture pressure. In this study, we specify the maximum BHP to be 50\% above the initial pressure at the top of the perforated interval for each injector. This type of criterion is widely used in practice~\citep{estublier2017simulation}. 

\begin{figure}[!ht]
\centering   
{\includegraphics[width = 90mm]{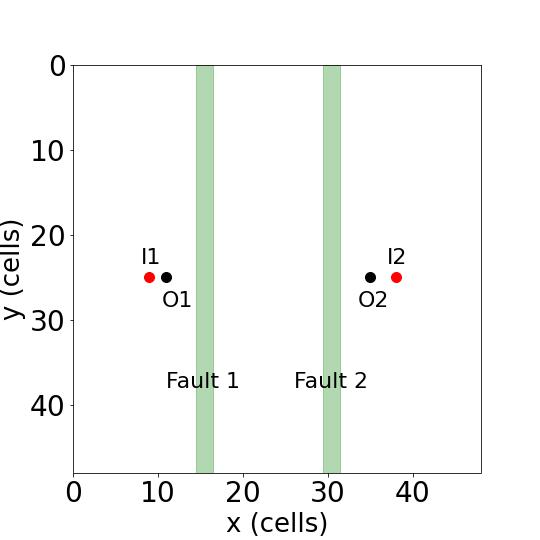}}
\caption{Areal view of the target aquifer, showing the locations of the two faults, injection wells I1 and I2, and observation wells O1 and O2.}
\label{multi-modal_well_location}
\end{figure}

The locations of the two vertical injection wells (denoted I1 and I2), along with the two faults and two vertical observation wells (denoted O1 and O2), are shown in Fig.~\ref{multi-modal_well_location}. Flow-only simulations (geomechanical effects are not considered) are performed using GEOS~\citep{bui2021multigrid, Settgast2024}. We simulate for a total time frame of 100~years, which includes the 50-year injection period and a 50-year post-injection equilibration period. A typical GEOS run for this setup requires about 13~minutes using 32 AMD EPYC-7543 CPU cores in parallel.

\begin{figure}[!ht]
\centering  
\includegraphics[width=16.5cm]{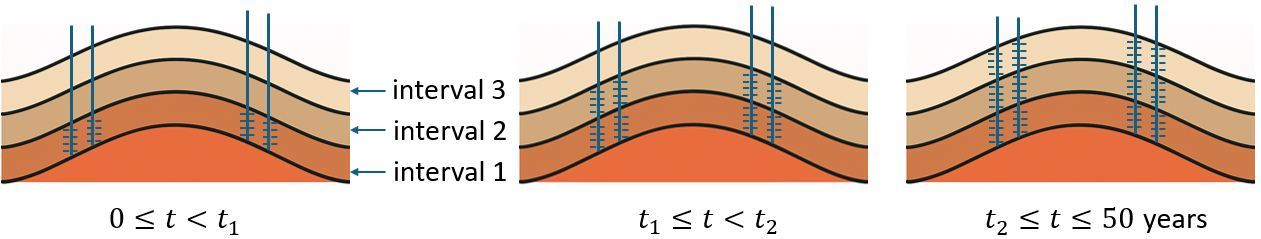}
\caption{Schematic of the bottom-up sequential perforation strategy. The target aquifer is perforated in three stages, applied at a sequence of times. These times, and the corresponding well injection rates, are the control variables considered in this study.}
\label{Peforation}
\end{figure}

As explained earlier, we consider a sequential perforation strategy, in which perforations are introduced in three equal-length intervals, starting at the bottom of the aquifer, at a sequence of times. In the first stage, perforation and injection are in the bottom third of the aquifer, while in the second stage perforations are added in the middle third (so injection is now over the lower two-thirds of the aquifer). The top third of the target aquifer is perforated in the final stage, at which point the full well is open to flow. This is illustrated in Fig.~\ref{Peforation} for a target aquifer with an anticlinal structure (as considered in this study). The perforation time for each stage is a control variable taken to be the same for both wells. The target injection rates, which are also control variables, differ in general between the two wells. The two well rates sum to 2~Mt/year, though the actual rates will be less if the BHP constraint is active. Note that, in our initial implementation~\citep{han2026surrogate}, slightly different control variables were used to define the perforation and injection strategy.

\begin{figure}[!ht]
\centering
\subfloat[Strategy~1]{\includegraphics[width=59mm]{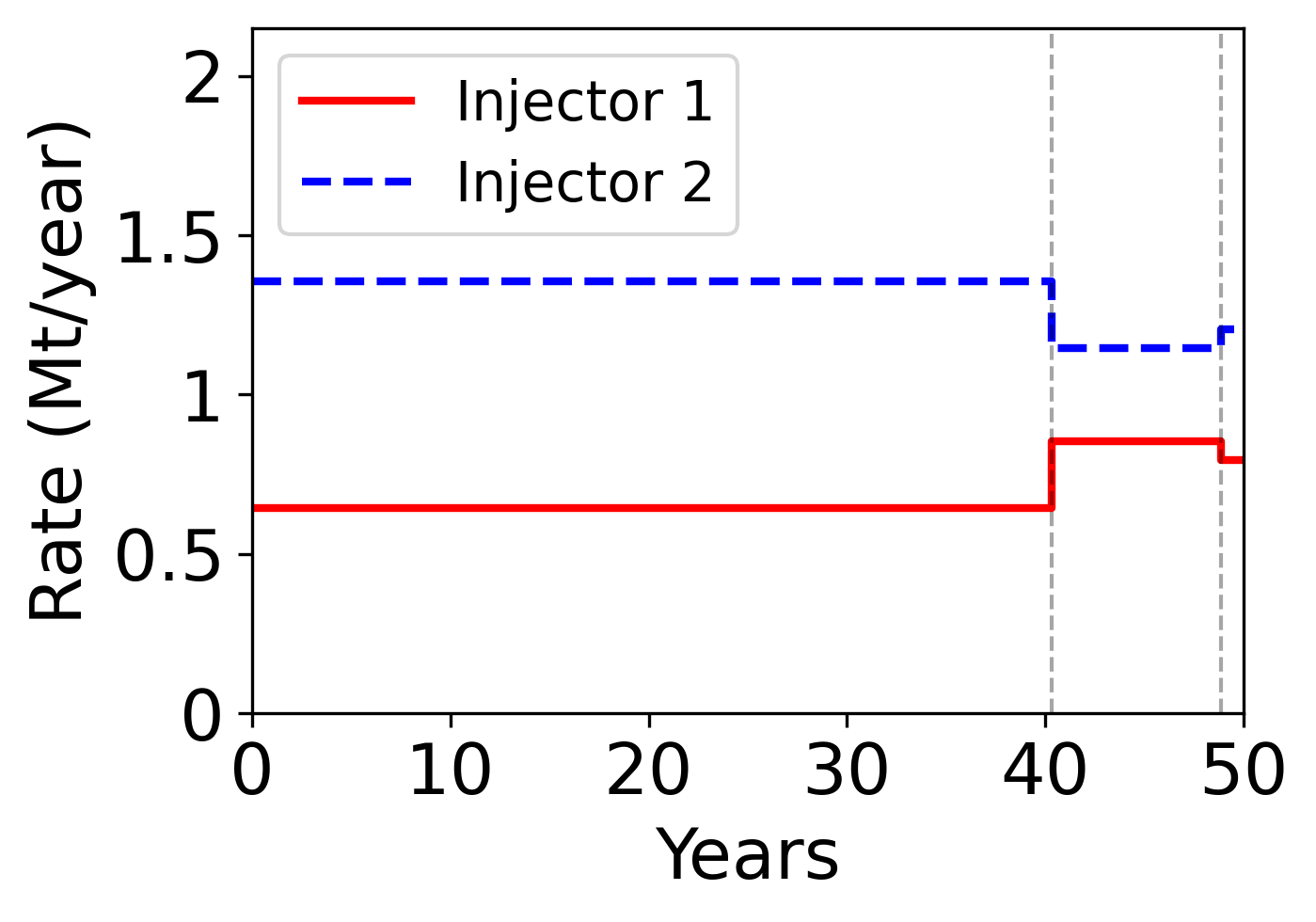}}
\hspace{1mm}
\subfloat[Strategy~2]{\includegraphics[width=59mm]{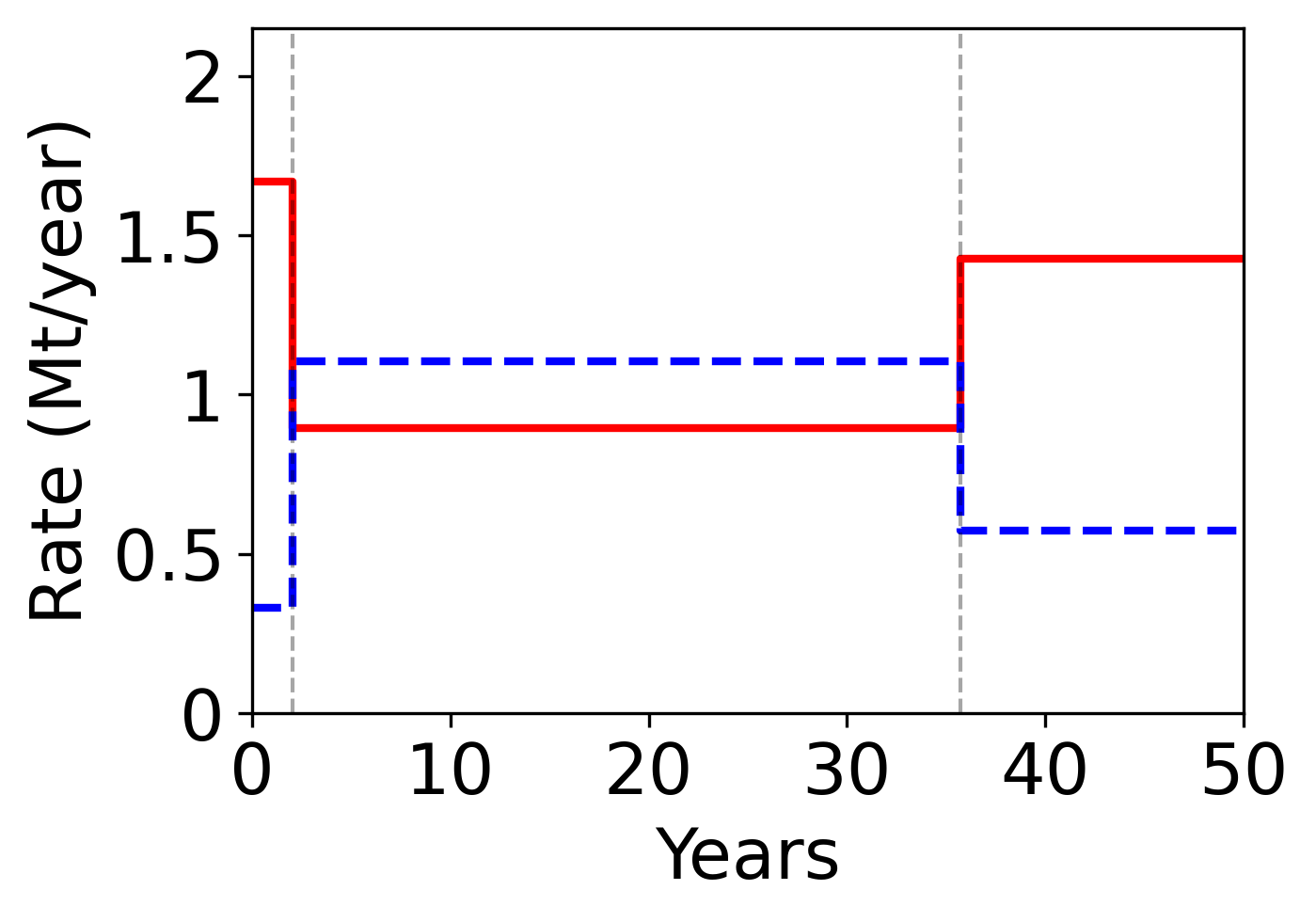}}
\hspace{1mm}
\subfloat[Strategy~3]{\includegraphics[width=59mm]{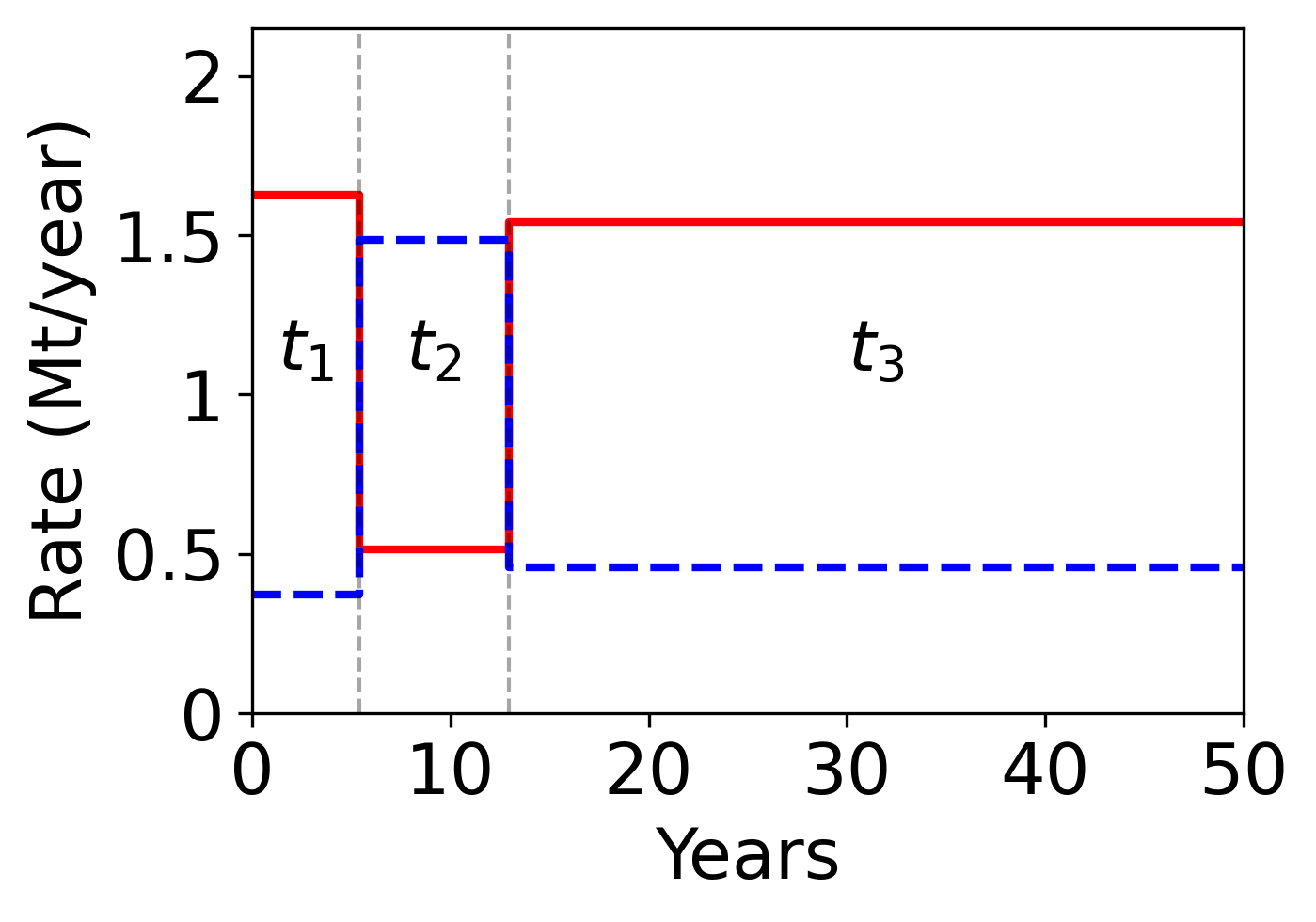}}
\\[1ex]
\caption{Three randomly generated perforation and injection strategies for the two injection wells. These operation strategies are considered for data assimilation in Section~\ref{Data Assimilation} for geomodel realization~1. Legend in (a) applies to all subplots.}
\label{variable_injection}
\end{figure}

The full set of control variables, denoted $\mathbf{t}_{\mathrm{control}} \in \mathbb{R}^{5}$, is
\begin{equation} \label{perforation_time}
{\mathbf t}_{\mathrm{control}} = [t_1, t_2, q^1_1, q^1_2, q^1_3],
\end{equation}
\noindent where $t_1$ is the injection duration for the first stage (when only interval~1 is open to flow), and $t_2$ is the injection duration for the second stage (when intervals~1 and~2 are open to flow). The third stage is also defined by these parameters since $t_1+t_2+t_3=50$~years. The variables $q^1_1$, $q^1_2$, and $q^1_3$ denote the target injection rates for injection well~I1 during the three perforation stages. These are constrained to lie between a minimum of 0.2~Mt/year and a maximum of 1.8~Mt/year. The target injection rates for injection well~I2, in units of Mt/year and denoted $q^2_1$, $q^2_2$, and $q^2_3$, are given by $q^2_1=2-q^1_1$, $q^2_2=2-q^1_2$, and $q^2_3=2-q^1_3$. We reiterate that the actual injection rates for each injector will be less if the well switches to BHP control, and this represents an important complication that the surrogate model needs to capture. Three randomly generated variable perforation and injection strategies for the two injection wells are shown in Fig.~\ref{variable_injection}.

\begin{figure}[!ht]
\centering   
\subfloat[Case 1]{\includegraphics[width=54mm]{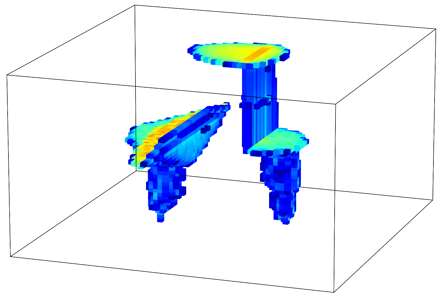}}
\hspace{2mm}
\subfloat[Case 2]{\includegraphics[width=54mm]{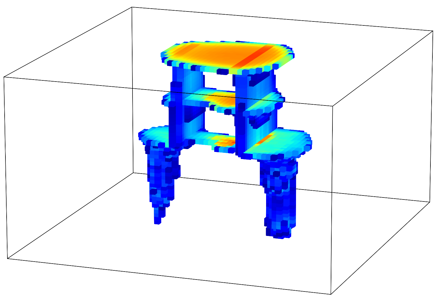}}
\hspace{2mm}
\subfloat[Case 3]{\includegraphics[width=54mm]{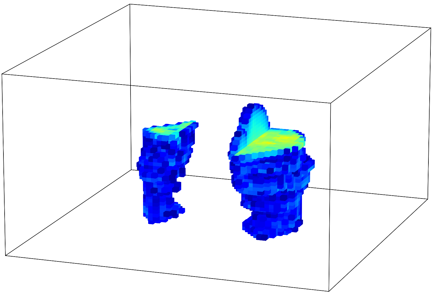}}
\hspace{1mm}
\includegraphics[width=8mm]{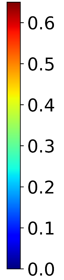}\\[1ex]
\caption{CO$_2$ saturation in the full domain at 100~years from GEOS flow simulation for  three random geomodel realizations (characterized by metaparameters given in Table~\ref{multi-modal_metaparameters}), each with a different injection strategy.}
\label{S:saturation_100_years}
\end{figure}

To illustrate the range of responses that can be observed with our setup, we now present detailed saturation fields from GEOS flow simulations for three cases. These correspond to  geomodel realizations~1, 2, and 3 in Table~\ref{multi-modal_metaparameters}, each with a different randomly generated perforation strategy. The 3D CO$_2$ saturation fields in the full domain at the end of the post-injection equilibration period (100~years) for these three cases are shown in Fig.~\ref{S:saturation_100_years}. These cases correspond to a range of flow and fault leakage behaviors. In case~1 (geomodel realization~1 with strategy~1 in Fig.~\ref{variable_injection}), fault~2 (on the right) is relatively permeable ($k_{f2}=88.7$~mD), resulting in leakage into the upper aquifer. Leakage does not occur through fault~1 because of its low permeability ($k_{f1}=0.02$~mD). For case~2, both faults are reasonably permeable ($k_{f1}=29.2$~mD, $k_{f2}=74.2$~mD), and leakage into both the middle and upper aquifers occurs. For case~3, both faults are of low permeability ($k_{f1} = 0.06$~md and $k_{f2} = 0.6$~md), so no leakage is observed into the middle and upper aquifers. As a result, the CO$_2$ plumes in the target aquifer in this case are larger than in cases~1 and 2.

Key goals with the variable perforation and injection strategy could be to (1)~inject the full 100~Mt of CO$_2$ over the 50-year injection time frame, and (2)~minimize the fraction of this CO$_2$ that is mobile and thus able to migrate or leak. These metrics are computed from GEOS simulations for 100 random perforation and injection strategies for geomodel realization~1. In generating these random strategies, we sample the stage durations from a Dirichlet distribution with a minimum duration of 0.25~years. The injection rate for well~1 at stage~1 is sampled as $q^1_1 \sim U(0.2,1.8)$~Mt, and similarly for $q^1_2$ and $q^1_3$. 

\begin{figure}[!ht]
\centering
\subfloat[Geomodel realization~1 at 50~years]{\includegraphics[width=85mm]{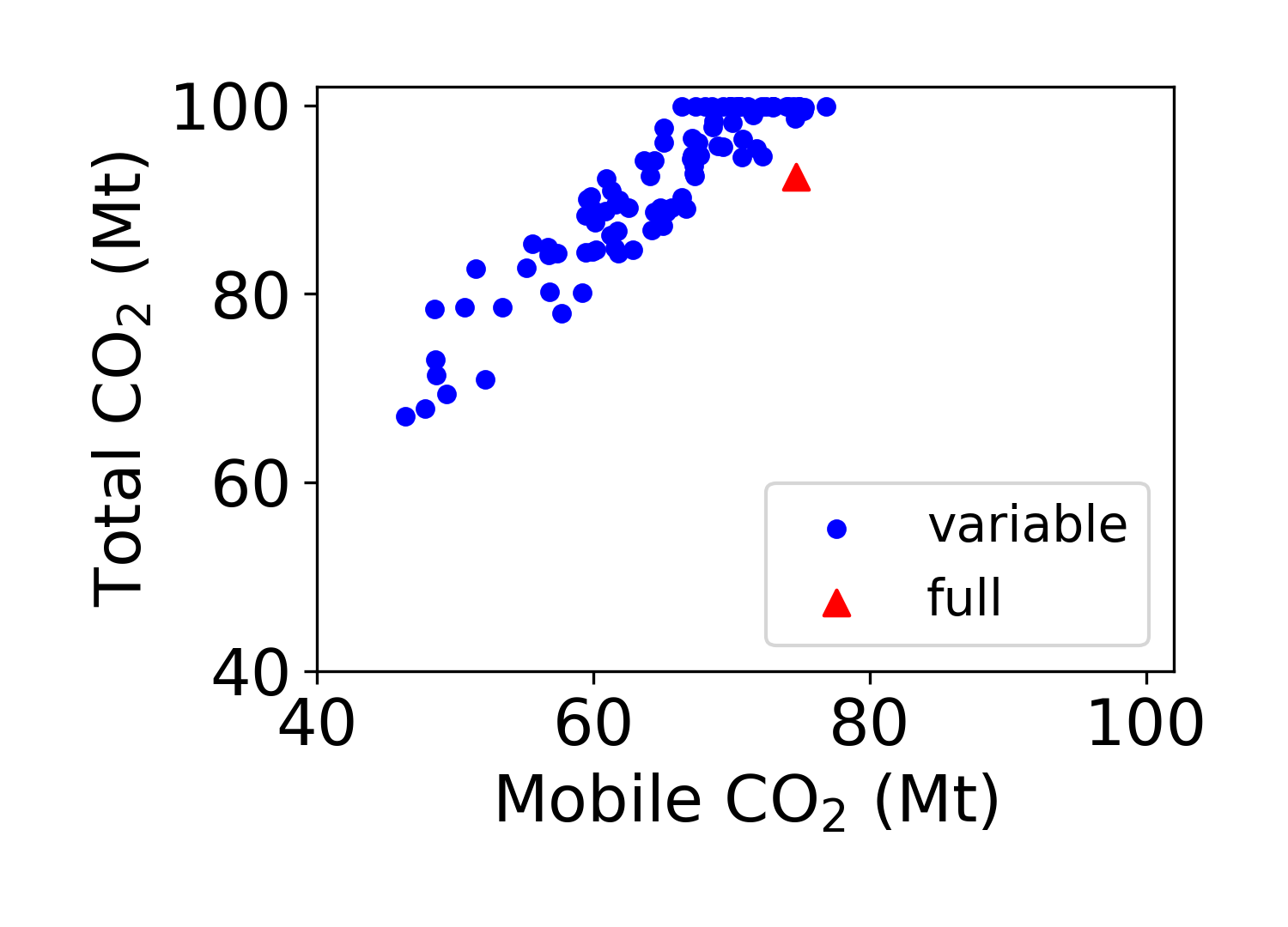}}
\hspace{4mm}
\subfloat[Geomodel realization~1 at 100~years]{\includegraphics[width=85mm]{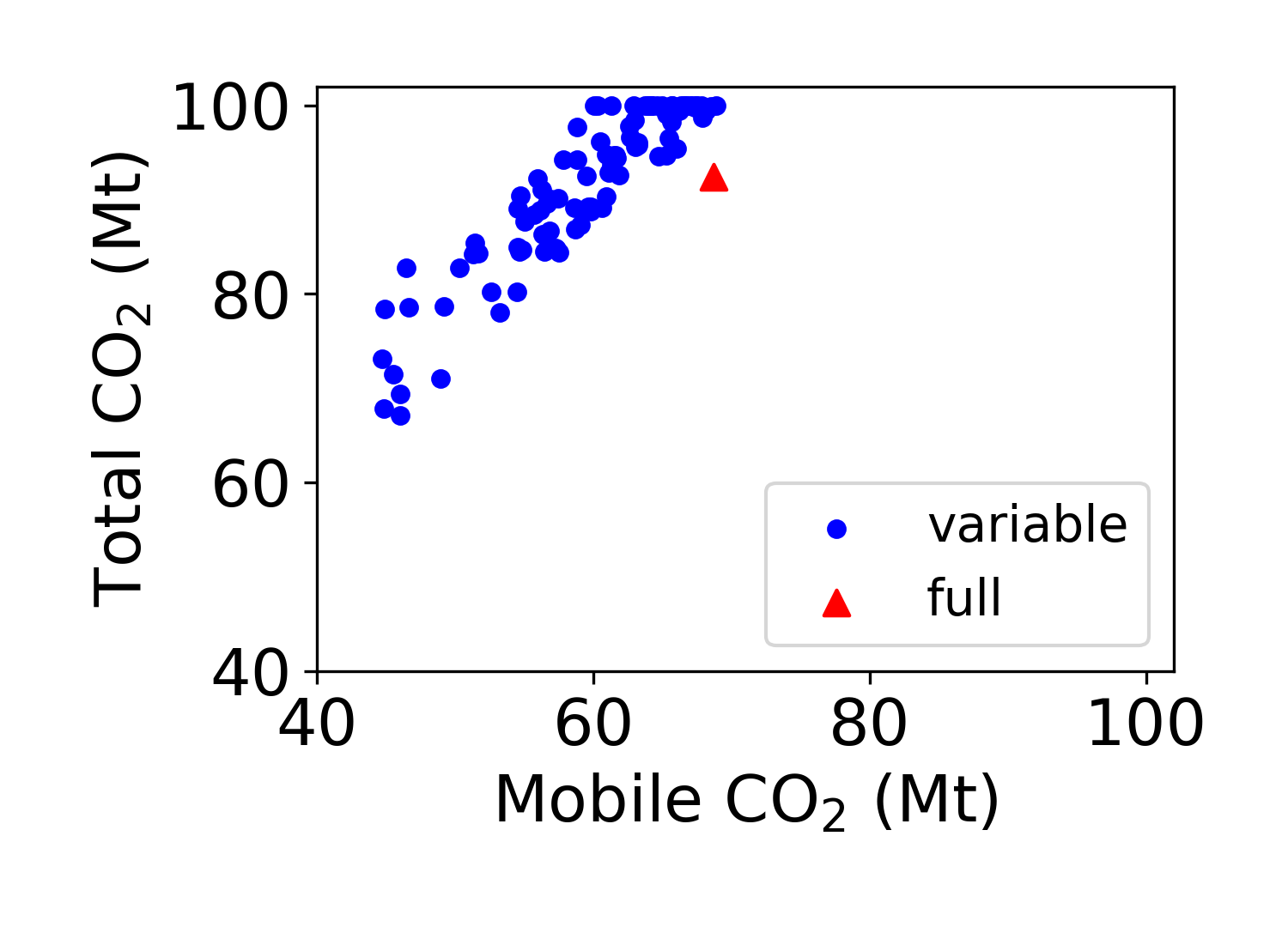}}
\\[1ex]
\caption{Total injected CO$_2$ mass and mobile CO$_2$ mass in the overall domain at the end of the injection period (50~years) and the end of the equilibration period (100~years) for geomodel realization~1 for 100 randomly generated variable perforation and injection strategies. Red triangles indicate the values for the base case involving fully perforating  wells and constant injection rates.}
\label{mobile_injection}
\end{figure}

Results for this setup are shown in Fig.~\ref{mobile_injection}. The red triangles indicate base case results involving wells that are fully perforated over the full simulation time frame, with each well prescribed to inject at a constant rate of 1~Mt/year. Because the wells reach the BHP limit, the total injected mass of CO$_2$ is 92.4~Mt at 50~years (rather than 100~Mt). The mobile CO$_2$ mass for this case is 74.7~Mt at 50~years and 68.7~Mt at 100~years. It is evident that, among the 100 (random) variable operation strategies considered, several achieve a cumulative injection of 100~Mt of CO$_2$, and many of these correspond to lower amounts of mobile CO$_2$ than the base case. The best such case results in 100~Mt of CO$_2$ injected, and mobile CO$_2$ mass of 66.4~Mt at 50~years and 60.1~Mt at 100~years. This case is superior to the base case in all three metrics, thus illustrating the potential benefits of the variable perforation and injection strategy considered in this study.  
 
\section{Multimodal Auto-regressive Transformer Model}
\label{Surrogate Model}
\begin{figure}[!ht]
\centering  
\includegraphics[width=16.5cm]{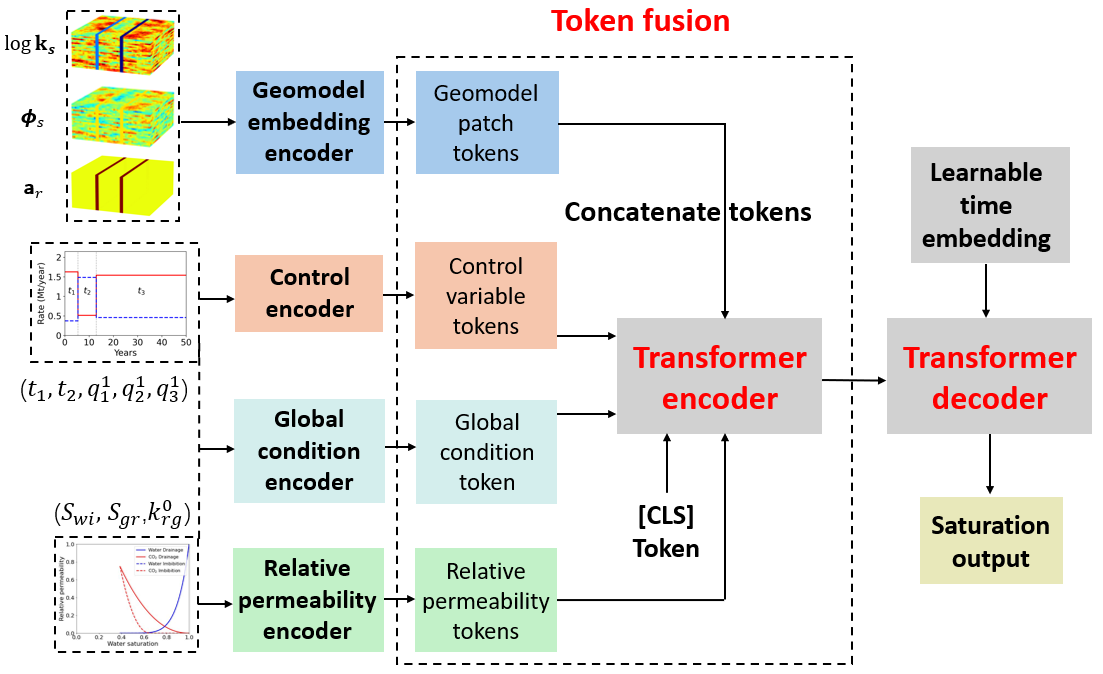}
\caption{Schematic of the multimodal auto-regressive architecture for predicting the temporal evolution of pressure and saturation at the monitoring locations, the total CO$_2$ injected, the mass of mobile CO$_2$, and the saturation footprint associated with each injection well.}
\label{Multi_Modal}
\end{figure}

The multimodal autoregressive transformer surrogate model developed in this study is able to treat uncertainty in the geomodel and relative permeability functions, along with variable control parameters characterizing the perforation and injection strategy. The multimodal architecture is well-suited for this problem because the surrogate model must fuse 3D spatial geomodel features with scalar parameters characterizing relative permeability curves, and time-dependent operational variables. The new surrogate model, shown schematically in Fig.~\ref{Multi_Modal}, integrates multiple input modalities and processes them through separate encoders. These include a geomodel embedding encoder for the 3D porosity, permeability, and permeability anisotropy ratio fields, a control encoder for the control variables, a relative permeability encoder for parameters characterizing the relative permeability (with hysteresis) curves, and a global condition encoder that processes the relative permeability and control parameters to produce a global latent representation. Details of these four encoders are provided in Tables~\ref{tab:patch_encoder},~\ref{tab:injection_encoder},~\ref{tab:relperm_encoder}, and~\ref{tab:global_encoder}. The encoded tokens from the encoders are then fused and input into a transformer encoder, which integrates them through multihead self-attention. Details of the multimodal auto-regressive architecture are provided in Table~\ref{table:multimodal_architecture}. In the tables, $N = 8$ denotes the batch size used during training, and $n_t = 20$ is the number of time steps at which predictions are provided. These predictions are generated every five years over the total simulation time frame of 100~years.

\begin{table}[!ht]
\footnotesize
\begin{center}
\caption{Architecture of the geomodel embedding encoder}
\label{tab:patch_encoder}
\renewcommand{\arraystretch}{1.15}
\begin{tabular}{ c @{\hspace{16mm}} c } 
\hline
\textbf{Layer} & \textbf{Output Shape} \\ 
\hline
input (geomodel) & ($N$, 3, 25, 48, 48) \\ 
\hline
Conv3D, 3 filters of size $3\times3\times3$, stride 1, padding 1, GELU & ($N$, 3, 25, 48, 48) \\ 
Conv3D, 3 filters of size $3\times3\times3$, stride 1, padding 1, GELU & ($N$, 3, 25, 48, 48) \\ 
Conv3D, 128 filters of size $5\times8\times8$, stride $(5,8,8)$ & ($N$, 128, 5, 6, 6) \\ 
flatten patches + positional embedding + LayerNorm & ($N$, 180, 128) \\ 
\hline
\end{tabular}
\end{center}
\end{table}

\begin{table}[!ht]
\footnotesize
\begin{center}
\caption{Architecture of the control encoder}
\label{tab:injection_encoder}
\renewcommand{\arraystretch}{1.15}
\begin{tabular}{ c @{\hspace{16mm}} c } 
\hline
\textbf{Layer} & \textbf{Output Shape} \\ 
\hline
input (control variables) & ($N$, 5) \\ 
\hline
linear layer + ReLU + LayerNorm & ($N$, 5, 128) \\ 
\hline
\end{tabular}
\end{center}
\end{table}

\begin{table}[!ht]
\footnotesize
\begin{center}
\caption{Architecture of the relative permeability encoder}
\label{tab:relperm_encoder}
\renewcommand{\arraystretch}{1.15}
\begin{tabular}{ c @{\hspace{16mm}} c } 
\hline
\textbf{Layer} & \textbf{Output Shape} \\ 
\hline
input (relative permeability parameters) & ($N$, 3) \\ 
\hline
linear layer + ReLU + LayerNorm & ($N$, 3, 128) \\ 
\hline
\end{tabular}
\end{center}
\end{table}

\begin{table}[!ht]
\footnotesize
\begin{center}
\caption{Architecture of the global condition encoder}
\label{tab:global_encoder}
\renewcommand{\arraystretch}{1.15}
\begin{tabular}{ c @{\hspace{16mm}} c } 
\hline
\textbf{Layer} & \textbf{Output Shape} \\ 
\hline
\begin{tabular}[c]{@{}l@{}}input (control variables and relative permeability parameters)\end{tabular} & ($N$, 8) \\ 
\hline
linear layer + ReLU + LayerNorm & ($N$, 1, 128) \\ 
\hline
\end{tabular}
\end{center}
\end{table}

\begin{table}[!ht]
\footnotesize
\begin{center}
\caption{Architecture of the multimodal auto-regressive transformer model}
\label{table:multimodal_architecture}
\renewcommand{\arraystretch}{1.15}
\begin{tabular}{ c @{\hspace{16mm}} c @{\hspace{16mm}} c } 
\hline
\textbf{Network} & \textbf{Layer} & \textbf{Output Shape} \\ 
\hline
\multirow{3}{*}{Input} 
 & geomodel & ($N$, 3, 25, 48, 48) \\ 
 & control variables & ($N$, 5) \\
 & relative permeability parameters & ($N$, 3) \\
\hline
\multirow{4}{*}{Input Encoder} 
 & geomodel embedding encoder & ($N$, 180, 128) \\
 & control encoder & ($N$, 5, 128) \\
 & relative permeability encoder & ($N$, 3, 128) \\
 & global condition encoder & ($N$, 1, 128) \\
\hline
\multirow{2}{*}{\begin{tabular}[c]{@{}c@{}}Token Fusion\\ Transformer Encoder\end{tabular}} 
 & concatenation + token type embeddings & ($N$, 190, 128) \\
 & transformer encoder (4 blocks, 8 heads) & ($N$, 190, 128) \\
\hline
\multirow{3}{*}{Temporal Decoder} 
 & learnable time embedding ($n_t = 20$) & ($N$, 20, 128) \\
 & transformer decoder (3 blocks, 8 heads) & ($N$, 20, 128) \\
 & linear projection & ($N$, 20, 8) \\
\hline
Output & Predictions for quantity of interest ($n_t = 20$) & ($N$, 20, 8) \\ 
\hline
\end{tabular}
\end{center}
\end{table}

The geomodel embedding encoder described in Table~\ref{tab:patch_encoder} operates on the 3D geomodel input. This input comprises three channels corresponding to the porosity, log-permeability, and permeability anisotropy ratio fields in the target aquifer. Two 3D convolutional layers, each with kernels of size $3\times3\times3$ and GELU activation, are first applied. These layers extract local geological features while preserving the dimensions of the input geomodels. The processed latent feature maps are then partitioned into non-overlapping 3D patches, each containing $8 \times 8 \times 5$ cells (in the $x$, $y$, and $z$ directions), using a strided 3D convolution with the kernel size and stride both equal to the patch size. This treatment, which corresponds to a 3D extension of the patch embedding strategy used in vision transformers~\citep{dosovitskiy2021image}, results in a total of $6 \times 6 \times 5 = 180$ patches. Each patch is then projected into an embedded token of dimension~128. A learnable positional embedding is added to each patch token to retain information on spatial location, and layer normalization is then applied. The resulting sequence of geomodel patch embedded tokens is denoted ${Z}_g \in \mathbb{R}^{180 \times 128}$.

The control encoder, relative permeability encoder, and global condition encoder (Tables~\ref{tab:injection_encoder}, \ref{tab:relperm_encoder}, and~\ref{tab:global_encoder}) map the scalar input variables into token representations that are dimensionally consistent with the geomodel patch embedded tokens. The control encoder acts on the five control variables in Eq.~\ref{perforation_time}. These variables are mapped, through a linear layer followed by ReLU activation and layer normalization, into five embedded tokens of dimension~128 (one embedded token per control variable), denoted ${Z}_c \in \mathbb{R}^{5 \times 128}$. The relative permeability encoder analogously maps the three relative permeability parameters, $S_{wi}$, $S_{gr}$, and $k_{rg}^0$, into three embedded tokens, denoted ${Z}_r \in \mathbb{R}^{3 \times 128}$. The global condition encoder concatenates the five control variables and the three relative permeability parameters into a single vector, which is mapped into a single global condition embedded token, denoted ${Z}_{gc} \in \mathbb{R}^{1 \times 128}$. This token provides a global representation of the overall operational strategy and relative permeability parameters. It is in addition to the tokens for each control variable and each relative permeability parameter generated by the control and relative permeability encoders.

The embedded tokens from the four encoders are concatenated, together with a learnable classification (CLS) token~\citep{devlin2019bert}, denoted ${Z}_{\mathrm{CLS}} \in \mathbb{R}^{1 \times 128}$. The CLS token provides an additional global latent representation of the geological properties and operational strategies. Because these tokens are derived from different modalities, a learnable token type embedding, which identifies each token as a geomodel, control variable, relative permeability, or global condition token, is added to each. This results in the fused embedded token sequence, denoted ${Z} = [{Z}_{\mathrm{CLS}}; {Z}_g; {Z}_c; {Z}_r; {Z}_{gc}] \in \mathbb{R}^{190 \times 128}$. 

The embedded token sequence ${Z}$ is then processed by the transformer encoder with four blocks and eight attention heads, as shown in Table~\ref{table:multimodal_architecture}, following the standard transformer architecture \citep{vaswani2017attention}. Each encoder block consists of multihead self-attention and feedforward layers, with layer normalization applied before each layer to improve training stability. All attention operations involve the scaled dot-product, which is given by
\begin{equation} \label{eq:attention}
\mathrm{Attention}(Q,K,V) = \mathrm{softmax}\!\left(\frac{{Q}{K}^{T}}{\sqrt{d_k}}\right){V}.
\end{equation}
\noindent Here ${Q}$, ${K}$, and ${V}$ are the query, key, and value matrices, respectively, and $d_k$ is the dimension of each attention head (here $d_k = 16$, corresponding to the embedding dimension of 128 divided by the eight attention heads). For the self-attention applied in the encoder, the queries, keys, and values are all computed from the fused embedded token sequence ${Z}$, with
\begin{equation} \label{eq:self_attention}
{Q}={Z}{W}^{Q}, \quad K=ZW^{K}, \quad V=ZW^{V},
\end{equation}
\noindent where ${W}^{Q}, {W}^{K}, {W}^{V} \in \mathbb{R}^{128 \times 16}$ are learnable projection matrices for each attention head. The outputs of the eight attention heads are concatenated, resulting in an embedding dimension of 128. Through self-attention, the geomodel patch tokens, the control variable tokens, the relative permeability tokens, the global condition token, and the CLS token all attend to one another. This allows the model to integrate geomodel properties, relative permeability parameters, and operational control variables. The output of the transformer encoder is the encoded multimodal feature representation, denoted ${M} \in \mathbb{R}^{190 \times 128}$, which is subsequently provided to the temporal decoder through cross-attention.

Time dependence is treated through a learnable sinusoidal time embedding in the decoder, which combines learnable time queries with fixed sinusoidal encoding~\citep{vaswani2017attention, dehghani2019universal}. Specifically, the time embedding $\mathbf{e}_t \in \mathbb{R}^{128}$ for time step $t$ is given by
\begin{equation} \label{eq:time_embedding}
\mathbf{e}_t = \alpha \, \mathbf{q}_t + (1-\alpha) \, \mathbf{p}_t, \quad t = 1, \ldots, n_t,
\end{equation}
\noindent where $\mathbf{q}_t \in \mathbb{R}^{128}$ is a learnable time query, $\mathbf{p}_t \in \mathbb{R}^{128}$ is the fixed sinusoidal encoding 
%\citep{vaswani2017attention} 
at time step $t$, and $\alpha \in (0, 1)$ is a learnable gating coefficient. Layer normalization is then applied to the time embeddings. With this treatment, the model determines, during training, the appropriate balance between the learnable time representation $\mathbf{q}_t$ and the fixed encoding $\mathbf{p}_t$, which encodes the sequential ordering of the time steps. The temporal decoder with three blocks and eight attention heads generates predictions auto-regressively. Each decoder block includes masked multihead self-attention, multihead encoder–decoder cross-attention, and feedforward layers.

In the temporal decoder, at each time step, the predictions from all previous time steps are projected into the embedding space through a linear layer. At the first time step, where previous predictions are not available, a learnable begin-of-sequence embedding is used. These embeddings are then added to the corresponding time embeddings $\mathbf{e}_t$ and the global condition embedded token ${Z}_{gc}$, with the latter providing conditioning on the operational strategy and relative permeability parameters at every time step. This results in the decoder token sequence, denoted ${H} \in \mathbb{R}^{t \times 128}$, at time step $t$. The sequence ${H}$ is processed using masked self-attention, which applies Eq.~\ref{eq:attention} with a causal mask such that the prediction at time step $t$ attends only to time steps $1, \ldots, t$. This masking is required because predictions are generated sequentially, meaning future predictions are not available at inference time. The output of this masked self-attention is denoted $\tilde{{H}} \in \mathbb{R}^{t \times 128}$.

The encoder-decoder cross-attention then enables each decoder block to attend to the encoded multimodal feature representation ${M}$. In this cross-attention case, the queries are computed from $\tilde{{H}}$, while the keys and values are computed from ${M}$, with
\begin{equation} \label{eq:cross_attention}
{Q}=\tilde{{H}}{W}^{Q}, \quad {K}={M}{W}^{K}, \quad {V}={M}{W}^{V}.
\end{equation}
A final linear layer maps each decoder hidden state to the quantities of interest at the corresponding time step, generating the predictions sequentially in time. The procedures used to train the multimodal auto-regressive transformer surrogate model are described in the next section.

\section{Surrogate Model Training and Evaluation} 
\label{Surrogate Evaluation}
In this section we first describe the detailed surrogate model training procedures. We then present the evaluation metrics used to assess model performance and compare error statistics for varying numbers of training samples. Comparisons between surrogate model predictions and reference GEOS simulations, both in terms of ensemble statistics and detailed results for a particular realization, are then presented. 

\subsection{Surrogate model training and accuracy assessment}
\label{sec:training}

Five networks, one for each quantity of interest, are trained separately. Four of the quantities of interest are predicted at 20 discrete time steps, corresponding to every five years over the 100-year simulation time frame. These networks provide (1) saturation at four different layers in the target aquifer at the two monitoring wells, (2) pressure in two layers of the target aquifer and in the top layers of the middle and upper aquifers at the two monitoring wells, (3) the mass of mobile CO$_2$ in the overall domain, and (4) the saturation footprint in the target aquifer associated with each injector. The fifth network, which provides the total mass of injected CO$_2$, is trained only for the 50-year injection period, corresponding to 10 time steps. 

Saturation footprint can be computed in various ways. Here we essentially apply the metric used by \citet{tang2025graph}, though we modify it to account for the fact that the injected CO$_2$ does not, in general, extend over the full aquifer thickness. Specifically, we define footprint as the bulk volume of a box-shaped, grid-aligned region enclosing each plume. These boxes do not extend vertically over the full target aquifer unless the plume itself spans the full aquifer thickness. Consistent with the earlier study, we apply a saturation threshold of 0.05 in these computations.  

The network training procedures are as follows. Normalization is applied prior to the training process. All input modalities and variables are normalized using the training-set mean and standard deviation for each quantity or field. All time-series output quantities of interest are normalized using the training-set mean and standard deviation computed separately at each time step. The training procedure entails determining optimal network parameters that minimize the difference between GEOS simulation results and surrogate model predictions. Specifically, the Smooth-$L_1$ loss between simulated and predicted (normalized) quantities of interest at all time steps is minimized. 

Importantly, scheduled sampling is applied within the autoregressive decoder to mitigate exposure bias, in which surrogate model prediction errors can accumulate because the decoder uses simulated results during training but its own predictions during rollout. Specifically, at each time step, the decoder input is taken to be the simulated result with probability equal to the scheduled sampling ratio, and the model prediction otherwise. During training, the scheduled sampling ratio decreases linearly from 1.0 to 0 over the first 30\% of the total training steps, after which the model relies entirely on its own predictions for auto-regressive rollout. This means that the decoder initially uses simulation results and then gradually transitions to using its own predictions.

The initial learning rate for the AdamW optimizer is set to $2\times 10^{-4}$, and a weight decay of $10^{-4}$ is applied to all weight matrices (bias and layer normalization parameters are excluded). A cosine-annealing schedule with 5~warm-up epochs (to prevent early overfitting) is applied, and the learning rate decays to a minimum value of $10^{-6}$ over a total of 120~epochs. A batch size of~8 is used, and gradient clipping with a threshold of 2 is applied to stabilize optimization. Training is performed using mixed-precision arithmetic to reduce memory requirements and computational cost. Early stopping with a patience of 12~epochs is applied based on the validation loss. Training is terminated once the loss plateaus. The maximum number of training samples considered is $n_f = 4000$. We specify a 9/1 training-validation split. Training the five multimodal autoregressive encoder-decoder transformer models with $n_f = 4000$ requires approximately 3.2~hours for saturation (91~epochs), 2.8~hours for pressure (67~epochs), 1~hour for total injected CO$_2$ (53~epochs), 3.2~hours for the mass of mobile CO$_2$ (91~epochs), and 3.6~hours for the saturation footprints (114~epochs) on a single NVIDIA A100 GPU. This corresponds to a total (serial) training time of about 14~hours.

We now evaluate the impact of the number of training samples ($n_f$) on surrogate model performance. The values considered are $n_f=1000$, 2000, 3000, and 4000. A new test set of $n_e=500$ geomodel realizations is then generated and simulated to evaluate the performance of the surrogate models. These entail new samples of all 10 metaparameters $\boldsymbol{\uptheta}_{\mathrm {meta}}$, which include permeability, porosity and relative permeability parameters, along with randomly sampled PCA parameters. A separate (distinct) perforation and injection strategy is also randomly sampled for each test case. 

For saturation, we compute the mean absolute error (MAE) at monitoring locations. Consistent with \citet{han2026recurrent}, contributions to this error are computed only where either the simulated saturation or the surrogate prediction exceeds a threshold value of 0.02 (otherwise this error would be artificially low because saturation is zero at many locations). Specifically, for test sample $i$ ($i = 1, \ldots, 500$), saturation MAE, denoted $\delta_S^i$, is  given by
\begin{equation} \label{surr_error_s}
\delta_S^i = \frac{1}{n_i n_t} \sum_{j=1}^{n_i} \sum_{t=1}^{n_t} \left|(\hat{S}_d)_{i,j}^t - (S_d)_{i,j}^t \right|, \quad \text{for } (S_d)_{i,j}^t > 0.02 \ \ \text{or} \ \ (\hat{S}_d)_{i,j}^t > 0.02.
\end{equation}
Here $(S_d)_{i,j}^t$ and $(\hat{S}_d)_{i,j}^t$ are CO$_2$ saturation predictions from the GEOS simulation and surrogate model, for test case $i$, monitoring location $j$, and surrogate model time step $t$, and $n_i$ is the number of saturation monitoring locations for which $(S_d)_{i,j}^t > 0.02$ or $(\hat{S}_d)_{i,j}^t > 0.02$. We note that previous studies have shown little sensitivity to the threshold saturation value over a reasonable range.

Relative pressure error for test sample $i$, denoted $\delta_p^i$, is calculated as 
\begin{equation} \label{surr_error_p}
\delta_p^i = \frac{1}{n_dn_t} \sum_{j=1}^{n_d} \sum_{t=1}^{n_t} \frac{| (\hat{p}_d)_{i,j}^t - (p_d)_{i,j}^t |}{(p_d)_{i,\mathrm{max}}^t - (p_d)_{i,\mathrm{min}}^t}.
\end{equation}
Here $(p_d)_{i,j}^t$ and $(\hat{p}_d)_{i,j}^t$ are pressure results from GEOS simulation and the surrogate model for test case $i$, monitoring location $j$, and surrogate output time step $t$, and $(p_d)_{i,\mathrm{max}}^t$ and $(p_d)_{i,\mathrm{min}}^t$ are the maximum and minimum simulated pressure for all monitoring locations in test case $i$ at time step $t$. This normalization acts to avoid artificially low errors, since the absolute pressure values can be large relative to the pressure range ($(p_d)_{i,\mathrm{max}}^t - (p_d)_{i,\mathrm{min}}^t$). The surrogate model errors for the total injected CO$_2$, the mass of mobile CO$_2$ in the overall domain, and the CO$_2$ saturation footprint associated with each injection well are evaluated using relative error equations of the same form as Eq.~\ref{surr_error_p}. The denominators in these relative error computations are the differences between the corresponding maximum and minimum values from the GEOS simulations, as in Eq.~\ref{surr_error_p}.

\begin{figure}[!ht]
\centering   
\subfloat[Saturation MAE]{\includegraphics[width = 80mm]{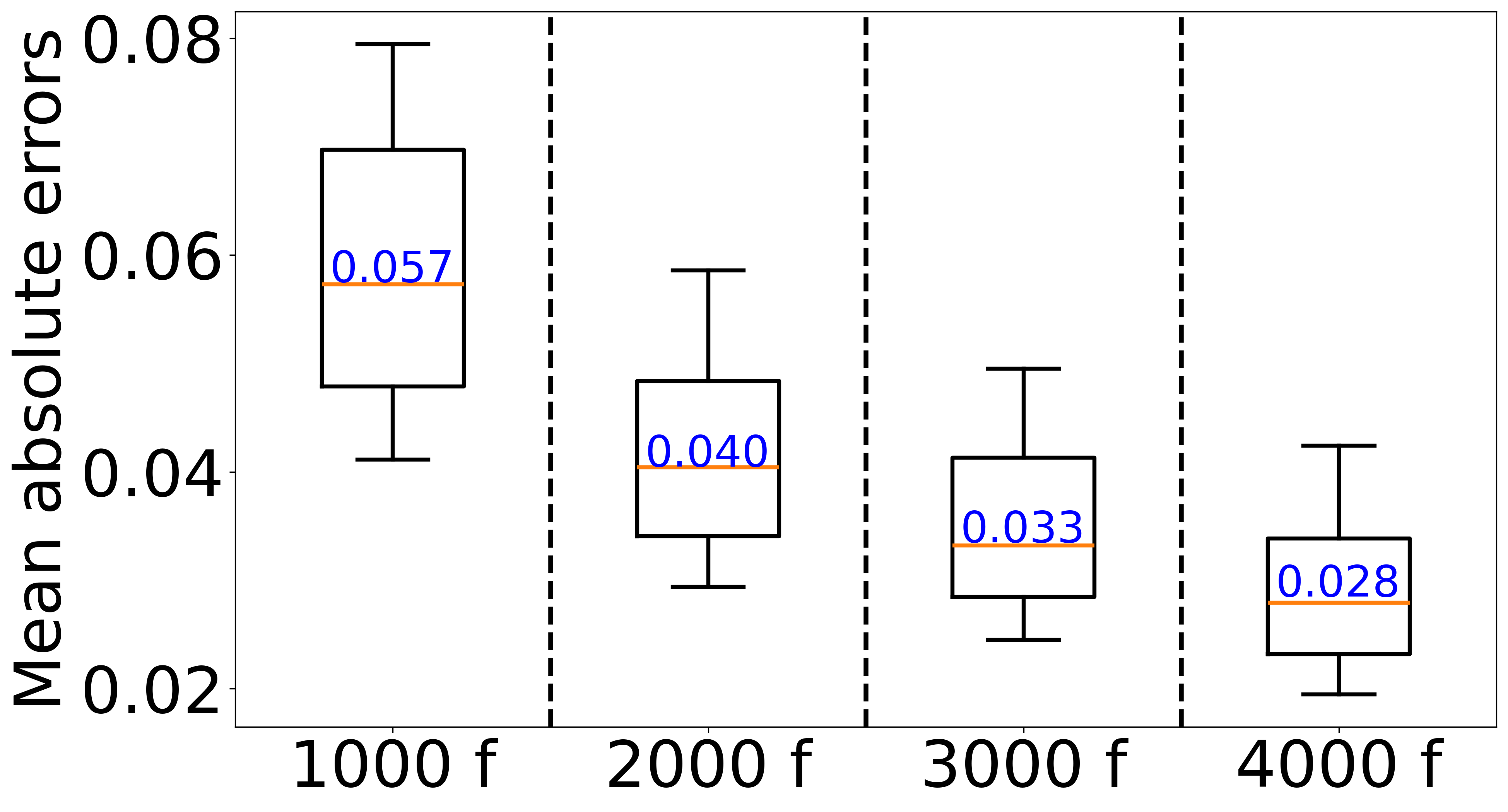}}
\hspace{10mm}
\subfloat[Pressure relative error]{\includegraphics[width = 80mm]{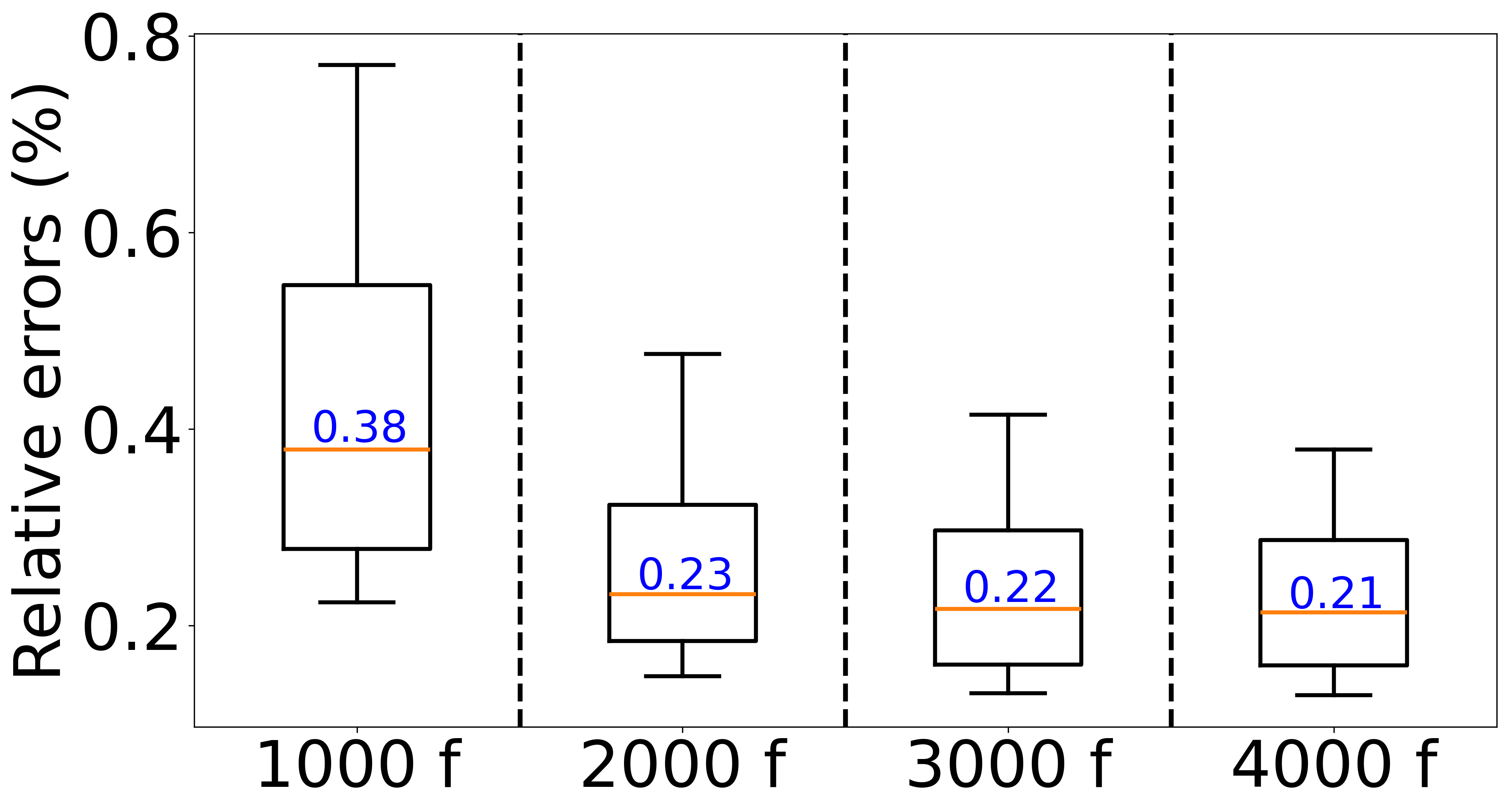}} \\
\subfloat[I1 saturation footprint relative error]{\includegraphics[width=80mm]{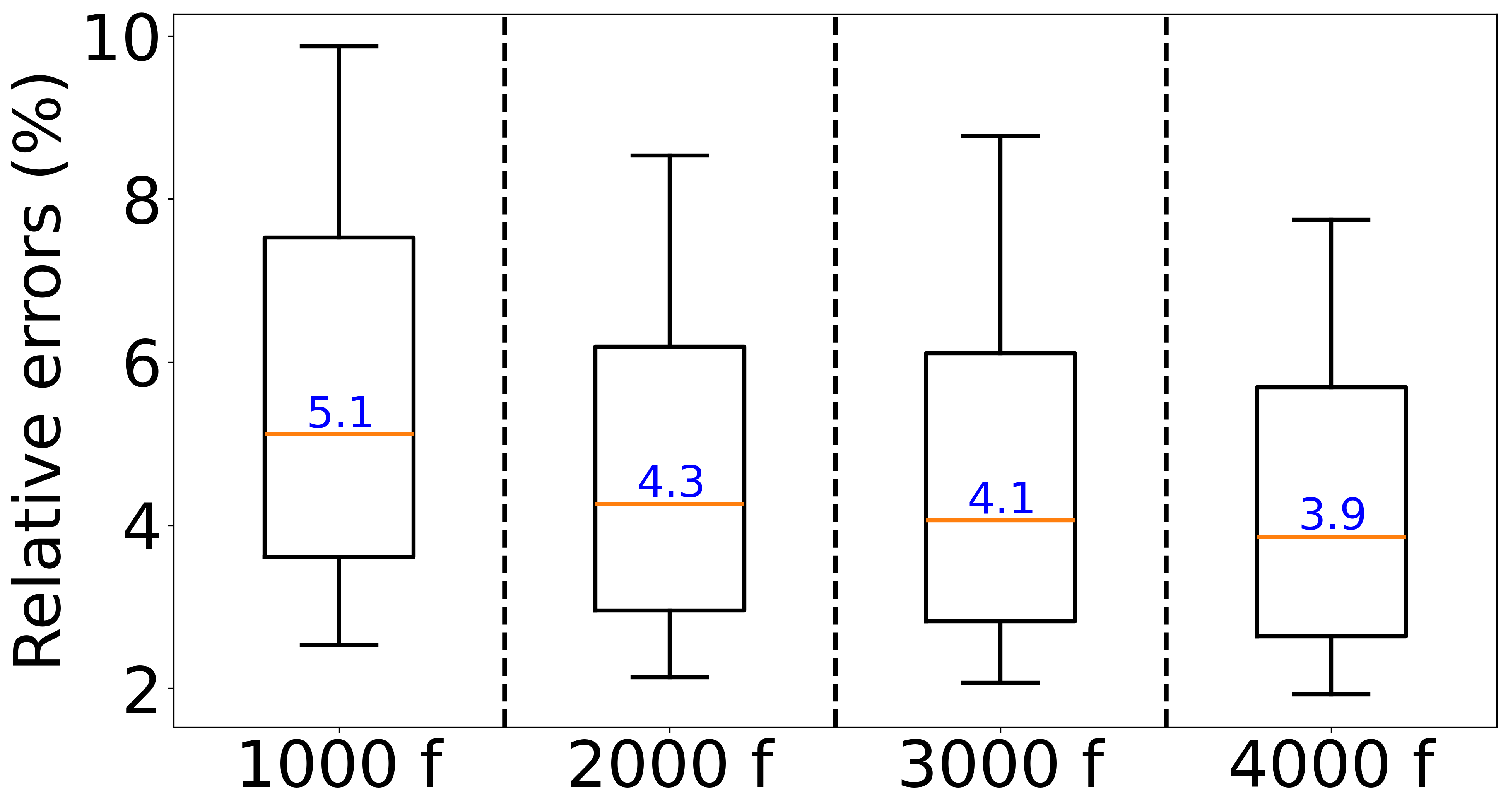}}
\hspace{10mm}
\subfloat[I2 saturation footprint relative error]{\includegraphics[width=80mm]{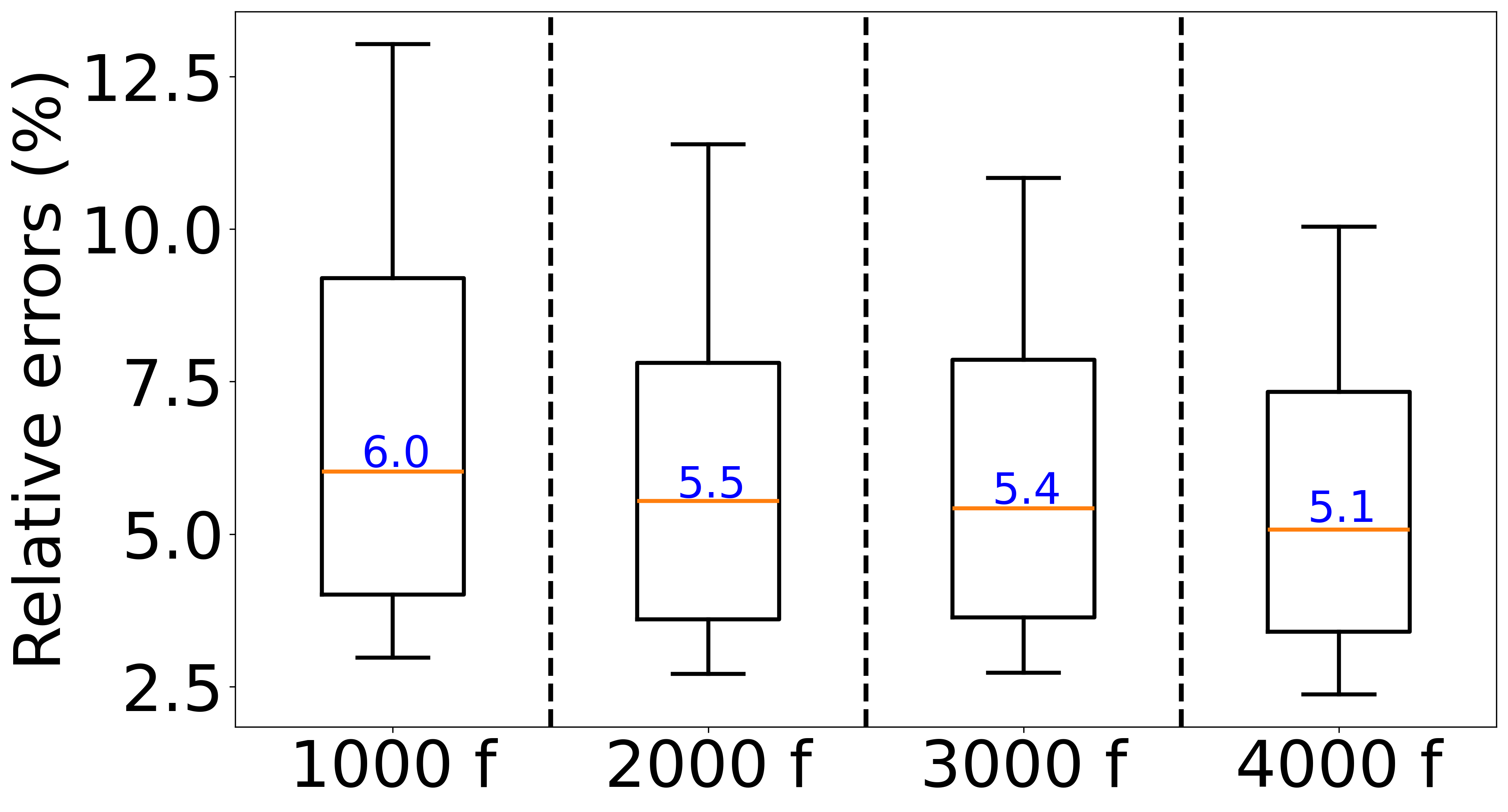}} \\
\subfloat[Total mass of CO$_2$ relative error]{\includegraphics[width=80mm]{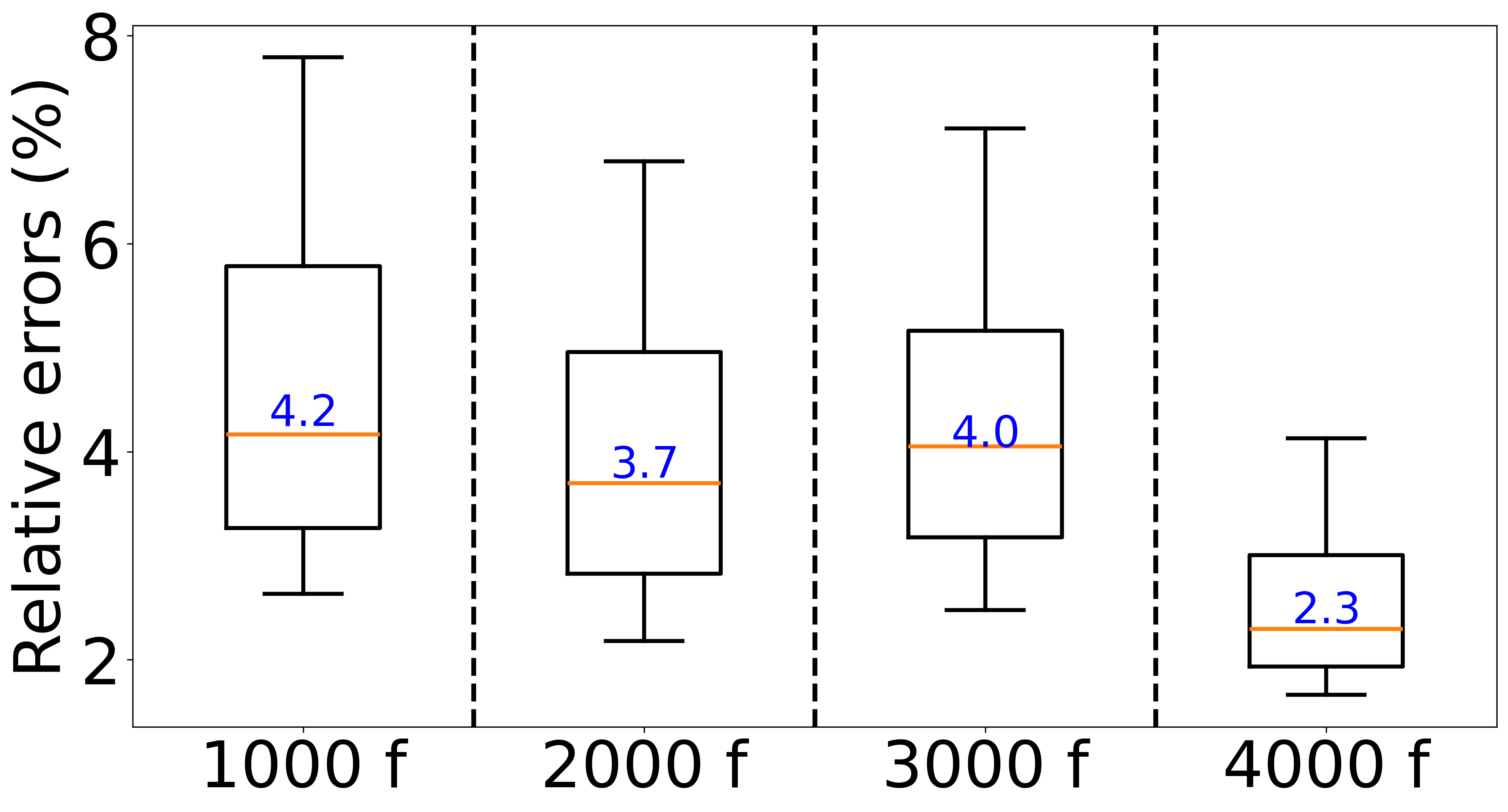}}
\hspace{10mm}
\subfloat[Mobile mass of CO$_2$ relative error]{\includegraphics[width=80mm]{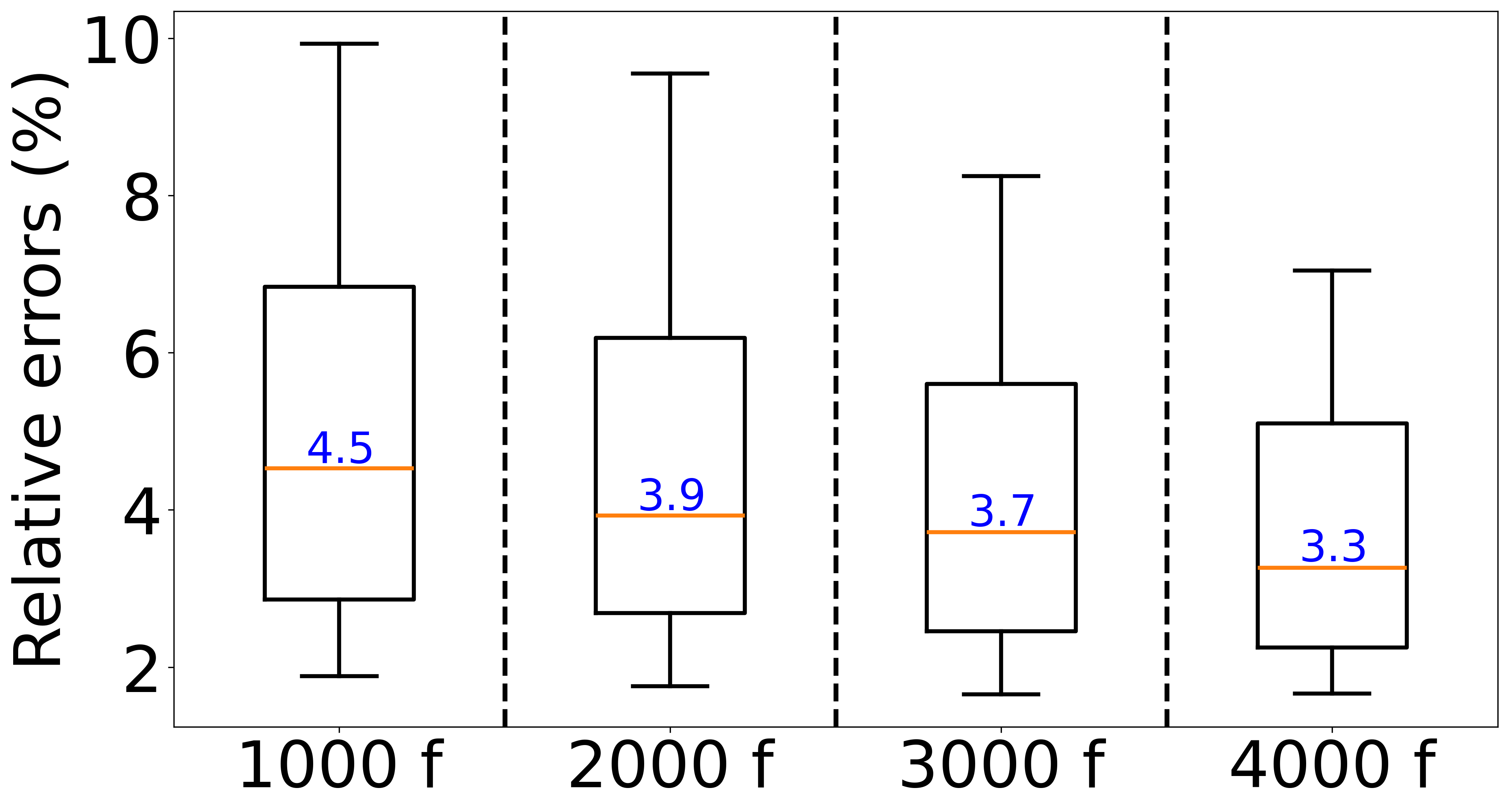}} \\
\caption{Saturation MAE and relative errors for pressure, saturation footprint associated with each injector, the total mass of injected CO$_2$, and the mass of mobile CO$_2$, for the 500~test cases. Results shown for surrogate models trained with 1000 flow simulation runs (1000~f), 2000 flow simulation runs (2000~f), 3000 flow simulation runs (3000~f), and 4000 flow simulation runs (4000~f). Boxes display P$_{90}$, P$_{75}$, P$_{50}$, P$_{25}$ and P$_{10}$ errors, numbers indicate the P$_{50}$ error.}
\label{multi_modal_errors_box}
\end{figure}

The errors for the six quantities, in terms of box plots, are shown in Fig.~\ref{multi_modal_errors_box}. The numbers in each box give the median error (indicated by the red lines). The upper and lower edges of the box indicate the P$_{75}$ and P$_{25}$ percentile errors, and the lines extending above and below the boxes show the P$_{90}$ and P$_{10}$ percentile errors. Overall, increasing the number of training samples generally leads to improved surrogate model performance, though the degree of improvement varies for the different quantities. Saturation exhibits the most substantial reduction in error, with the median MAE decreasing from 0.057 to 0.028. For the total injected mass of CO$_2$, the relative error decreases from 4.2\% to 2.3\%. The reductions in the other quantities are less pronounced, i.e., from 0.38\% to 0.21\% for pressure, from 4.5\% to 3.3\% for the mass of mobile CO$_2$, from 5.1\% to 3.9\% for the saturation footprint associated with I1, and from 6.0\% to 5.1\% for the saturation footprint associated with I2. In all subsequent results, we will use the surrogate model trained with 4000 samples.

\subsection{Comparison of ensemble statistics and individual realization results} 
\label{sec:ensemble_results}

\begin{figure}[!ht]
\centering
\subfloat[Saturation at O1 in layer 1]{\includegraphics[width = 80mm]{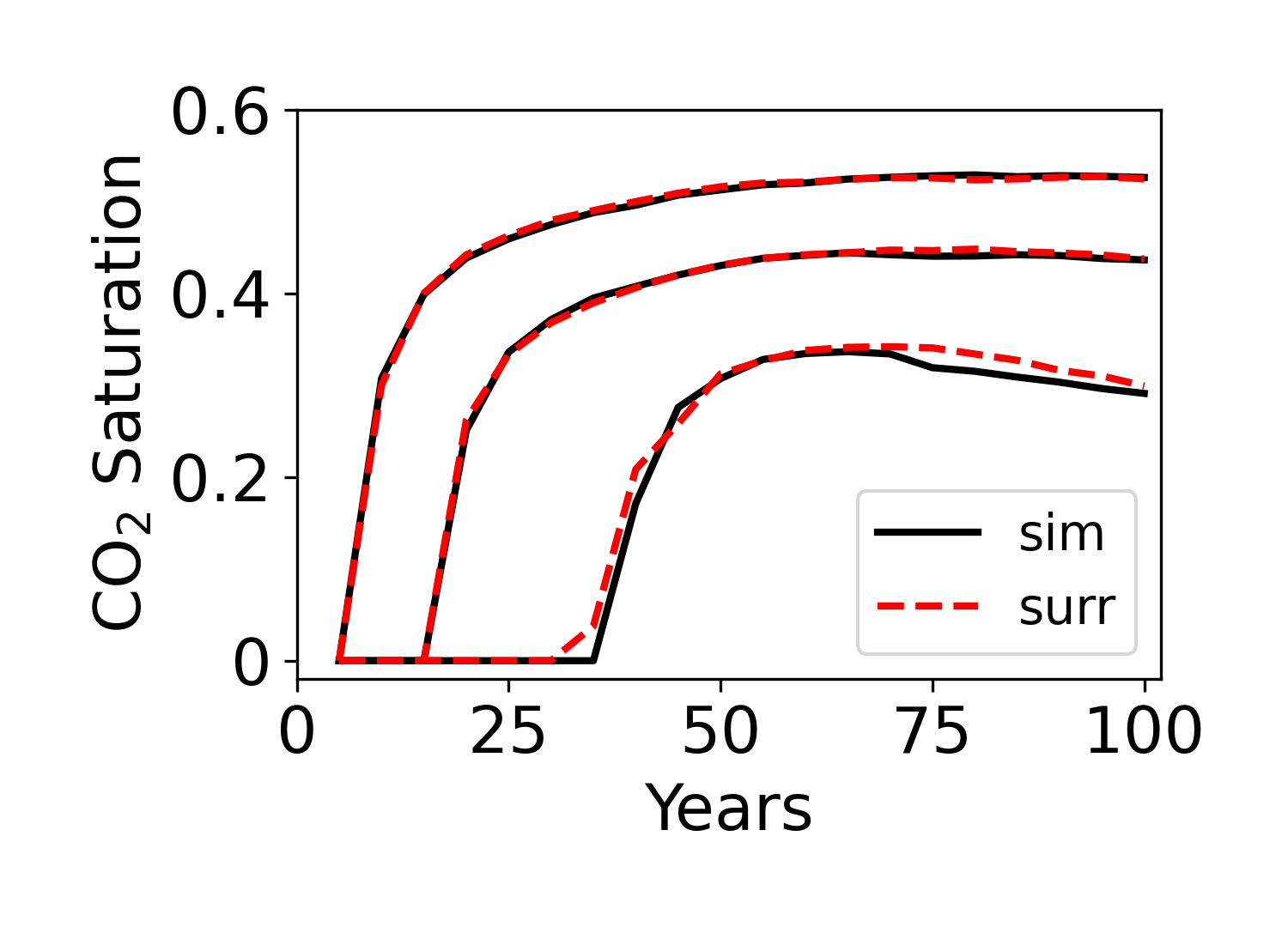}}
\hspace{6mm}
\subfloat[Saturation at O2 in layer 7]{\includegraphics[width = 80mm]{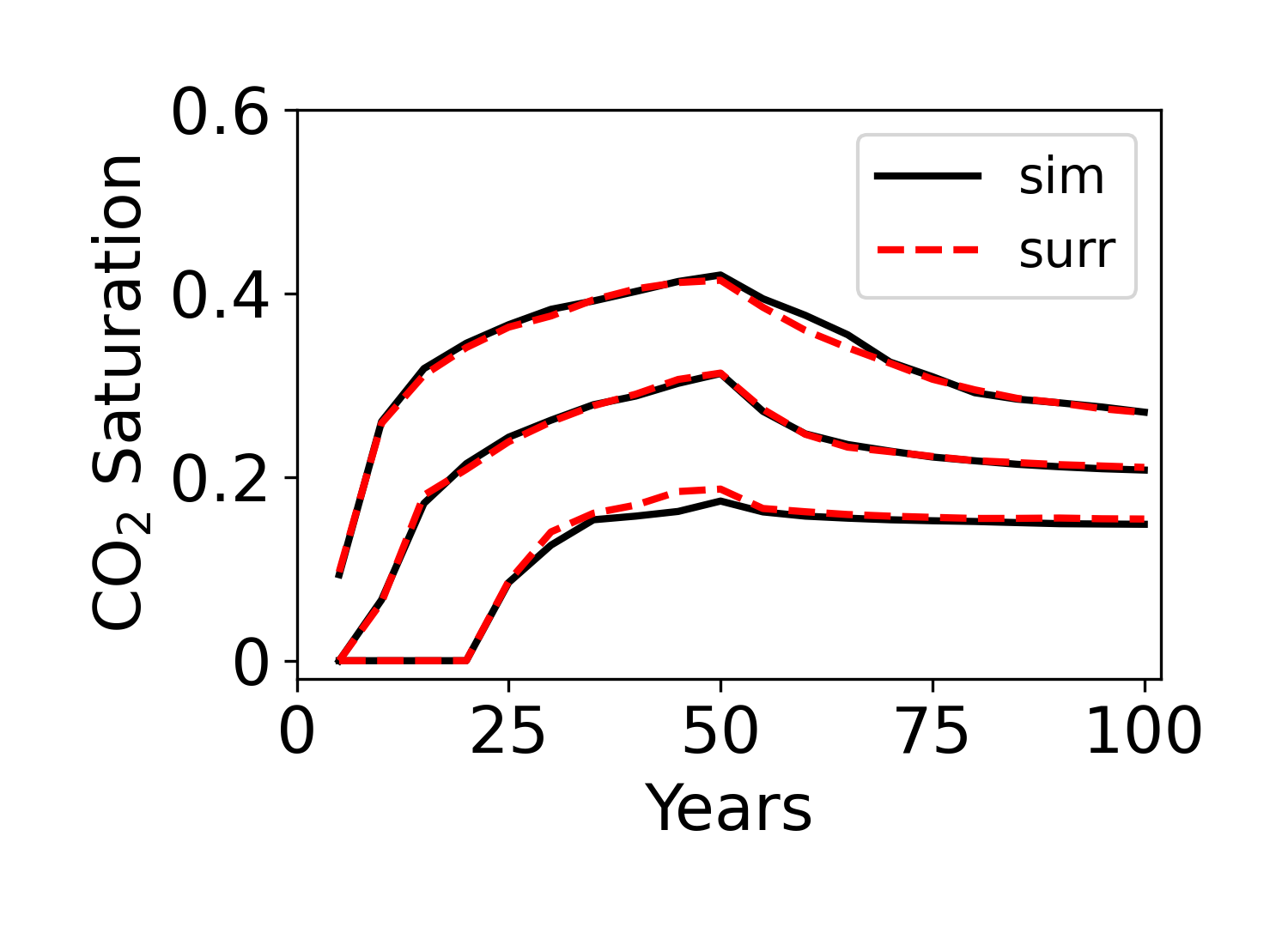}}
\\[1ex]
\subfloat[Saturation at O2 in layer 13]{\includegraphics[width = 80mm]{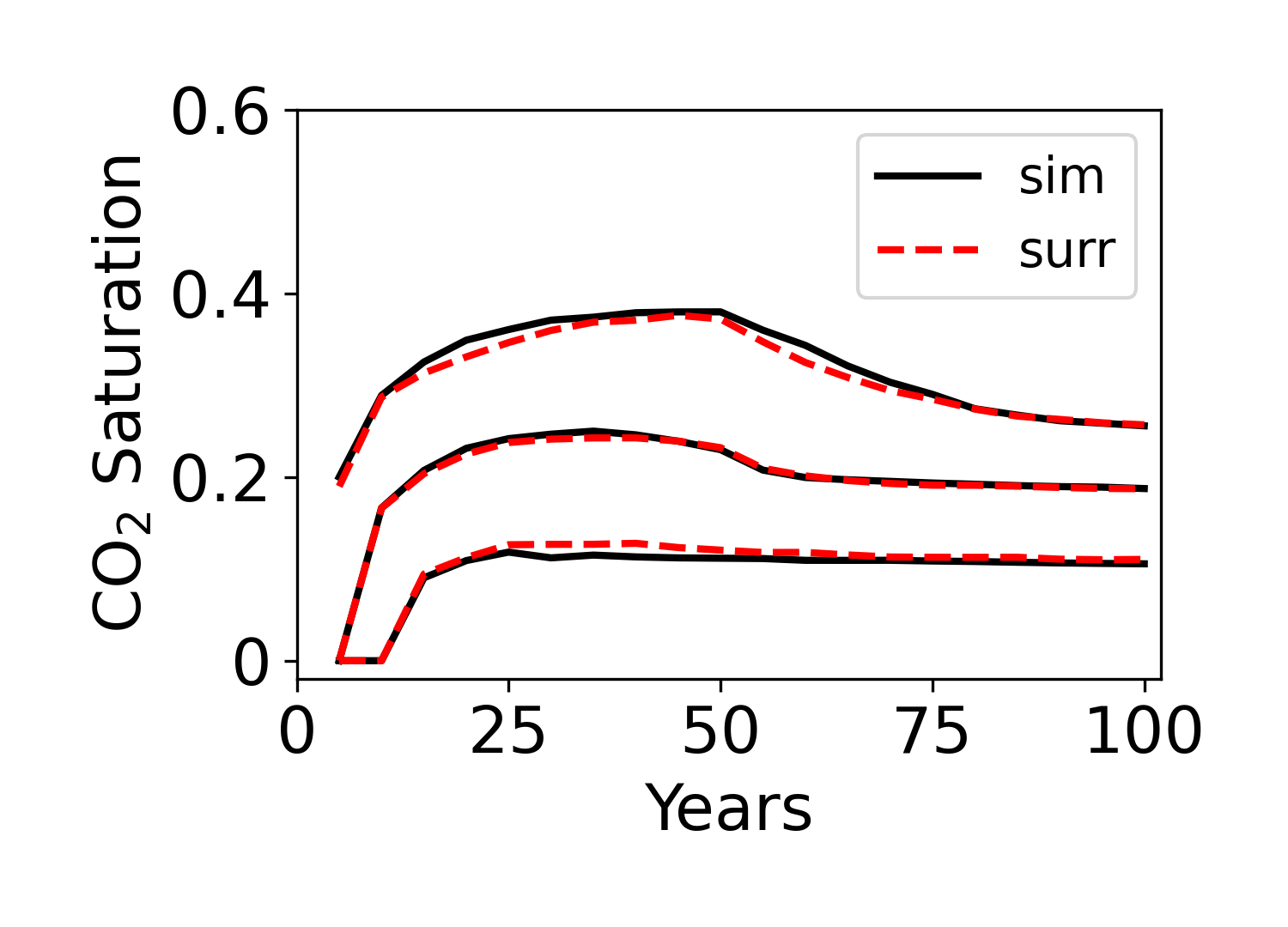}}
\hspace{6mm}
\subfloat[Saturation at O2 in layer 19]{\includegraphics[width = 80mm]{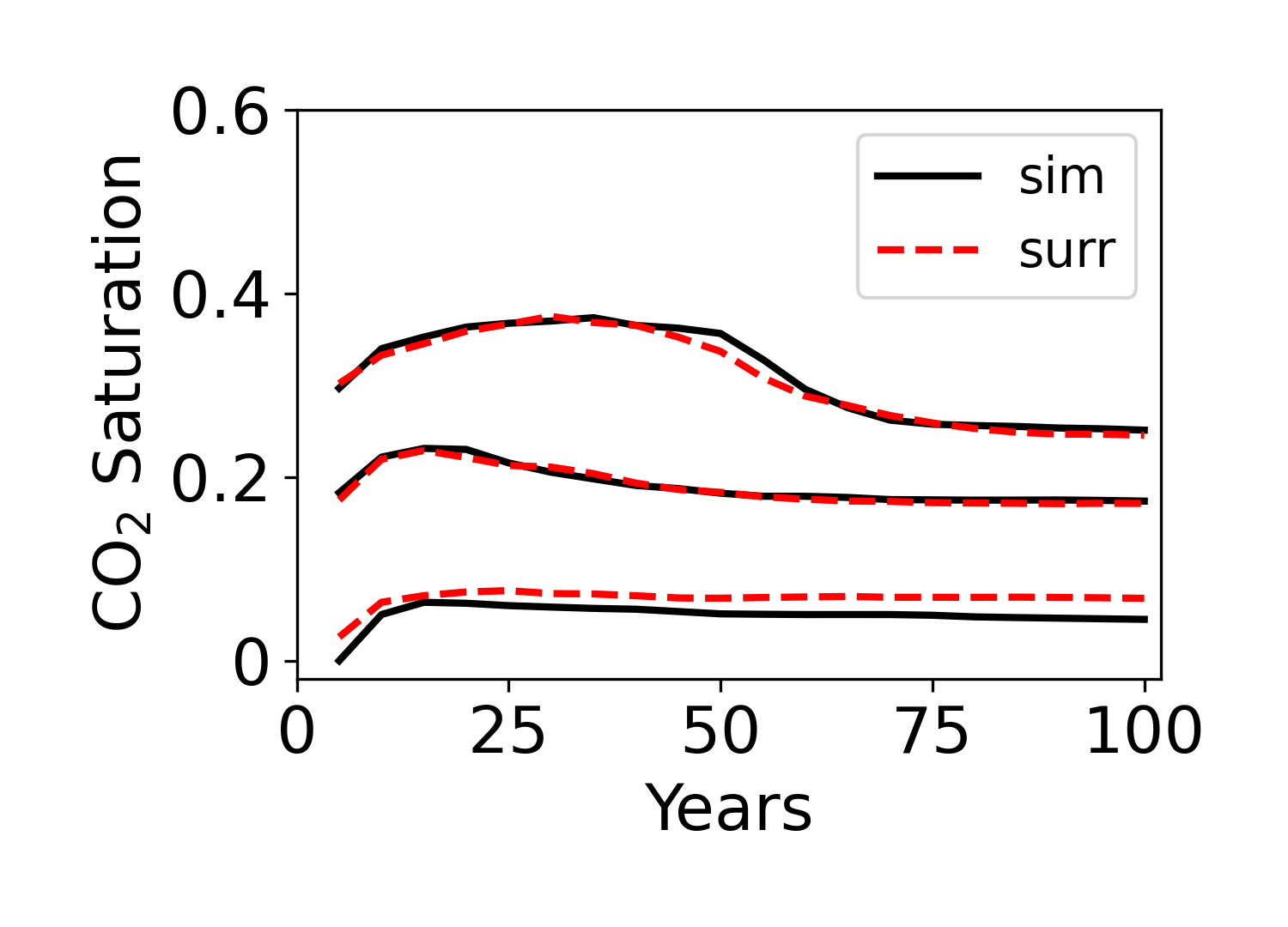}}
\\[1ex]
\caption{Saturation ensemble statistics from GEOS (black solid curves) and surrogate model (red dashed curves) at four observation locations in the target aquifer. The upper, middle and lower curves correspond to P$_{90}$, P$_{50}$ and P$_{10}$ results over the 500 test cases.}
\label{multi_modal_statistics_s}
\end{figure}

To assess the overall statistical correspondence between the GEOS and surrogate model results, we now present ensemble statistics through time for saturation and pressure, in different layers at the observation wells. Results for saturation are shown in Fig.~\ref{multi_modal_statistics_s}. The solid black curves represent GEOS results, and the dashed red curves are surrogate model results at each of the 20~time steps. The upper curves are the P$_{90}$ results, the middle curves are the P$_{50}$ results, and the lower curves the P$_{10}$ results. Close agreement is observed between the two sets of curves, which demonstrates statistical consistency in saturation results at the monitoring well locations. Layer~1 is at the top of the target aquifer. Note that in many cases CO$_2$ appears later in layers~1 and~7, consistent with the fact that injection is initially in the lower portion of the aquifer. Some of curves have a kink at 50~years, corresponding to the end of injection. After 50~years, saturation generally decreases in time except in layer~1. The decrease in saturation in the lower layers is due to both dissolution and vertical migration. CO$_2$ migrates upward into layer~1 (where it is trapped), which acts to increase saturation in that layer, but it also dissolves into brine. The net effect is that layer~1 saturation is near constant in many realizations.

\begin{figure}[!ht]
\centering
\subfloat[Pressure at O1 in layer~1 of upper aquifer]{\includegraphics[width = 78mm]{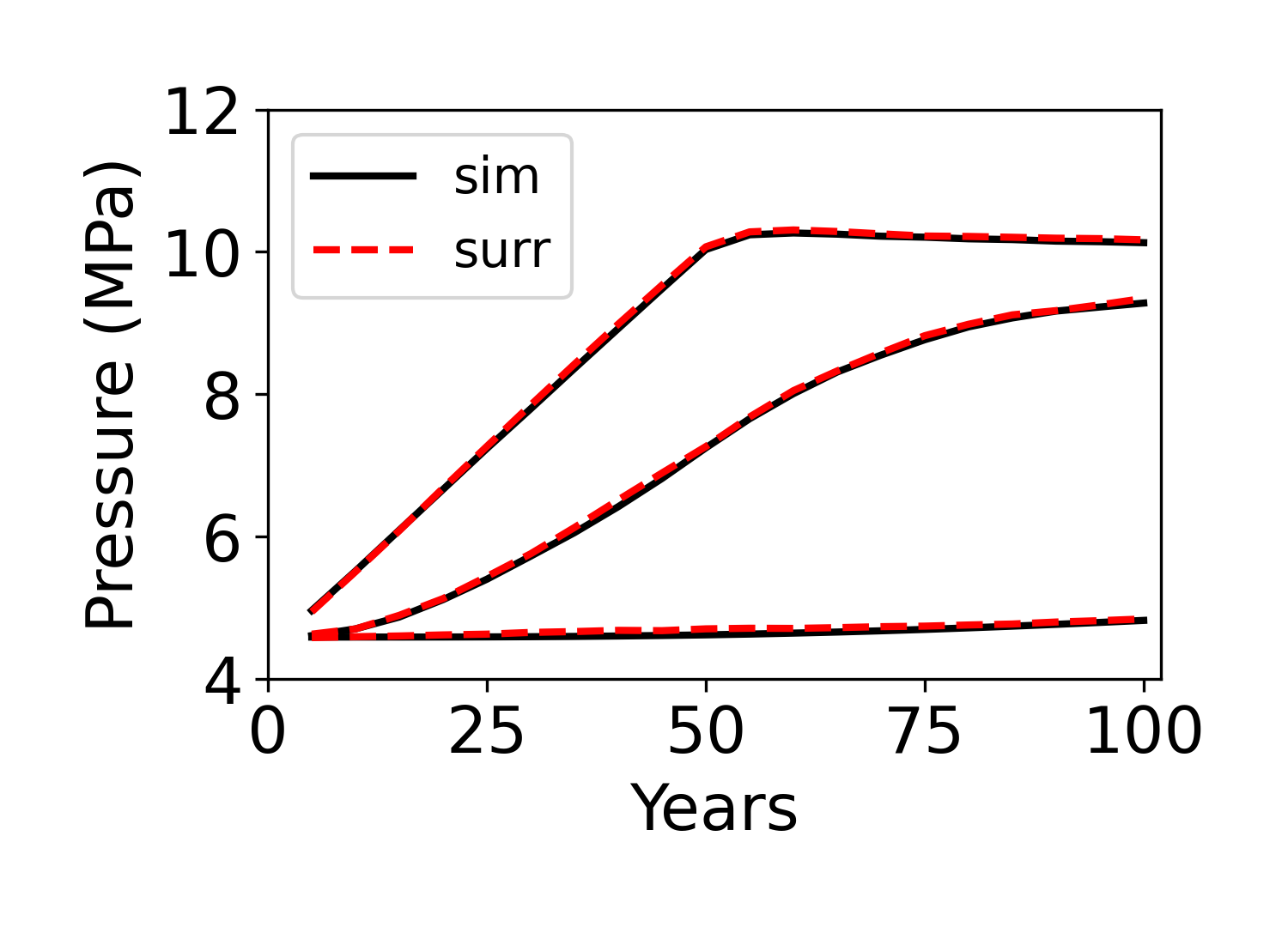}}
\hspace{6mm}
\subfloat[Pressure at O2 in layer~1 of middle aquifer]{\includegraphics[width = 78mm]{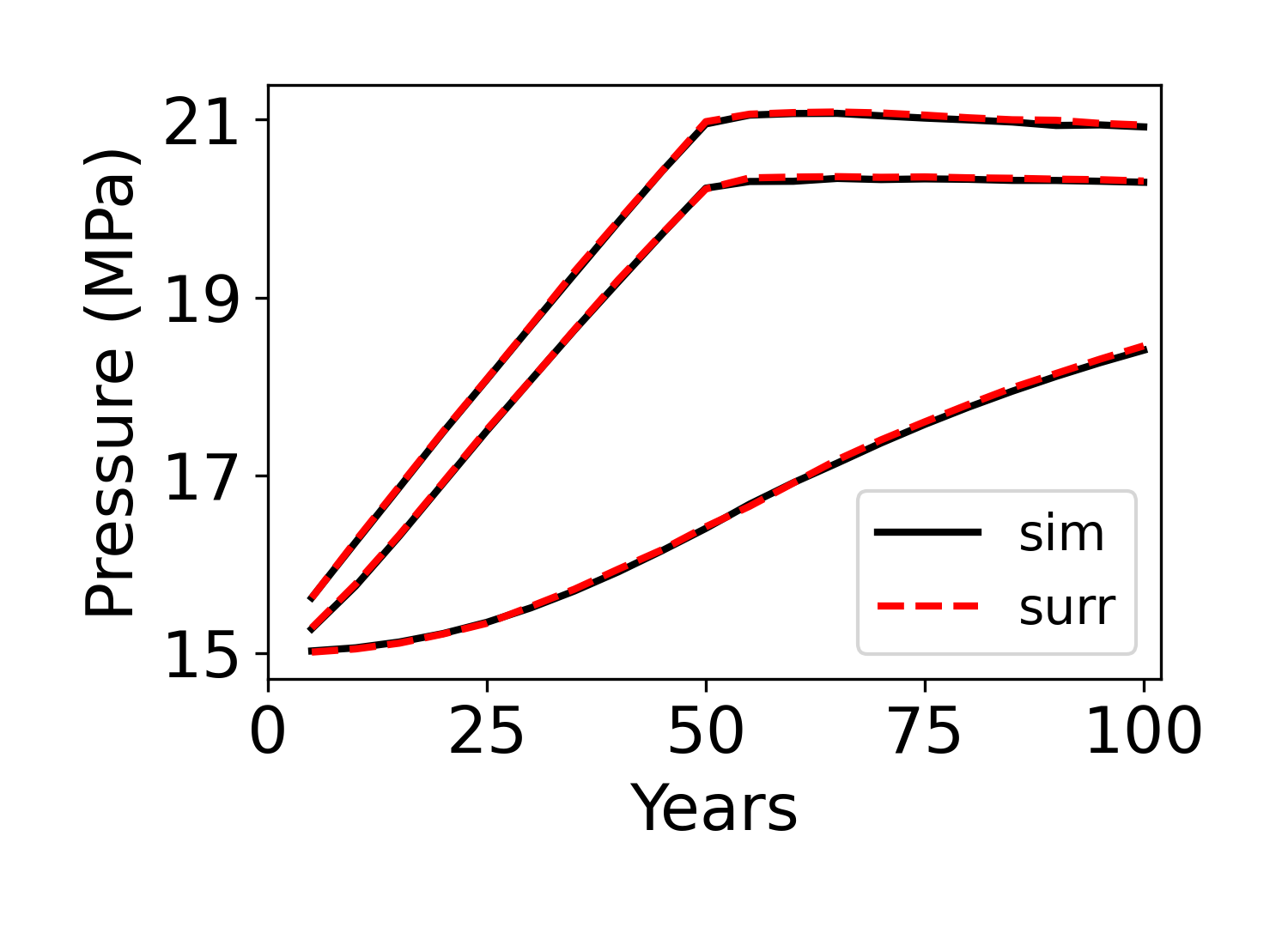}}
\\[1ex]
\subfloat[Pressure at O2 in layer~1 of target aquifer]{\includegraphics[width = 78mm]{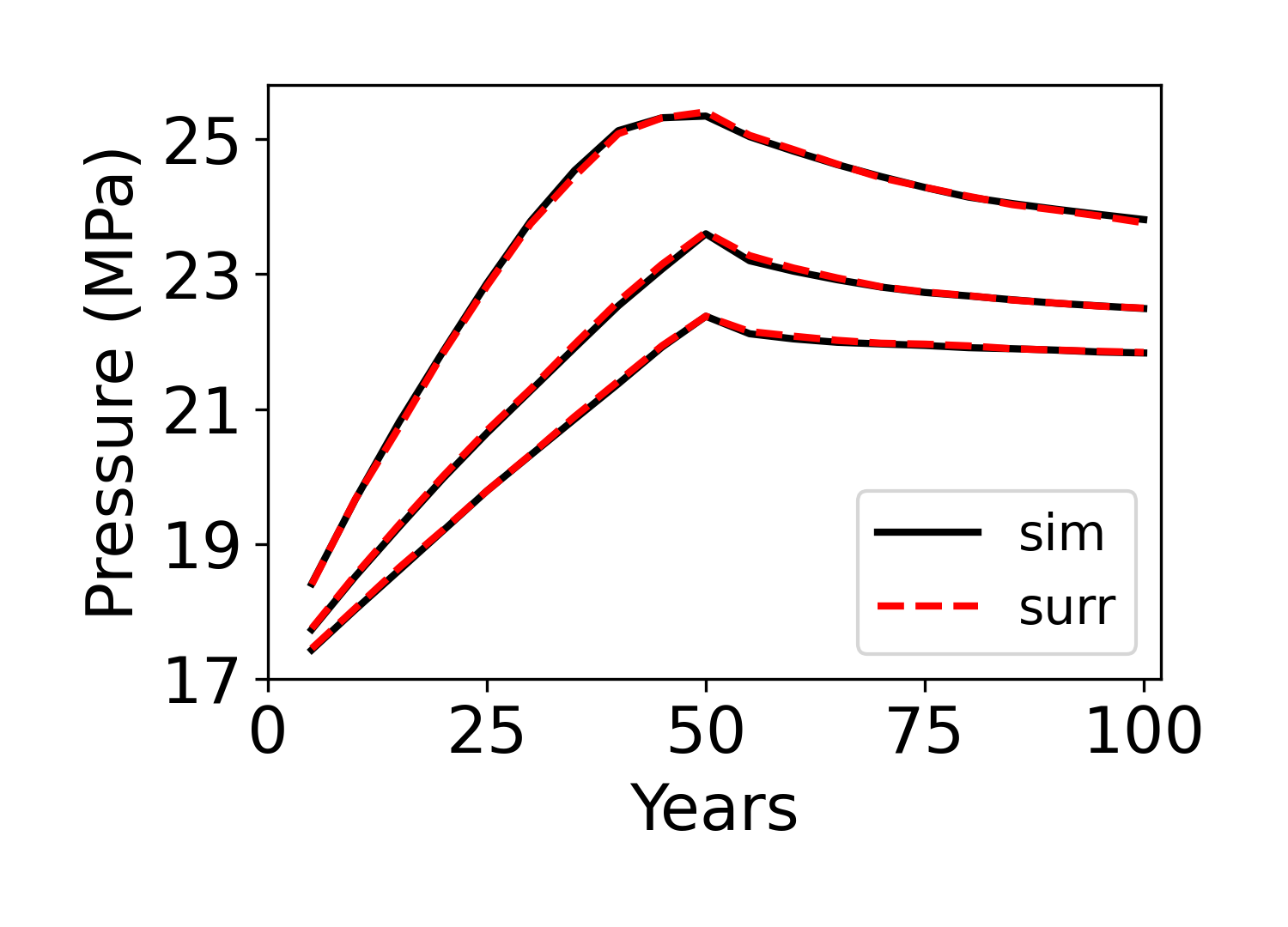}}
\hspace{6mm}
\subfloat[Pressure at O2 in layer~19 of target aquifer]{\includegraphics[width = 78mm]{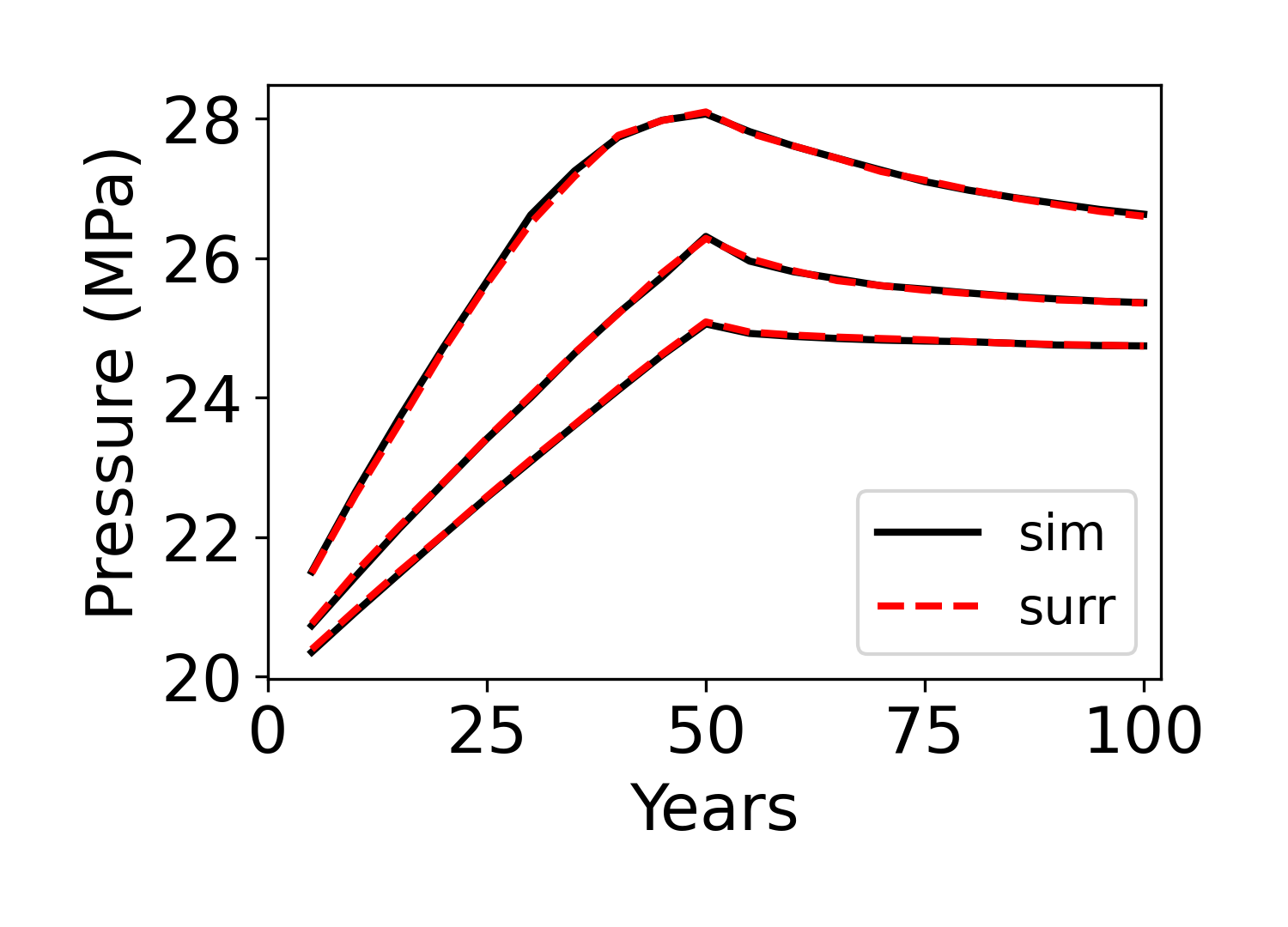}}
\caption{Pressure ensemble statistics from GEOS (black solid curves) and surrogate model (red dashed curves) at three observation locations. The upper, middle and lower curves correspond to P$_{90}$, P$_{50}$ and P$_{10}$ results over the 500 test cases.}
\label{multi_modal_statistics_p}
\end{figure}

Results for pressure are shown in Fig.~\ref{multi_modal_statistics_p}. We see that pressure increases significantly with depth, as expected. The P$_{10}$ result in the upper aquifer (Fig.~\ref{multi_modal_statistics_p}a) shows essentially constant pressure, reflecting the fact that there is little or no leakage into this aquifer in some cases. The results for the target aquifer display a shift in behavior at 50~years, when injection stops. At this time there is a decrease due to pressure dissipation throughout the system. The P$_{90}$ pressures in the target aquifer peak slightly earlier, consistent with a switch from rate control to BHP control in a fraction of the cases. Importantly, these behaviors, along with the various saturation responses (shown in Fig.~\ref{multi_modal_statistics_s}), are accurately captured by the surrogate. We reiterate that the test cases considered here involve randomly sampled geomodel realizations, relative permeability parameters, and perforation and injection strategies.

\begin{figure}[!ht]
\centering
\subfloat[Total mass of CO$_2$]{\includegraphics[width = 85mm]{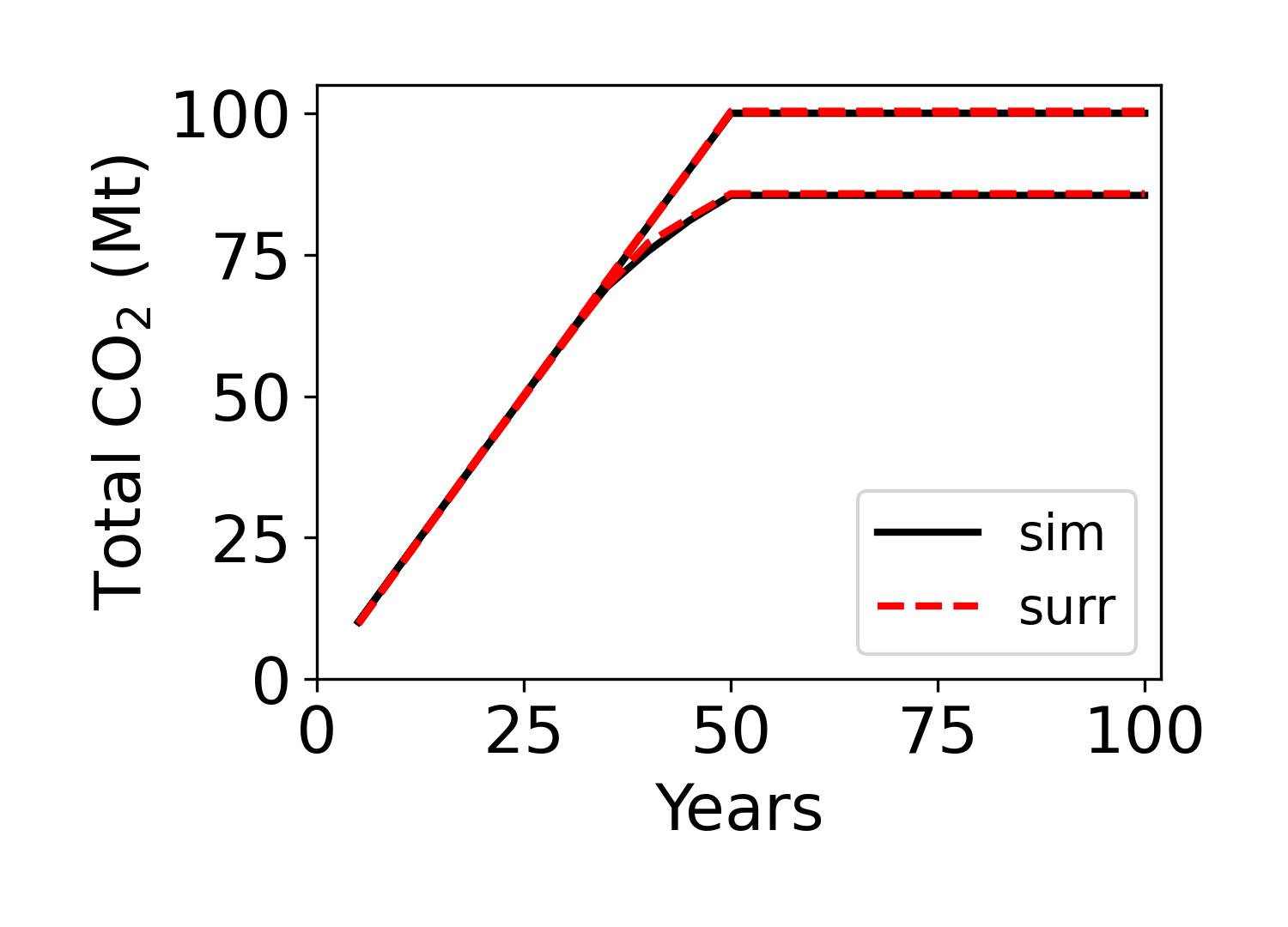}}
\hspace{4mm}
\subfloat[Mobile mass of CO$_2$]{\includegraphics[width = 85mm]{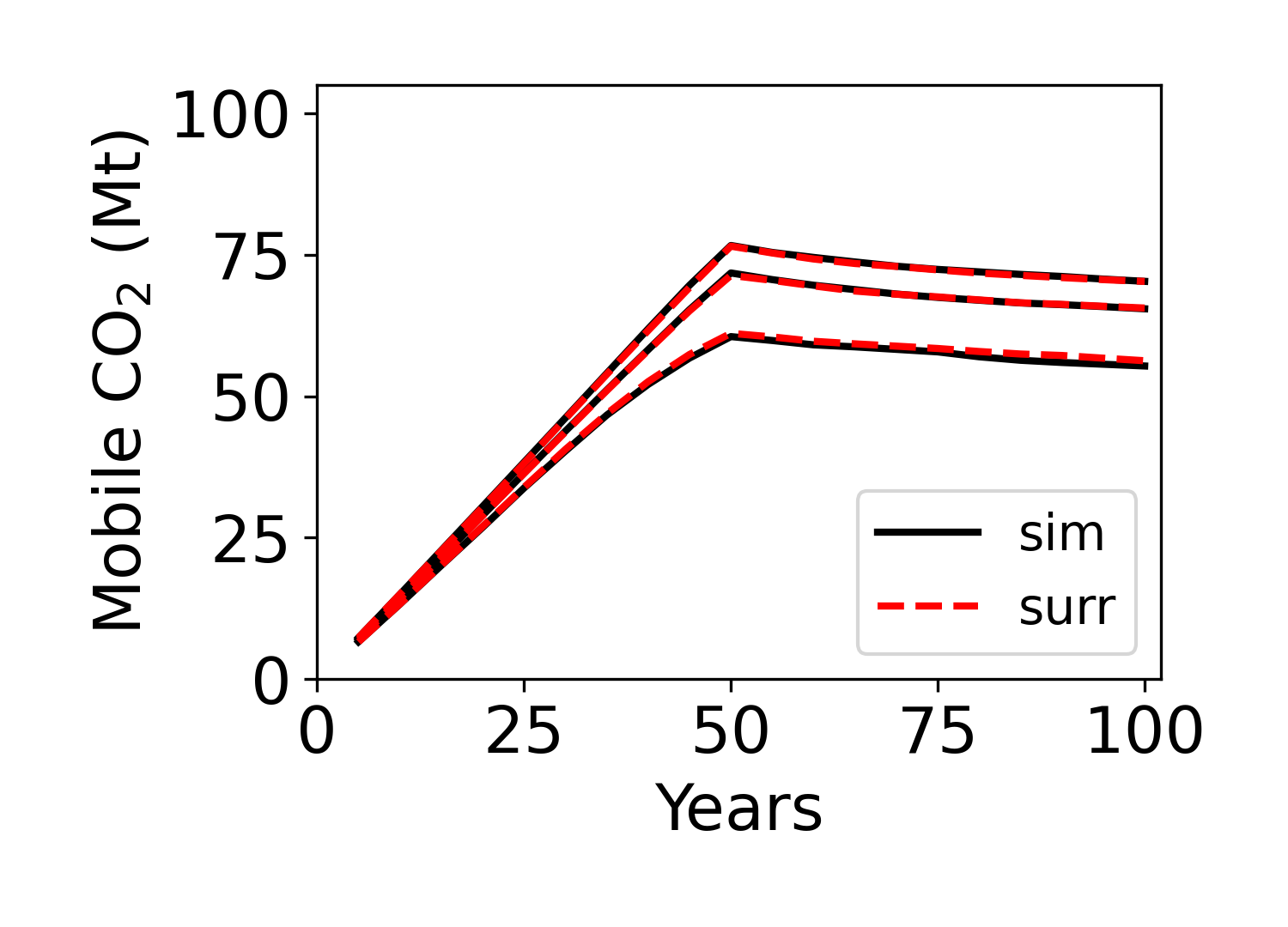}}
\\[1ex]
\caption{Total and mobile CO$_2$ mass ensemble statistics from GEOS (black solid curves) and surrogate model (red dashed curves). The upper, middle and lower curves correspond to P$_{90}$, P$_{50}$ and P$_{10}$ results over the 500 test cases. P$_{90}$ and P$_{50}$ curves overlap in (a).}
\label{multi_modal_statistics_mobile_total_CO2}
\end{figure}

Ensemble statistics for the total injected CO$_2$ and the mass of mobile CO$_2$ in the overall domain are shown in Fig.~\ref{multi_modal_statistics_mobile_total_CO2}. Close agreement between the GEOS results and surrogate model predictions is again observed. For total mass of CO$_2$ injected (Fig.~\ref{multi_modal_statistics_mobile_total_CO2}a), the P$_{90}$ and P$_{50}$ curves essentially overlap, with both reaching the 100~Mt injection target. The P$_{10}$ curve corresponds to a total injected mass of approximately 86~Mt. This is due to one or both injection wells reaching the BHP constraint and then switching from rate control to BHP control, thus reducing the total mass of CO$_2$ injected. The GEOS simulations account for residual and solubility trapping, which both act to immobilize the injected CO$_2$. The mass of mobile CO$_2$, shown in Fig.~\ref{multi_modal_statistics_mobile_total_CO2}b, increases over the injection period and then gradually decreases as CO$_2$ becomes residually trapped or dissolves during the post-injection period.  

\begin{figure}[!ht]
\centering
\subfloat[Total mass of CO$_2$ at 40 years]{\includegraphics[width = 80mm]{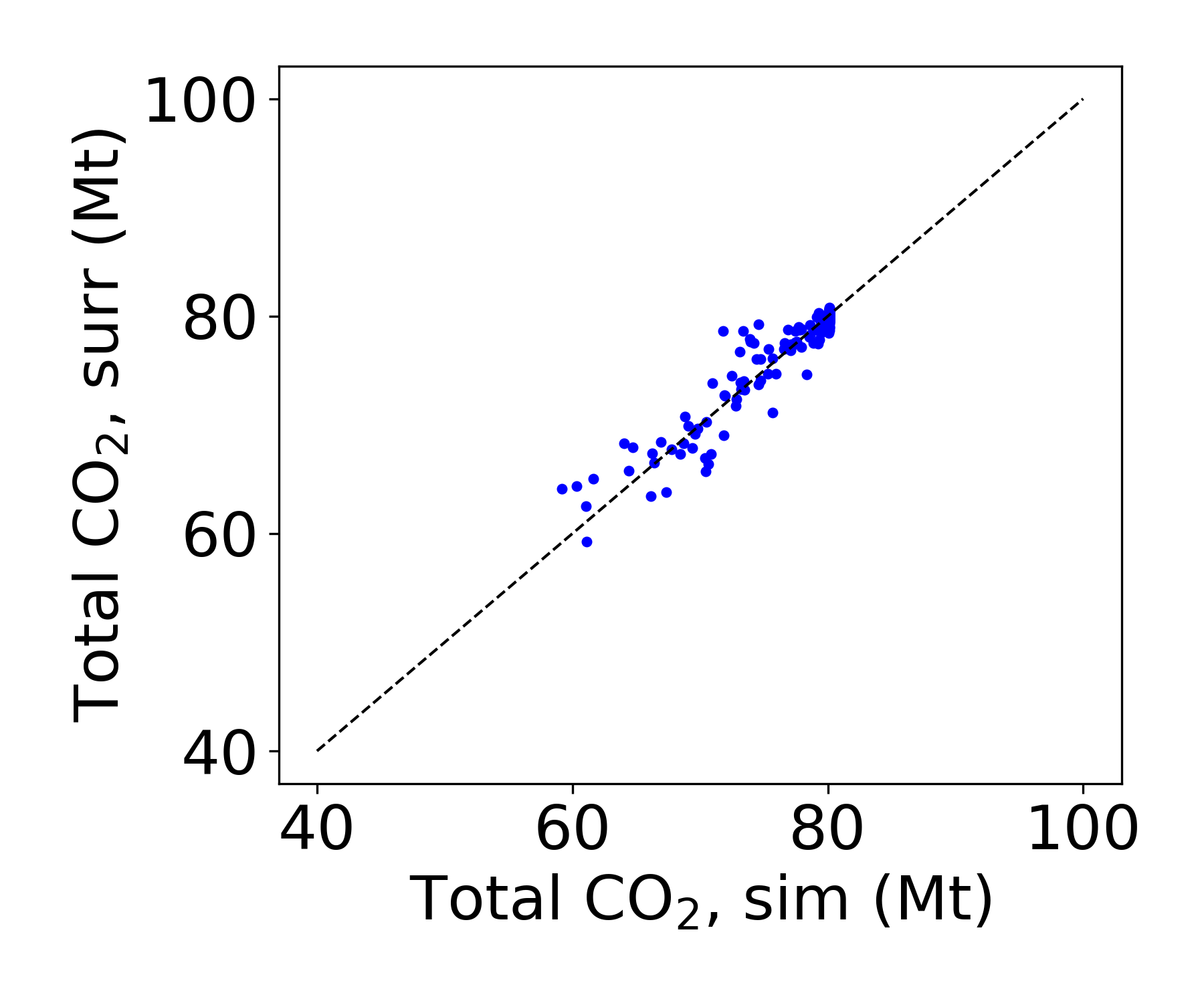}}
\hspace{4mm}
\subfloat[Total mass of CO$_2$ at 50 years]{\includegraphics[width = 80mm]{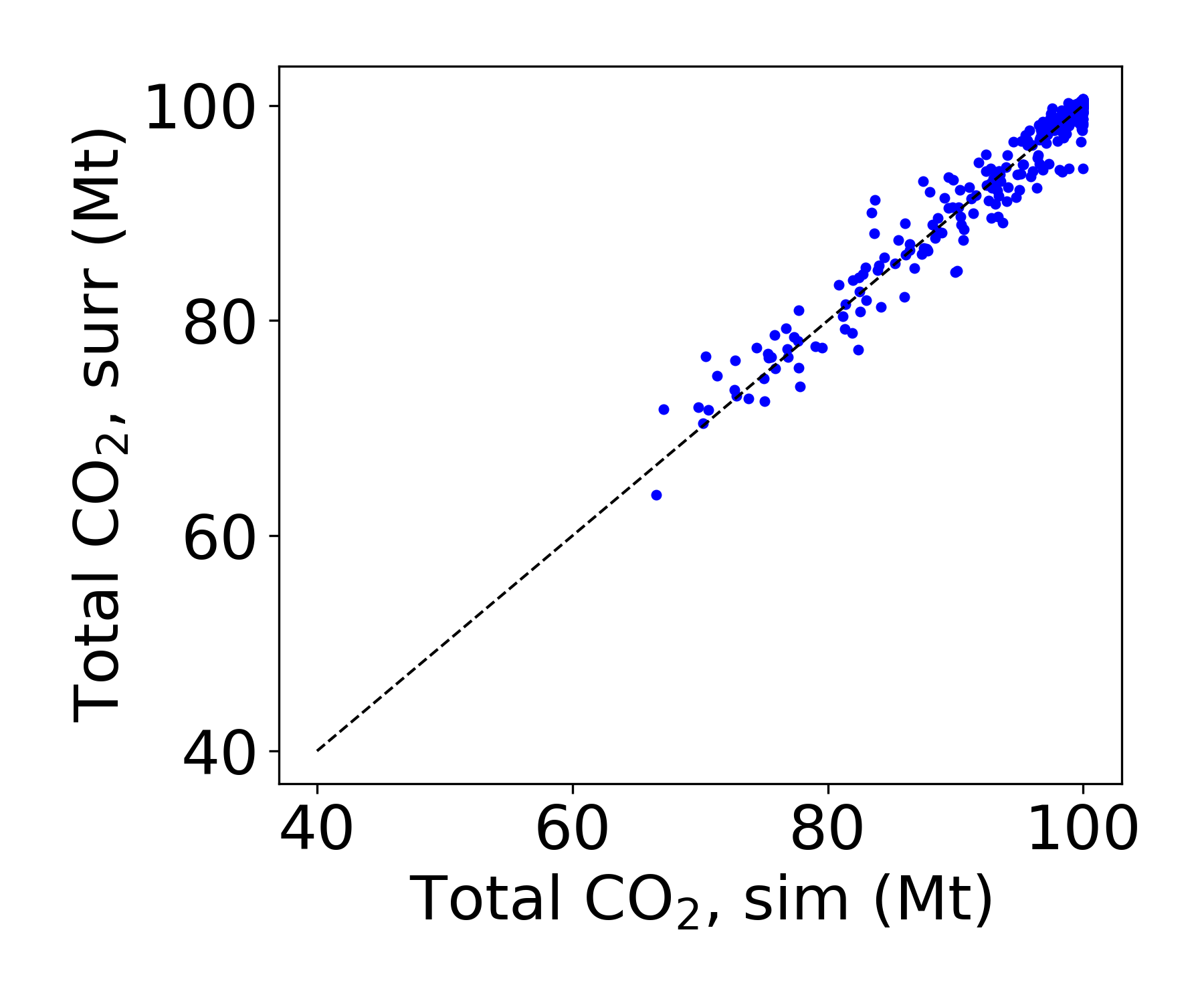}}
\\[1ex]
\subfloat[Mobile mass of CO$_2$ at 50 years]{\includegraphics[width = 80mm]{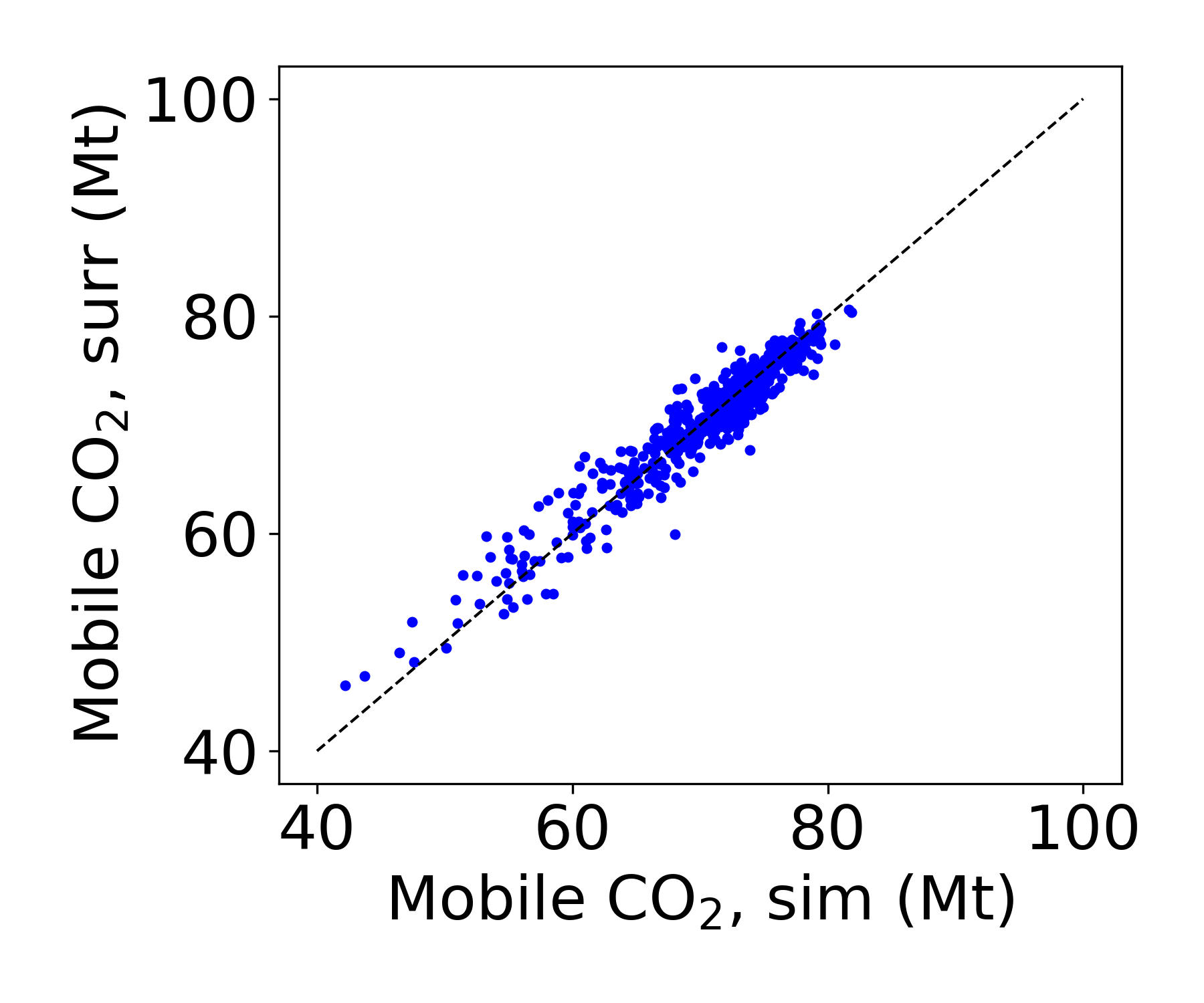}}
\hspace{4mm}
\subfloat[Mobile mass of CO$_2$ at 100 years]{\includegraphics[width = 80mm]{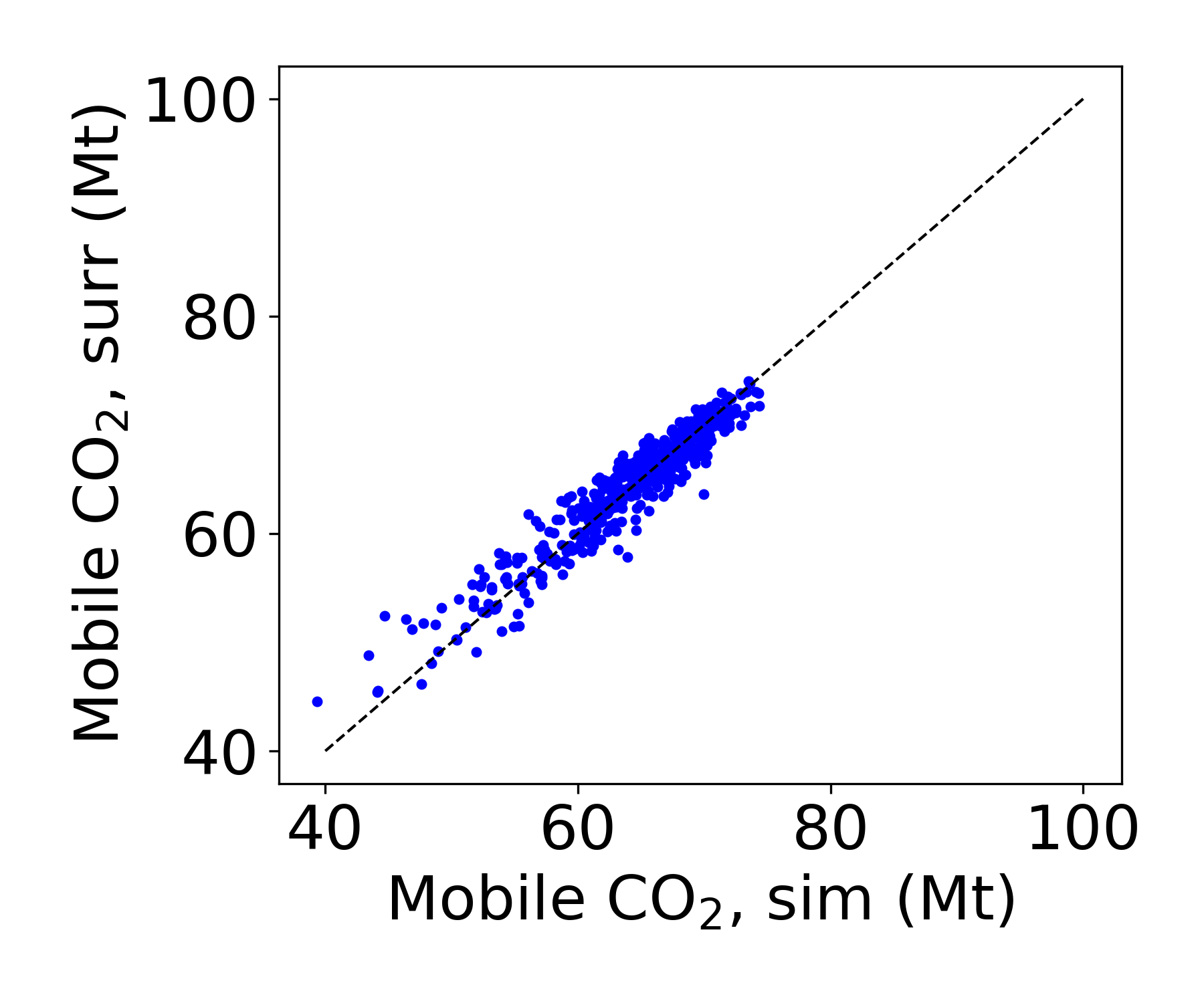}}
\caption{Total injected CO$_2$ mass cross plots at 40 and 50~years in (a) and (b), and mass of mobile CO$_2$ cross plots at 50 and 100~years in (c) and (d).}
\label{multi_modal_statistics_mobile_total}
\end{figure}

Cross plots for injected mass and mobile mass of CO$_2$, at particular times, are presented in Fig.~\ref{multi_modal_statistics_mobile_total}. In these plots, each point corresponds to one of the 500 test cases, and the dashed line indicates perfect agreement. The total injected CO$_2$ mass results at 40~years (Fig.~\ref{multi_modal_statistics_mobile_total}a) display a dense cluster of points at 80~Mt, corresponding to test cases in which both wells have remained on rate control up to this time. The points at lower values correspond to cases in which the BHP constraint has become active. Similar behavior is evident at 50~years (Fig.~\ref{multi_modal_statistics_mobile_total}b), where a large fraction of the cases achieve the full 100~Mt injection. The mobile CO$_2$ mass results at 50 and 100~years (Fig.~\ref{multi_modal_statistics_mobile_total}c and d) cluster around the 45$^\circ$ line, with slight scatter appearing for cases with smaller amounts of mobile CO$_2$. These results demonstrate that the surrogate model captures, with reasonable accuracy, the impact of switching from rate control to BHP control on a realization-by-realization basis. This is an important though complicated effect that depends on the geomodel, relative permeability curves, and perforation and injection strategy.

\begin{figure}[!ht]
\centering
\subfloat[I1 footprint ensemble statistics]{\includegraphics[width=80mm]{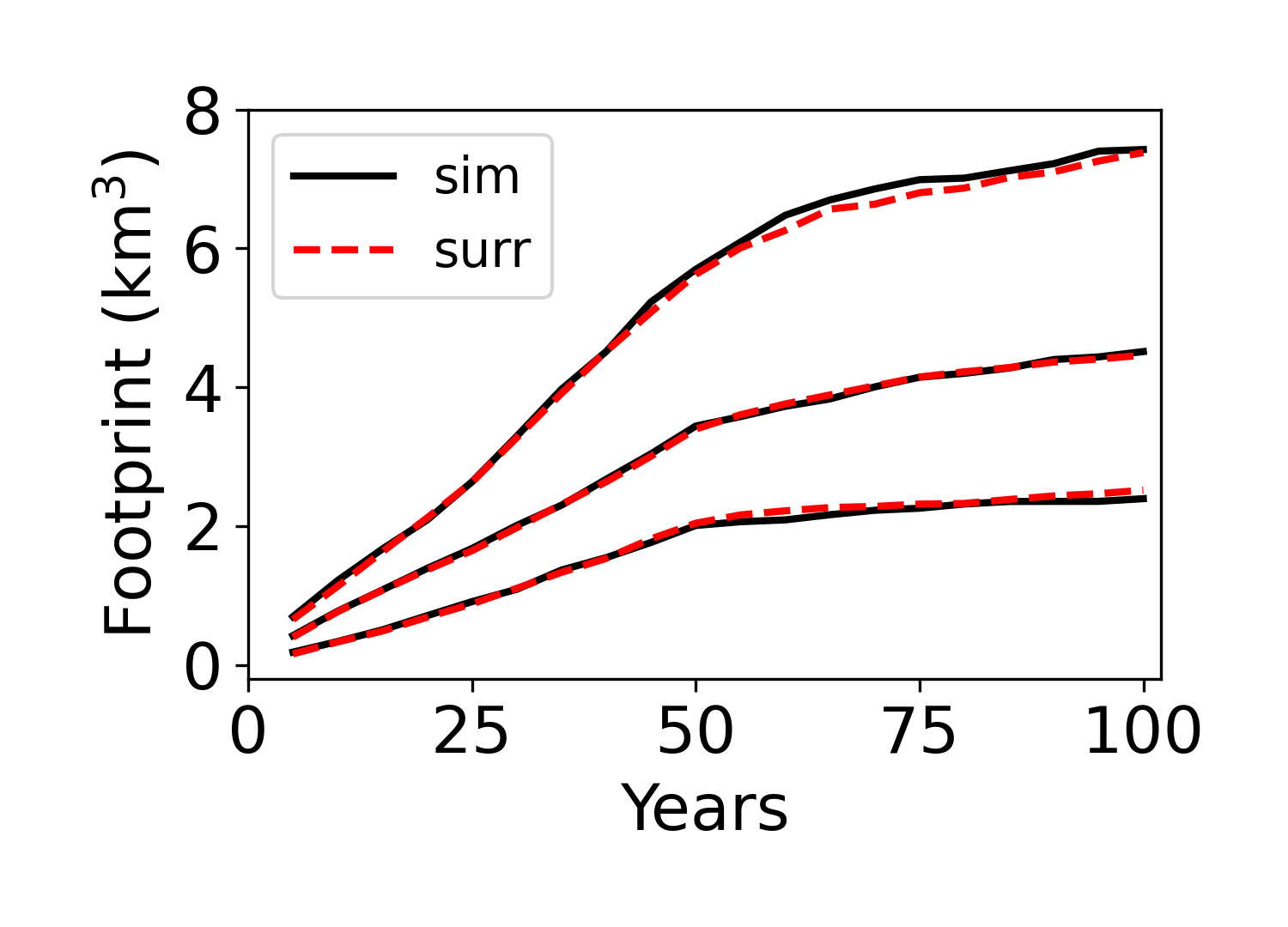}}
\hspace{5mm}
\subfloat[I2 footprint ensemble statistics]{\includegraphics[width=80mm]{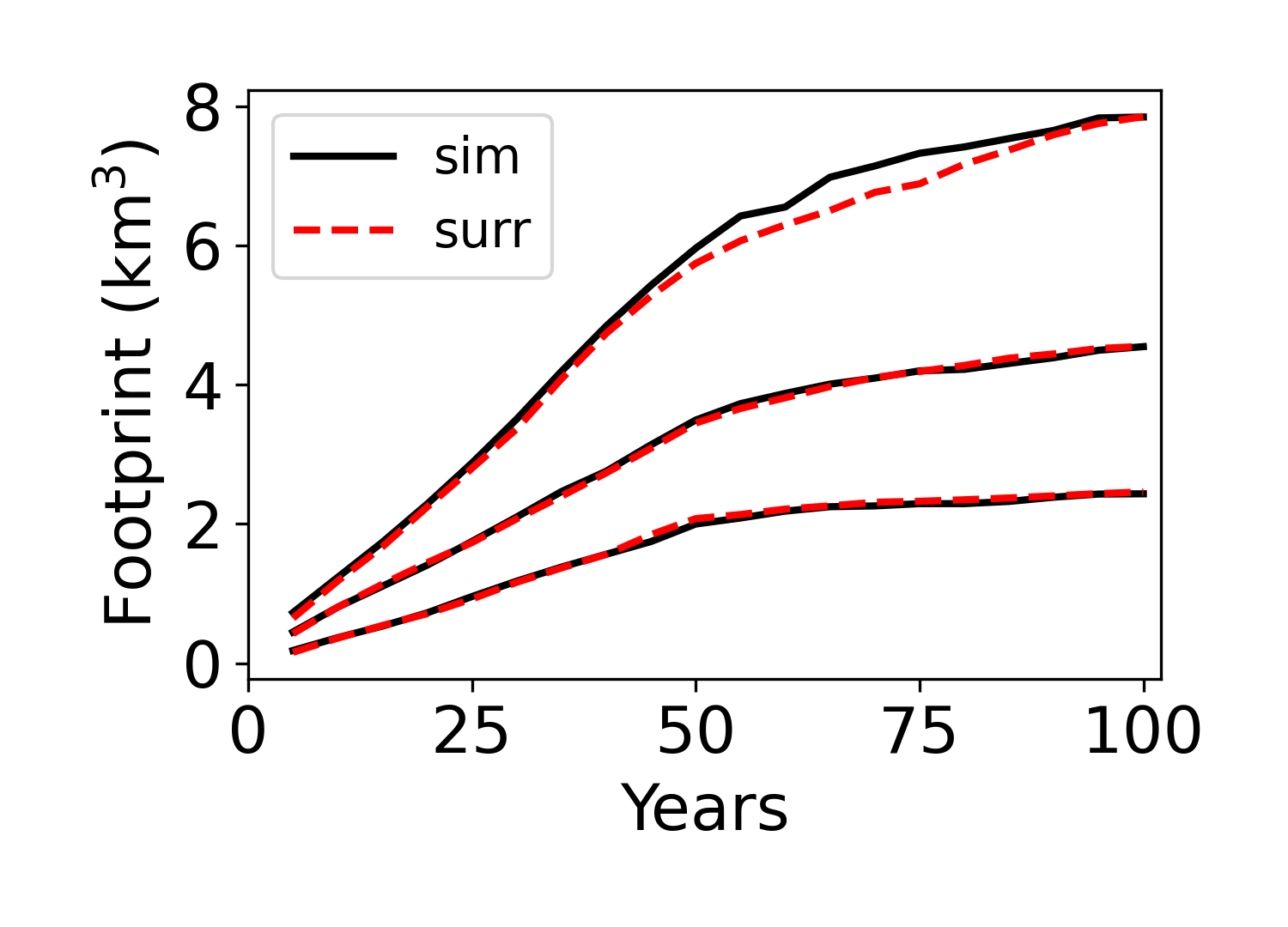}}
\\[1ex]
\subfloat[Saturation footprint at 50~years]{\includegraphics[width=82mm]{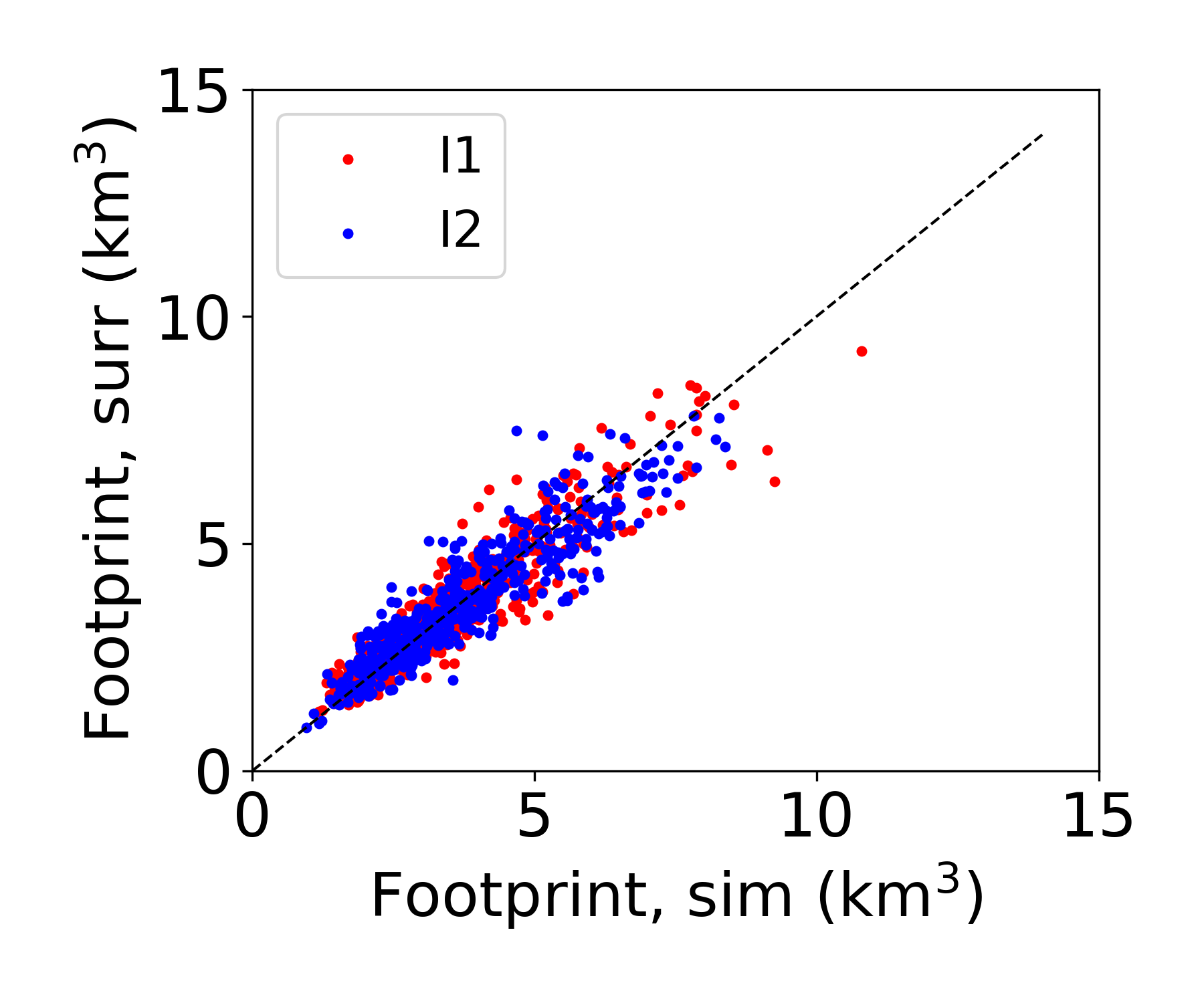}}
\hspace{5mm}
\subfloat[Saturation footprint at 100~years]{\includegraphics[width=82mm]{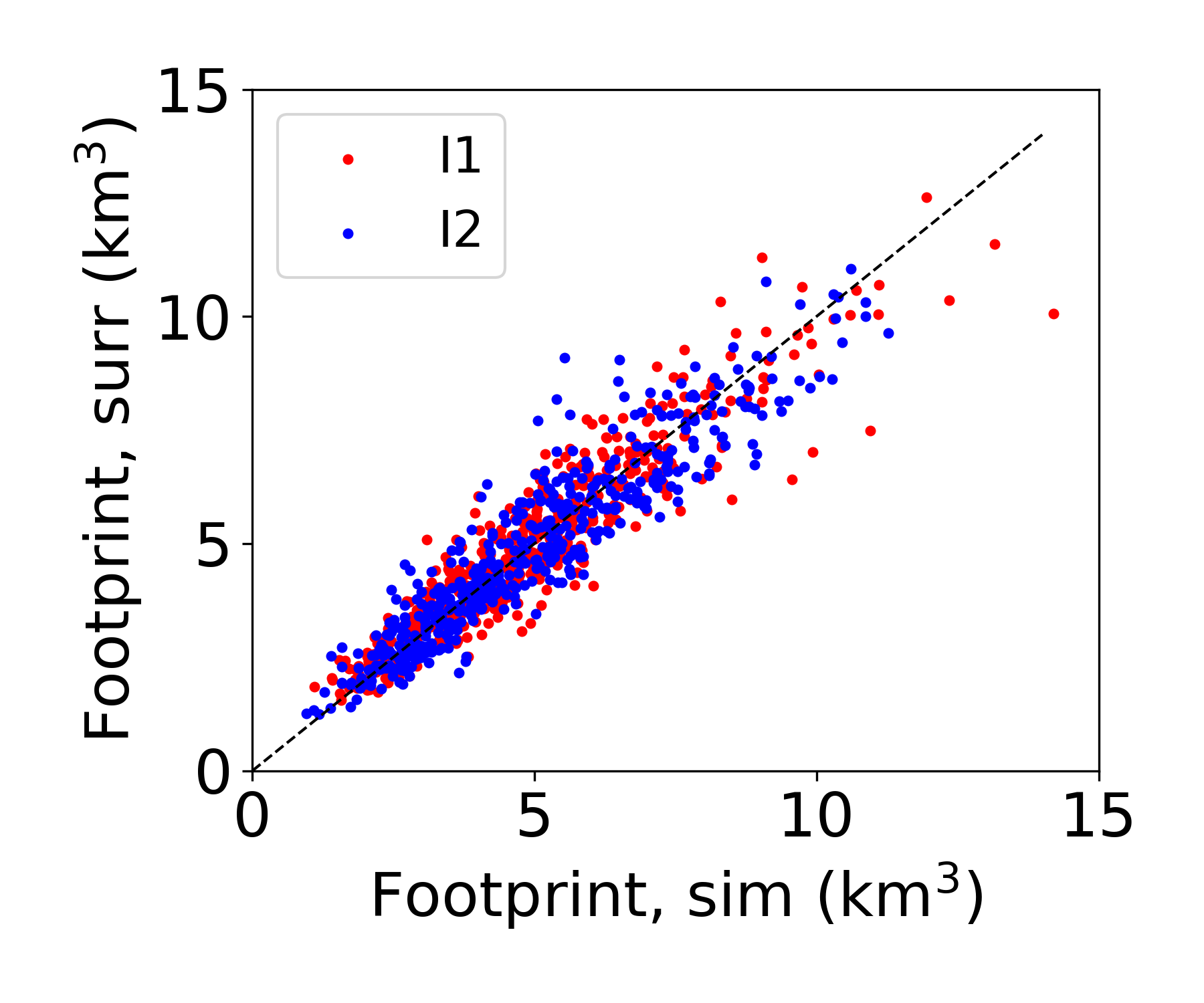}}
\\[1ex]
\caption{Saturation footprint associated with the two injection wells. P$_{10}$, P$_{50}$ and P$_{90}$ ensemble statistics from GEOS (black solid curves) and the surrogate model (red dashed curves) for injection wells I1 and I2 in (a) and (b), along with cross plots of saturation footprints at 50~years and 100~years for the two injectors in (c) and (d).}
\label{multi_modal_statistics_footprint}
\end{figure}

Results for the saturation footprints associated with the two injection wells (in the target aquifer) are shown in Fig.~\ref{multi_modal_statistics_footprint}. Close statistical correspondence between the surrogate model and the GEOS simulations, in terms of the P$_{10}$, P$_{50}$, and P$_{90}$ responses, is evident in Fig.~\ref{multi_modal_statistics_footprint}a and b. Cross plots for predicted versus simulated footprints at 50 and 100~years, presented in Fig.~\ref{multi_modal_statistics_footprint}c and~d, show that the predicted saturation footprints generally cluster around the 45$^\circ$ line, though some scatter is observed, particularly at 100~years. This may be due in part to the large (order of magnitude) variation in the size of the footprints for the two injection wells.

Finally, we evaluate surrogate model performance under different perforation and injection strategies for a particular geomodel realization. For this assessment we consider realization~1, characterized by the metaparameters given in Table~\ref{multi-modal_metaparameters}. The perforation and injection strategies are shown in  Fig.~\ref{variable_injection}. Note that these three strategies represent a range of cases and include some short-duration settings.

\begin{figure}[!ht]
\centering
\subfloat[Total mass of CO$_2$]{\includegraphics[width=85mm]{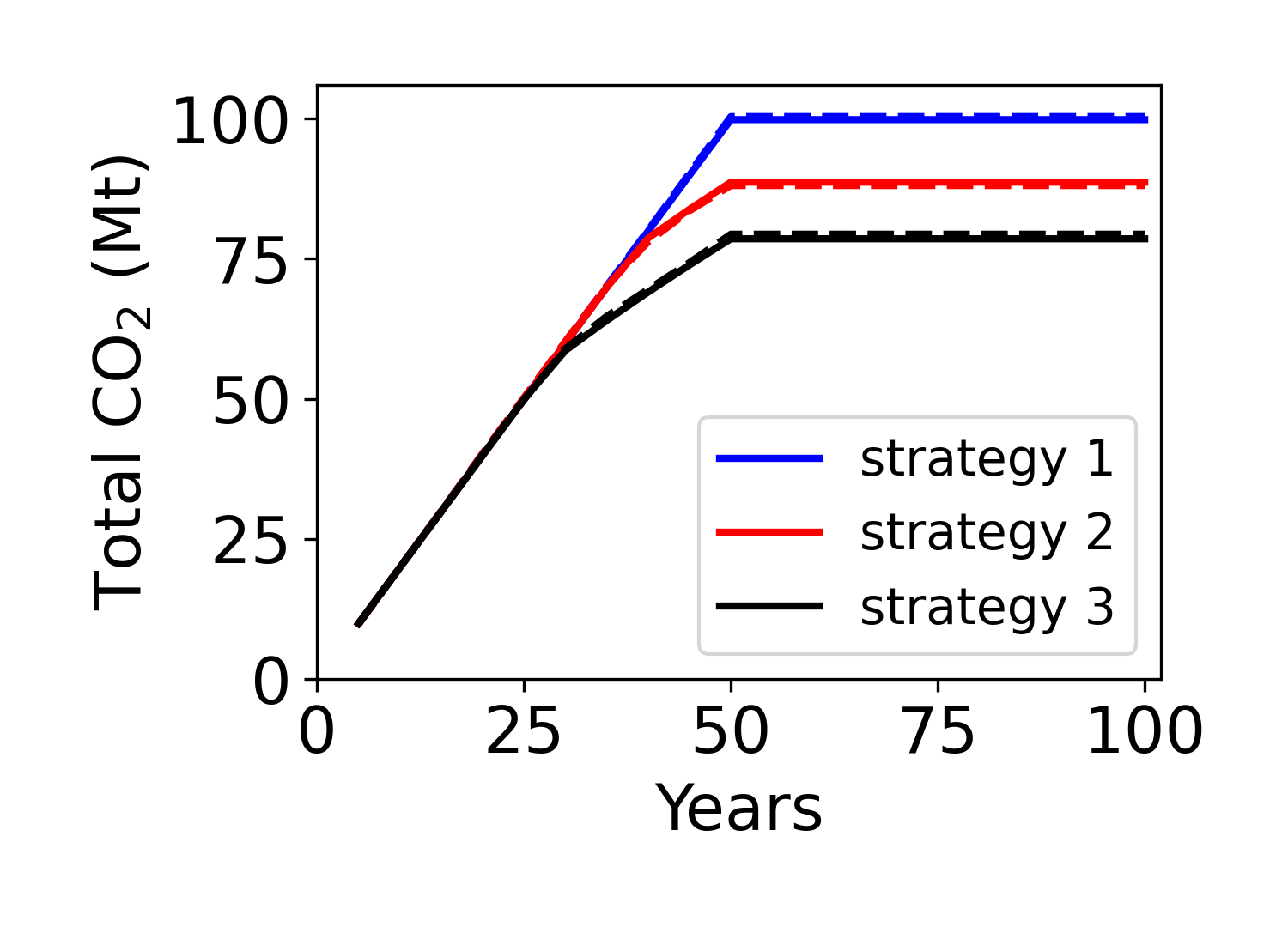}}
\hspace{5mm}
\subfloat[Mobile mass of CO$_2$]{\includegraphics[width=85mm]{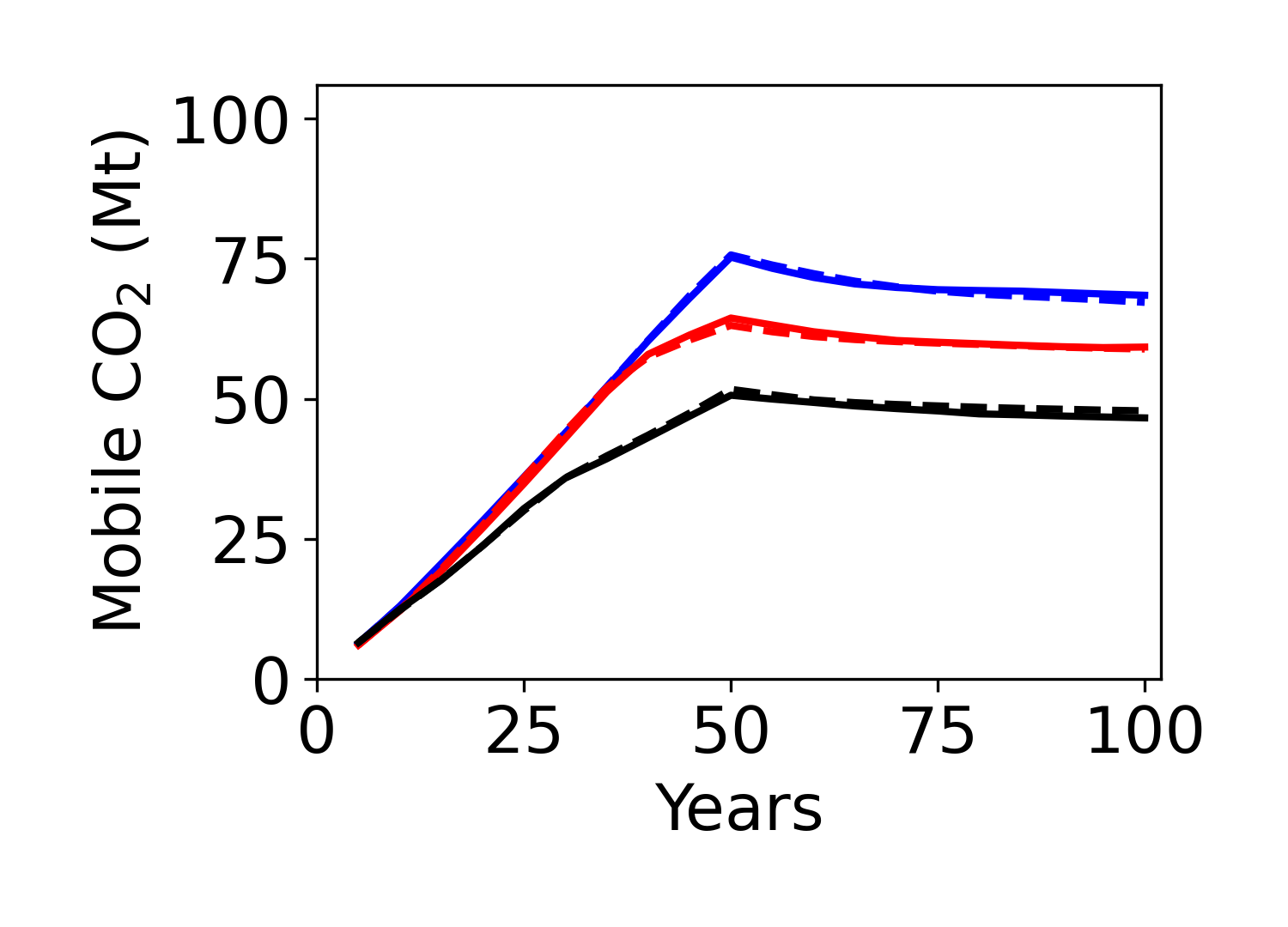}}
\\[1ex]
\caption{Total mass of injected and mobile CO$_2$ for geomodel realization~1 from GEOS (solid curves) and surrogate model (dashed curves) for the three variable perforation strategies shown in Fig.~\ref{variable_injection}. For each case, the geomodel and relative permeability parameters are fixed, and only the perforation and injection strategy is varied.}
\label{CO2_prediction}
\end{figure}

Comparisons between the surrogate model and GEOS results for the three strategies, for total mass of injected CO$_2$ and mass of mobile CO$_2$ in the overall domain, are shown in Fig.~\ref{CO2_prediction}. It is evident that the three strategies lead to different injection and mobile CO$_2$ results and that the surrogate model remains accurate for each of the strategies. Strategy~1 injects the target mass of 100~Mt, while strategies~2 and~3 inject only about 88.7~Mt and 78.6~Mt, respectively. The amount of mobile CO$_2$ at 100~years for the three strategies is about 68.5~Mt, 59.2~Mt and 46.6~Mt. These results reiterate that, for a given realization, different injection and perforation strategies can lead to substantial variation in the mass of injected and mobile CO$_2$. These differences indicate the potential to optimize the perforation and injection strategy with the goals of, e.g., injecting the full 100~Mt target over the 50-year injection period while minimizing the fraction of this CO$_2$ that is mobile (and thus able to migrate or leak).

\section{Data Assimilation using Surrogate Model} 
\label{Data Assimilation}
In this section, the multimodal autoregressive transformer surrogate model is used in a hierarchical MCMC-based data assimilation procedure to reduce uncertainty in the metaparameters, associated geomodels, and key quantities of interest for a synthetic true model (geomodel realization~1). Because the surrogate model treats the perforation and injection strategy as an input, data assimilation can be readily performed under different operational settings. This is a key difference between this work and most previous studies. Here we consider the three perforation and injection strategies shown in Fig.~\ref{variable_injection}, which allows us to assess the impact of the operational strategy on data assimilation results.

\subsection{Data assimilation problem setup}
\label{sec:da_setup}

The data used in our assessments derive from a synthetic ``true'' model, denoted $\textbf{m}_{\mathrm{true}}$. This model corresponds to geomodel realization~1, with metaparameters given in Table~\ref{multi-modal_metaparameters}. In this model, fault~1 is of very low permeability ($k_{f1} = 0.02$~md) while fault~2 is relatively permeable ($k_{f2} = 88.7$~md). The CO$_2$-brine relative permeability (with hysteresis) and capillary pressure curves for this model are shown in Fig.~\ref{fig:multi_modal_relper_pc}. The true model is simulated in GEOS under each of the three perforation and injection strategies, and the simulation results at observation wells O1 and O2 (locations shown in Fig.~\ref{multi-modal_well_location}) provide the true data. Measurement error, in the form of random noise, is then added to these data to give the observed data used in data assimilation. 

The GEOS simulation results for the 3D CO$_2$ saturation field at 100~years for $\textbf{m}_{\mathrm{true}}$, under operational strategy~1, is shown in Fig.~\ref{S:saturation_100_years}a. Some leakage of CO$_2$ into the upper aquifer is apparent. Consistent with the quantities predicted by the surrogate model, the observed data comprise saturation in four layers of the target aquifer, pressure in two layers of the target aquifer, and pressure in the top layers of the middle and upper aquifers at the two monitoring wells. These data are collected every five years over the first 25~years after the start of injection, resulting in a total of $n_m = 80$ measurements (40~saturation and 40~pressure measurements) for each strategy.

Error associated with the precision of the measurement devices and with data interpretation impacts the observed data. Consistent with our previous studies, the standard deviation of this error is set to 0.02 for all saturation measurements and 0.095~MPa for pressure measurements. The observed data with measurement error, denoted $\textbf{d}_{\mathrm{obs}} \in \mathbb{R}^{n_m}$, are expressed as
\begin{equation} \label{eq:dobs}
\textbf{d}_{\mathrm{obs}} = f(\textbf{m}_{\mathrm{true}}) + \bm{\epsilon},
\end{equation}
\noindent where $f(\textbf{m}_{\mathrm{true}}) \in \mathbb{R}^{n_m}$ denotes the GEOS simulation results at the observation locations for the true model, and $\bm{\epsilon} \in \mathbb{R}^{n_m}$ is the measurement error, sampled from $\mathcal{N}(\textbf{0}, C_{\mathrm{D}})$. Here $C_{\mathrm{D}} \in \mathbb{R}^{n_m \times n_m}$ is the diagonal covariance matrix constructed from the measurement error standard deviations.

Error also arises from the use of the surrogate model in place of the high-fidelity flow simulator. Following the treatment in \citet{han2025accelerated}, this error is expressed in terms of the covariance of the surrogate model error, $C_{\mathrm{surr}} \in \mathbb{R}^{n_m \times n_m}$, which is computed from the differences in GEOS and surrogate model predictions for the test set. The total error covariance, which is the covariance used in data assimilation, is then given by $C_{\mathrm{tot}} = C_{\mathrm{D}} + C_{\mathrm{surr}}$. The $C_{\mathrm{surr}}$ and $C_{\mathrm{tot}}$ matrices are nondiagonal, as required to capture spatial and temporal correlations. More specialized treatments for representing model error are discussed later.

\subsection{Hierarchical MCMC procedure}
\label{sec:da_mcmc}

The hierarchical MCMC sampling method used in this study is based on the noncentered preconditioned Crank--Nicolson within Gibbs algorithm introduced by \citet{chen2018dimension} and applied in geological carbon storage problems by \citet{han2023surrogate, han2025accelerated} and \citet{han2026recurrent}. The uncertain variables include the metaparameters $\boldsymbol{\uptheta}_{\mathrm{meta}} \in \mathbb{R}^{10}$ defined in Eq.~\ref{multi-modal_faulted_metaparameters} and the PCA latent variables $\bm{\xi} \in \mathbb{R}^{n_{\xi}}$, which characterize the geological realization of the target aquifer. In this study, $n_{\xi} = 850$, which acts to retain approximately 95\% of the energy in the 1000 original SGeMS realizations used to construct the PCA basis matrix. See~\citet{han2026recurrent} for more details on the construction and use of the PCA representation for the target aquifer. 

From Bayes' theorem, the posterior probability density function for the metaparameters and PCA latent variables, conditioned to the observed data, is given by
\begin{equation} \label{eq:bayes}
p(\boldsymbol{\uptheta}_{\mathrm{meta}}, \bm{\xi} \mid \textbf{d}_{\mathrm{obs}}) = \frac{p(\boldsymbol{\uptheta}_{\mathrm{meta}}, \bm{\xi}) \, p(\textbf{d}_{\mathrm{obs}} \mid \boldsymbol{\uptheta}_{\mathrm{meta}}, \bm{\xi})}{p(\textbf{d}_{\mathrm{obs}})},
\end{equation}
\noindent where $p(\boldsymbol{\uptheta}_{\mathrm{meta}}, \bm{\xi})$ is the prior probability density function and $p(\textbf{d}_{\mathrm{obs}})$ is a normalization constant. The likelihood function $p(\textbf{d}_{\mathrm{obs}} \mid \boldsymbol{\uptheta}_{\mathrm{meta}}, \bm{\xi})$ is computed as
\begin{equation} \label{eq:likelihood}
p(\textbf{d}_{\mathrm{obs}} \mid \boldsymbol{\uptheta}_{\mathrm{meta}}, \bm{\xi}) = c \exp\left(-\frac{1}{2} \left(\textbf{d}_{\mathrm{obs}} - \hat{f}\left(\textbf{m}\left(\boldsymbol{\uptheta}_{\mathrm{meta}}, \bm{\xi}\right)\right)\right)^{T} C_{\mathrm{tot}}^{-1} \left(\textbf{d}_{\mathrm{obs}} - \hat{f}\left(\textbf{m}\left(\boldsymbol{\uptheta}_{\mathrm{meta}}, \bm{\xi}\right)\right)\right)\right).
\end{equation}
\noindent Here $c$ is a normalization constant, $\textbf{m}(\boldsymbol{\uptheta}_{\mathrm{meta}}, \bm{\xi})$ is the geomodel constructed from a set of metaparameters and PCA latent variables, and $\hat{f}(\textbf{m}(\boldsymbol{\uptheta}_{\mathrm{meta}}, \bm{\xi})) \in \mathbb{R}^{n_m}$ is the surrogate model prediction for saturation and pressure at the observation locations. The matrix $C_{\mathrm{tot}}$ is the total error covariance, described above. 

The MCMC sampling procedure involves simulating $m$ Markov chains in parallel (here $m=20$). For each chain, the initial metaparameters are sampled from their prior distributions, and each component of the initial set of PCA latent variables is sampled independently from $\mathcal{N}(0, 1)$. The Metropolis-within-Gibbs approach requires sampling metaparameters and latent variables independently. At iteration $k+1$ for chain $j$ ($j = 1, \ldots, m$), a new set of PCA latent variables $\bm{\xi}_j^{'}$ is proposed through application of
\begin{equation} \label{eq:pcn}
\bm{\xi}_j^{'} = \sqrt{1 - \beta^{2}} \, \bm{\xi}_j^{k} + \beta \bm{\eta},
\end{equation}
\noindent where $\bm{\xi}_j^{k}$ is the sample at iteration $k$, $\beta$ is a coefficient controlling the magnitude of the update (here $\beta = $~0.35, which provides an appropriate acceptance rate), and $\bm{\eta} \in \mathbb{R}^{n_{\xi}}$ is a random vector with each component sampled independently from $\mathcal{N}(0, 1)$. A new set of metaparameters is then proposed by sampling from a multivariate Gaussian distribution centered at the current sample, with the proposal standard deviation for each metaparameter set to 1/40 of its prior range. 

The proposed PCA latent variables and metaparameters are accepted or rejected (independently) based on the Metropolis--Hastings criteria~\citep{hastings1970monte}. Please see \citet{han2023surrogate} for implementation details and \citet{han2026recurrent} for our treatment of parallel chains. Convergence is evaluated using the Gelman--Rubin potential scale reduction factor~\citep{gelman1992inference}, which compares the variance within individual chains with that across the $m$ chains. Values close to one indicate that the chains have converged to very similar distributions. We require values below 1.05 for all 10~metaparameters. This is a tighter threshold than the commonly used value of 1.1.

\subsection{Data assimilation results}
\label{sec:da_results}

In terms of timings, each surrogate model evaluation for saturation, pressure, mobile CO$_2$ mass, and saturation footprint (individually) require about 0.064~seconds on a single Nvidia A100 GPU. The evaluation of total injected CO$_2$, which involves only 10~time steps, requires about 0.035~seconds. Within the MCMC procedure, the saturation and pressure networks provide the predictions required for the likelihood computation in Eq.~\ref{eq:likelihood}. Surrogate model evaluations for the 20~chains at each iteration are performed in parallel, in a single batch. The batched evaluation for saturation or pressure (individually) requires about 0.072~seconds, only slightly more than a single evaluation, corresponding to a speedup of about $17.6\times$ relative to sequential evaluation. The networks for the saturation footprints and total injected and mobile CO$_2$ mass are applied after the data assimilation procedure, for the accepted post-burn-in samples, to generate corresponding posterior predictions. 

The MCMC procedure is performed separately for each of the three perforation and injection strategies. The three runs converge after 34,400, 50,700, and 79,500 iterations, respectively. These correspond to 1,376,000 (34,400 $\times$ 20 chains $\times$ 2 evaluations per iteration for $\boldsymbol{\uptheta}_{\mathrm{meta}}$ and $\bm{\xi}$), 2,028,000, and 3,180,000 function evaluations. After the burn-in period, 134,458, 200,220, and 334,932 posterior samples ($\boldsymbol{\uptheta}_{\mathrm{meta}}$ and $\bm{\xi}$) are collected. The total elapsed times on a single NVIDIA A100 GPU for the three data assimilation runs are approximately 4.3, 5.9, and 10.4~hours, respectively. These include PCA-based geomodel construction, surrogate model evaluation, hierarchical MCMC execution, likelihood computation, and convergence diagnostics. This type of MCMC-based procedure would not be viable using high-fidelity flow simulations, as a single GEOS run requires about 13~minutes using 32~CPU cores in parallel.

\begin{figure}[!ht]
\centering
\subfloat[$\mu_{\mathrm{log}k}$, $\sigma_{\mathrm{log}k}$ (strategy~1)]{\includegraphics[width=58mm]{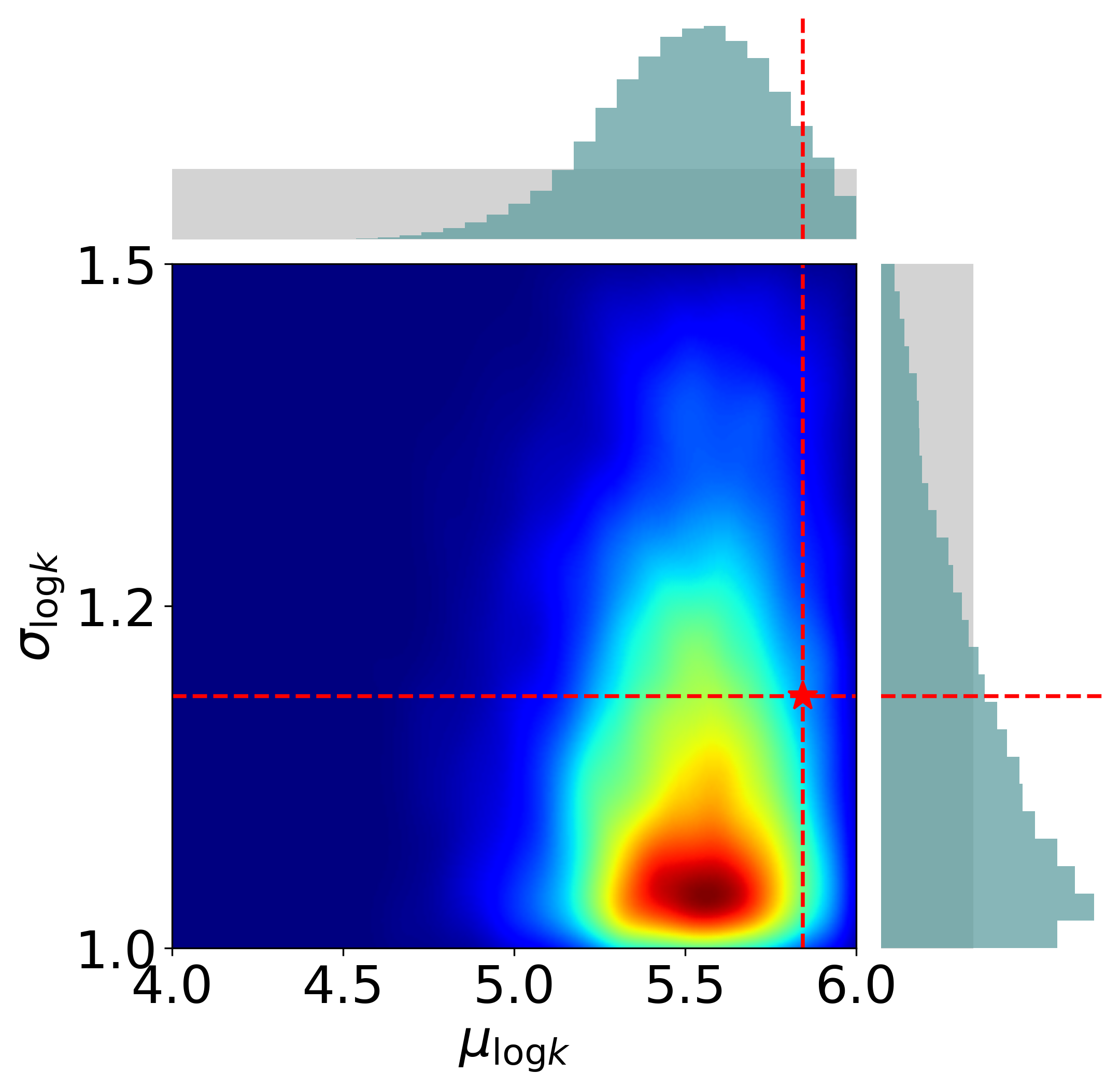}}
\hspace{2mm}
\subfloat[$\mu_{\mathrm{log}k}$, $\sigma_{\mathrm{log}k}$ (strategy~2)]{\includegraphics[width=58mm]{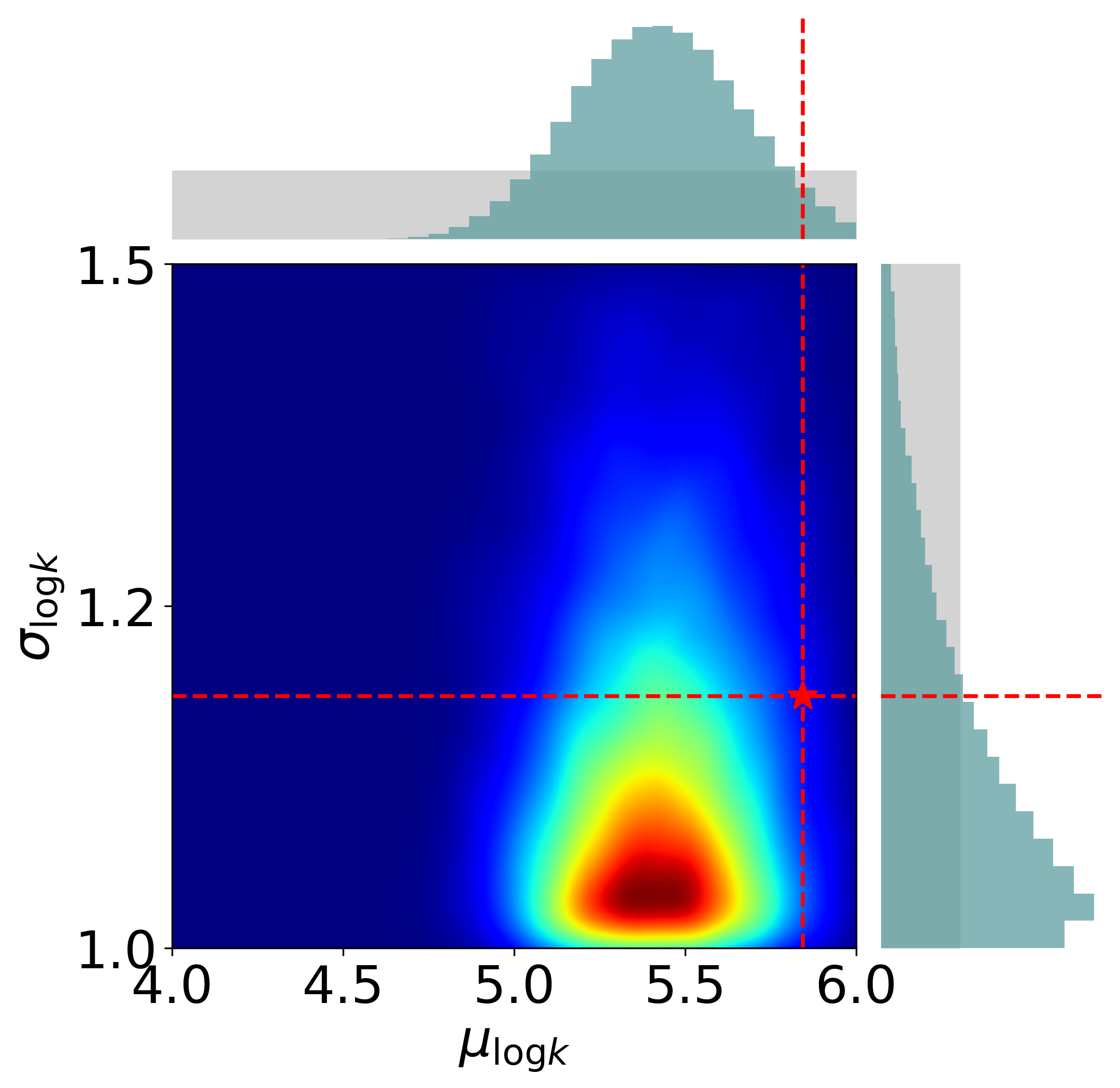}}
\hspace{2mm}
\subfloat[$\mu_{\mathrm{log}k}$, $\sigma_{\mathrm{log}k}$ (strategy~3)]{\includegraphics[width=58mm]{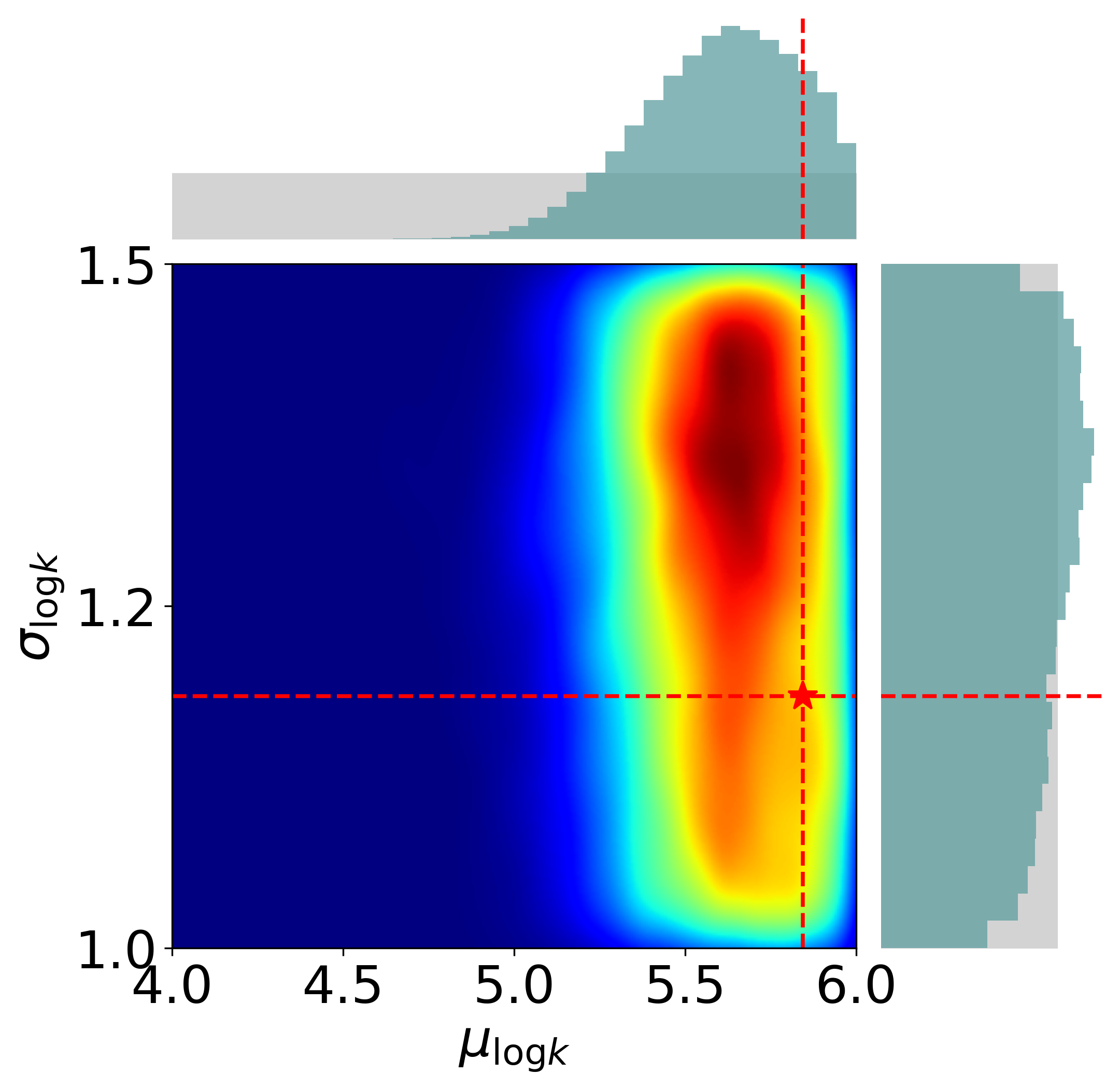}}
\\[1ex]
\subfloat[$\mu_{\phi}$, $\sigma_{\phi}$ (strategy~1)]{\includegraphics[width=58mm]{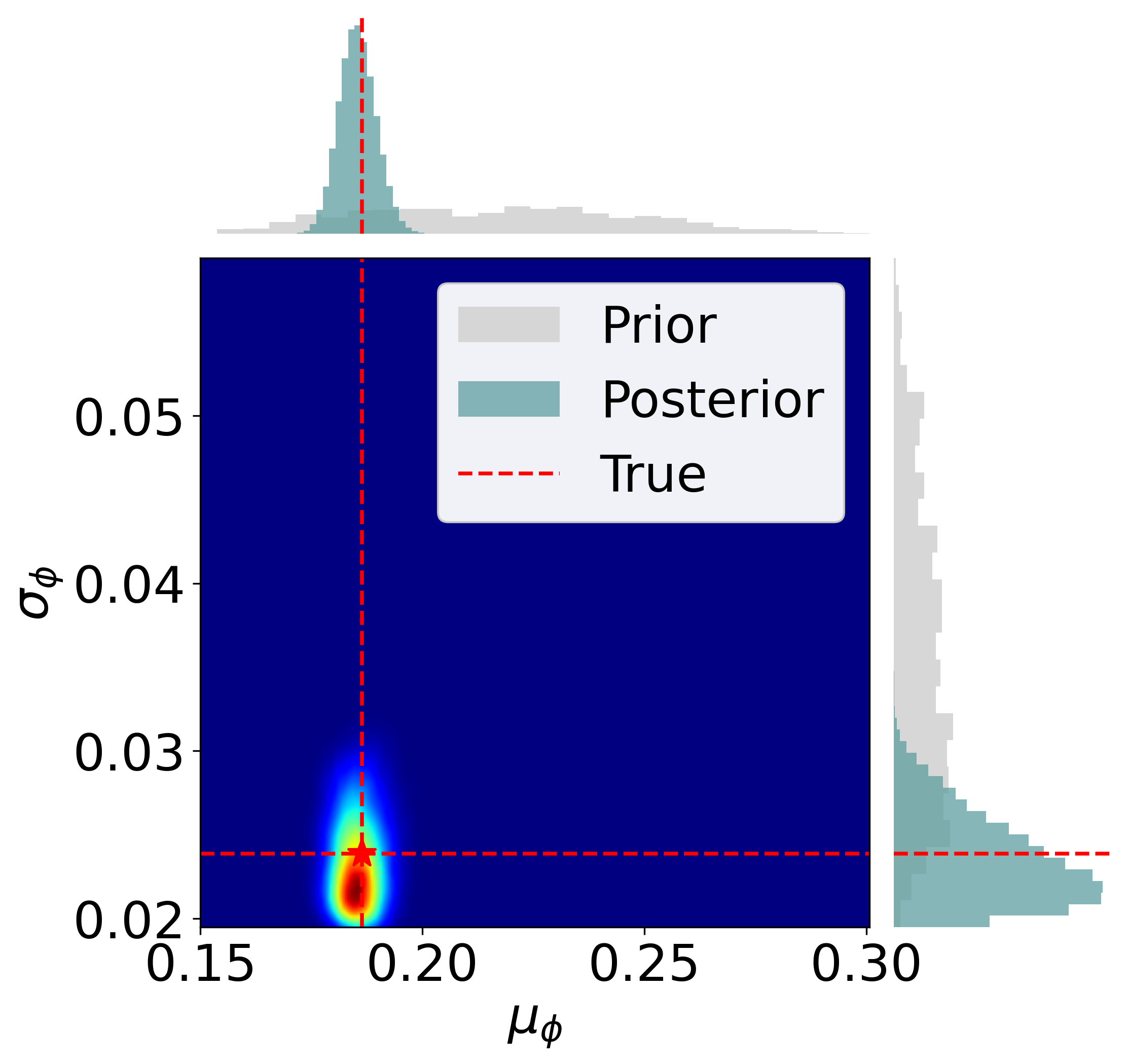}}
\hspace{2mm}
\subfloat[$\mu_{\phi}$, $\sigma_{\phi}$ (strategy~2)]{\includegraphics[width=58mm]{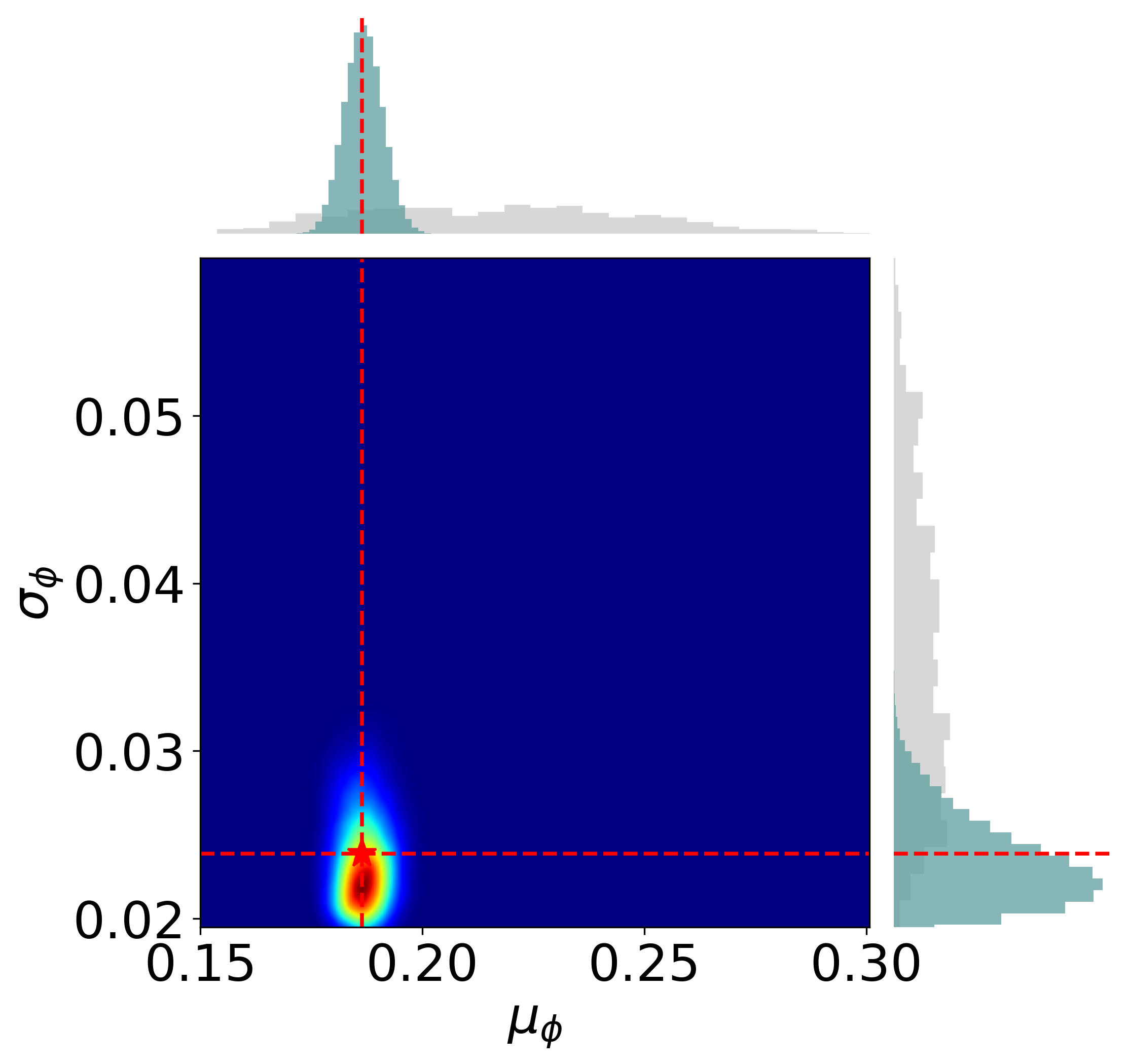}}
\hspace{2mm}
\subfloat[$\mu_{\phi}$, $\sigma_{\phi}$ (strategy~3)]{\includegraphics[width=58mm]{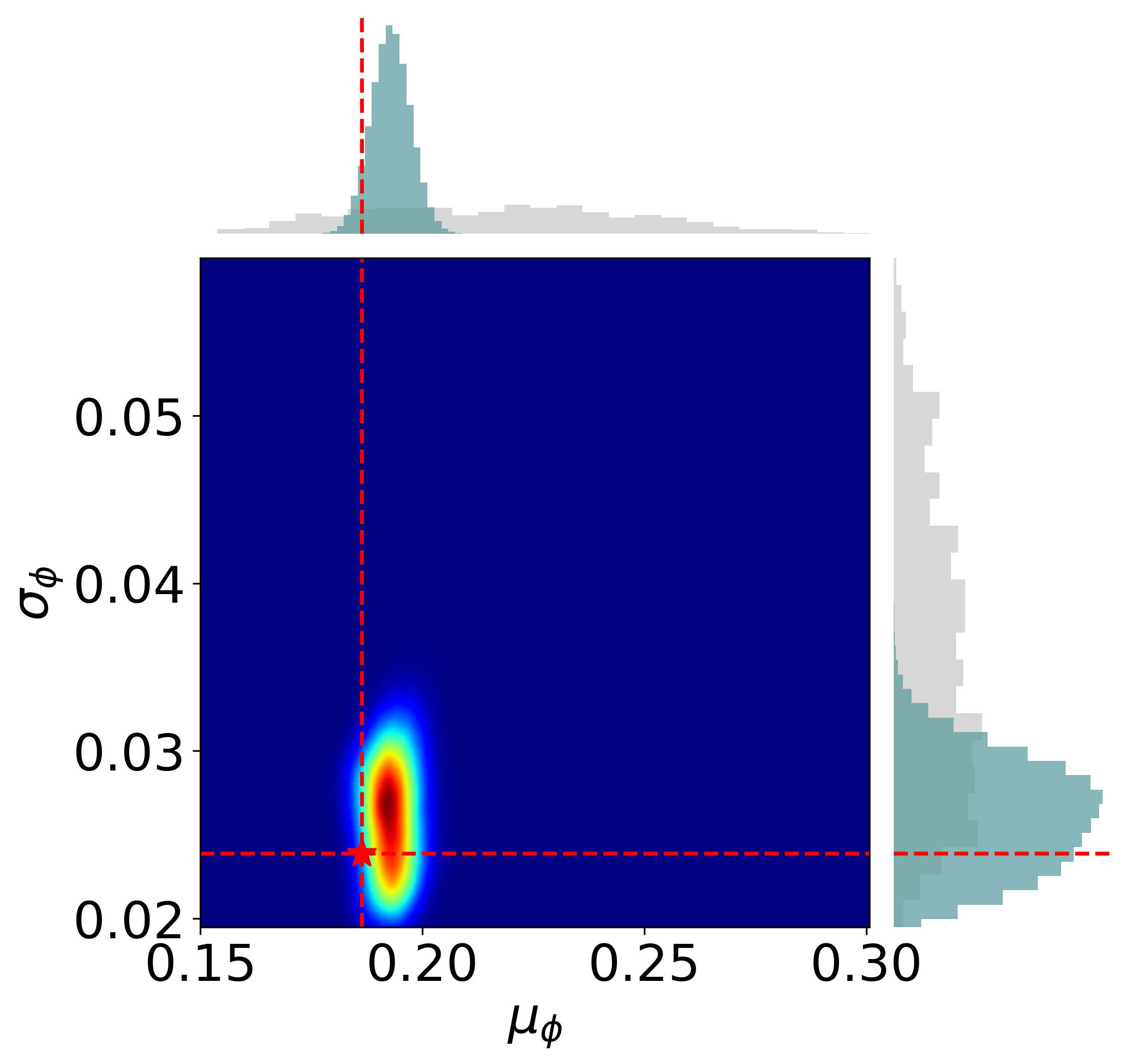}}
\caption{Data assimilation results for the joint and marginal distributions of the metaparameters $\mu_{\mathrm{log}k}$, $\sigma_{\mathrm{log}k}$, $\mu_{\phi}$, and $\sigma_{\phi}$ for the three perforation and injection strategies. Gray regions (histograms) represent prior distributions, blue histograms are posterior marginal distributions, and red lines and stars denote true values. Heat maps are the joint posterior densities for pairs of related metaparameters. Legend in (d) applies to all subplots.}
\label{meta_1_true}
\end{figure}

\begin{figure}[!ht]
\centering
\subfloat[$k_{f1}$, $k_{f2}$ (strategy~1)]{\includegraphics[width=56mm]{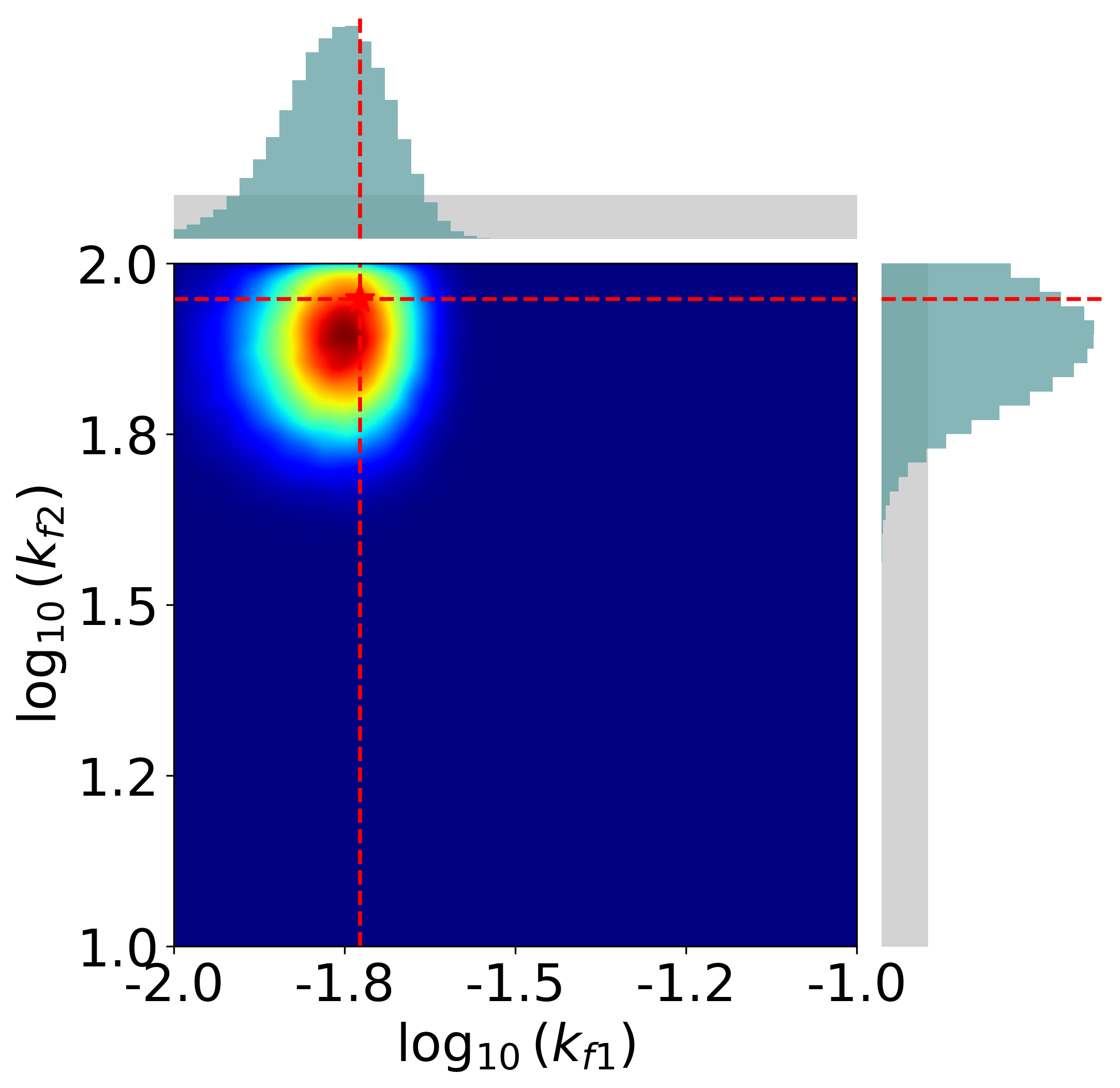}}
\hspace{2mm}
\subfloat[$k_{f1}$, $k_{f2}$ (strategy~2)]{\includegraphics[width=56mm]{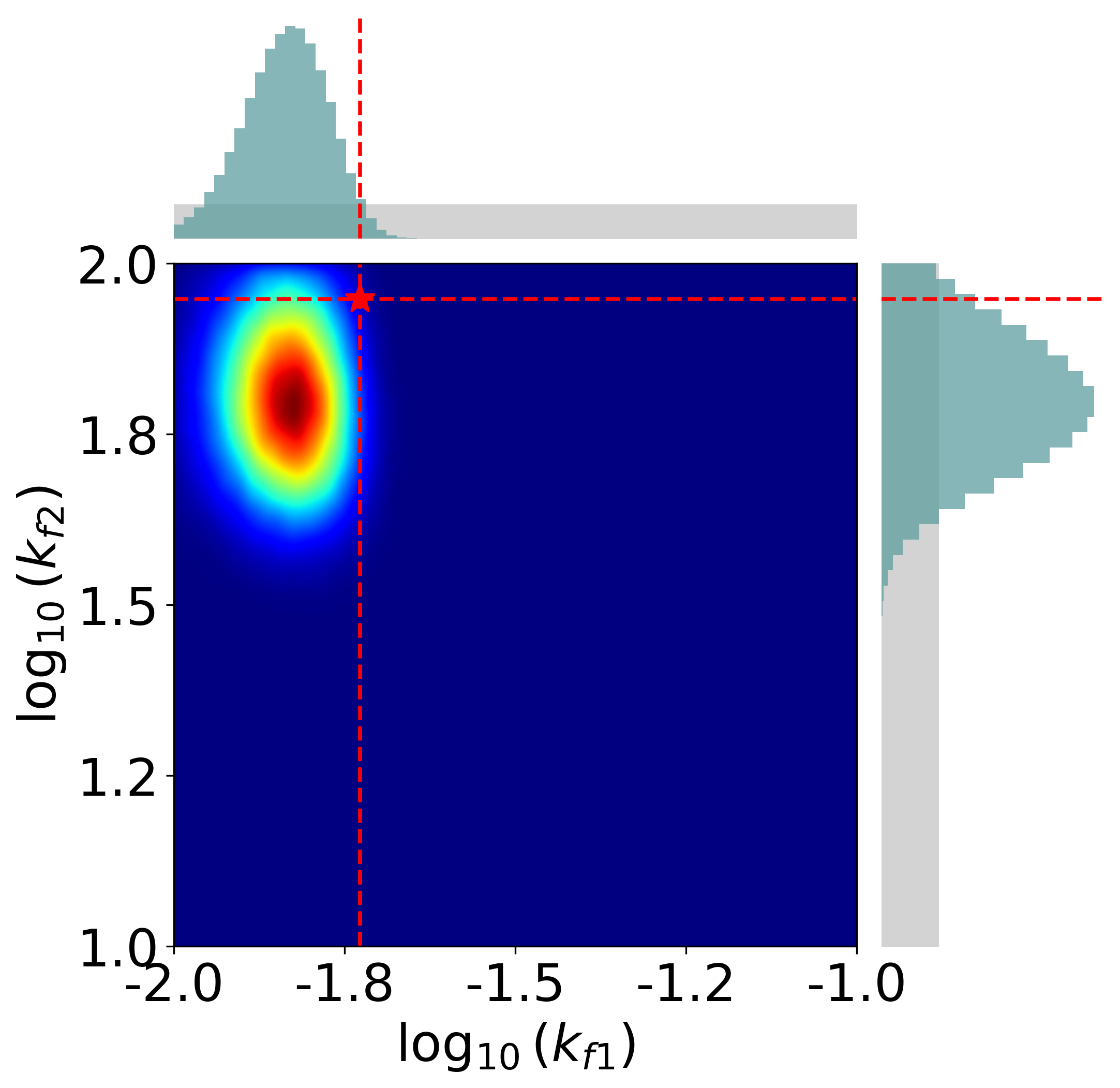}}
\hspace{2mm}
\subfloat[$k_{f1}$, $k_{f2}$ (strategy~3)]{\includegraphics[width=56mm]{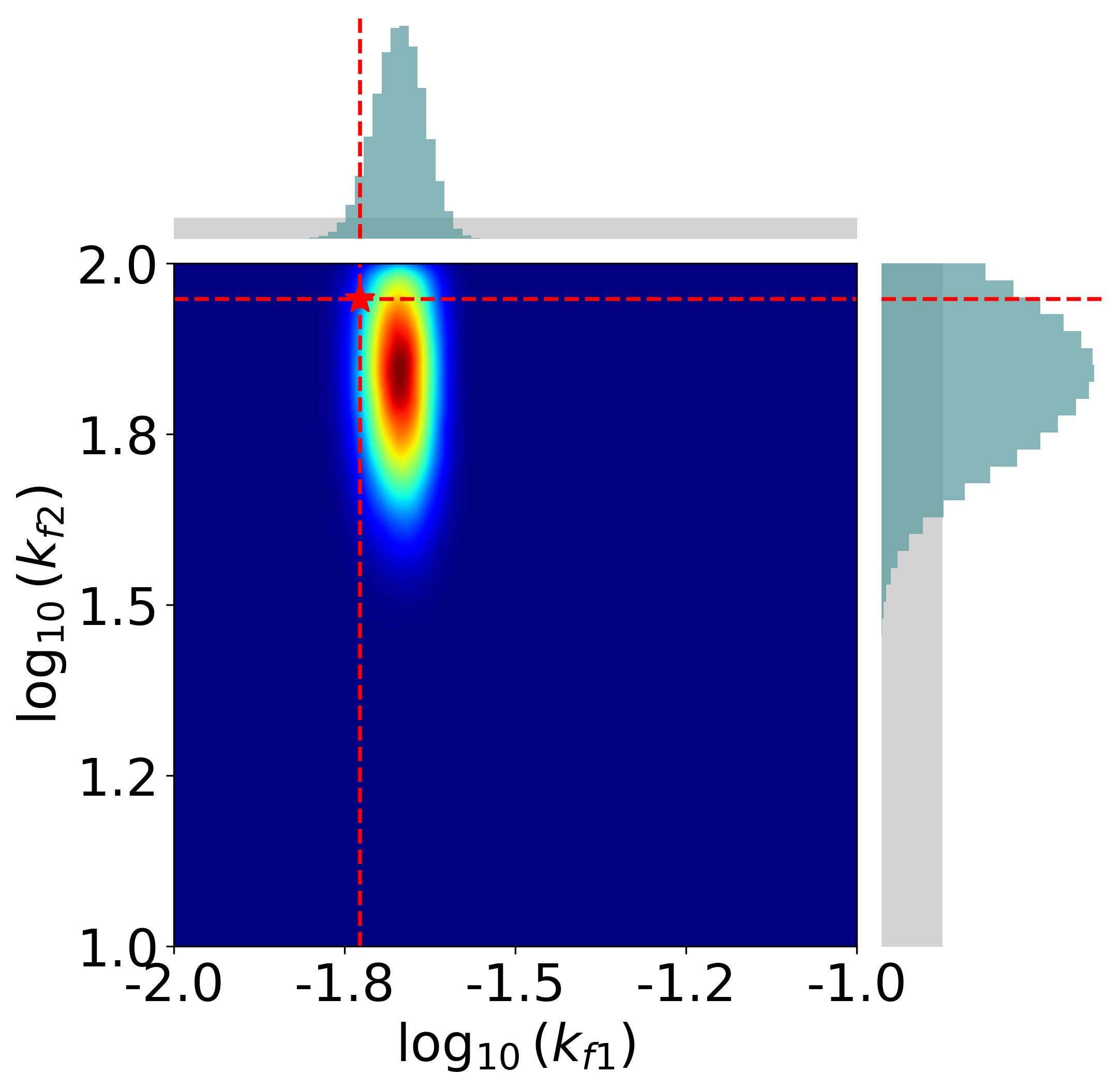}}
\\[1ex]
\subfloat[$S_{wi}$, $S_{gr}$ (strategy~1)]{\includegraphics[width=56mm]{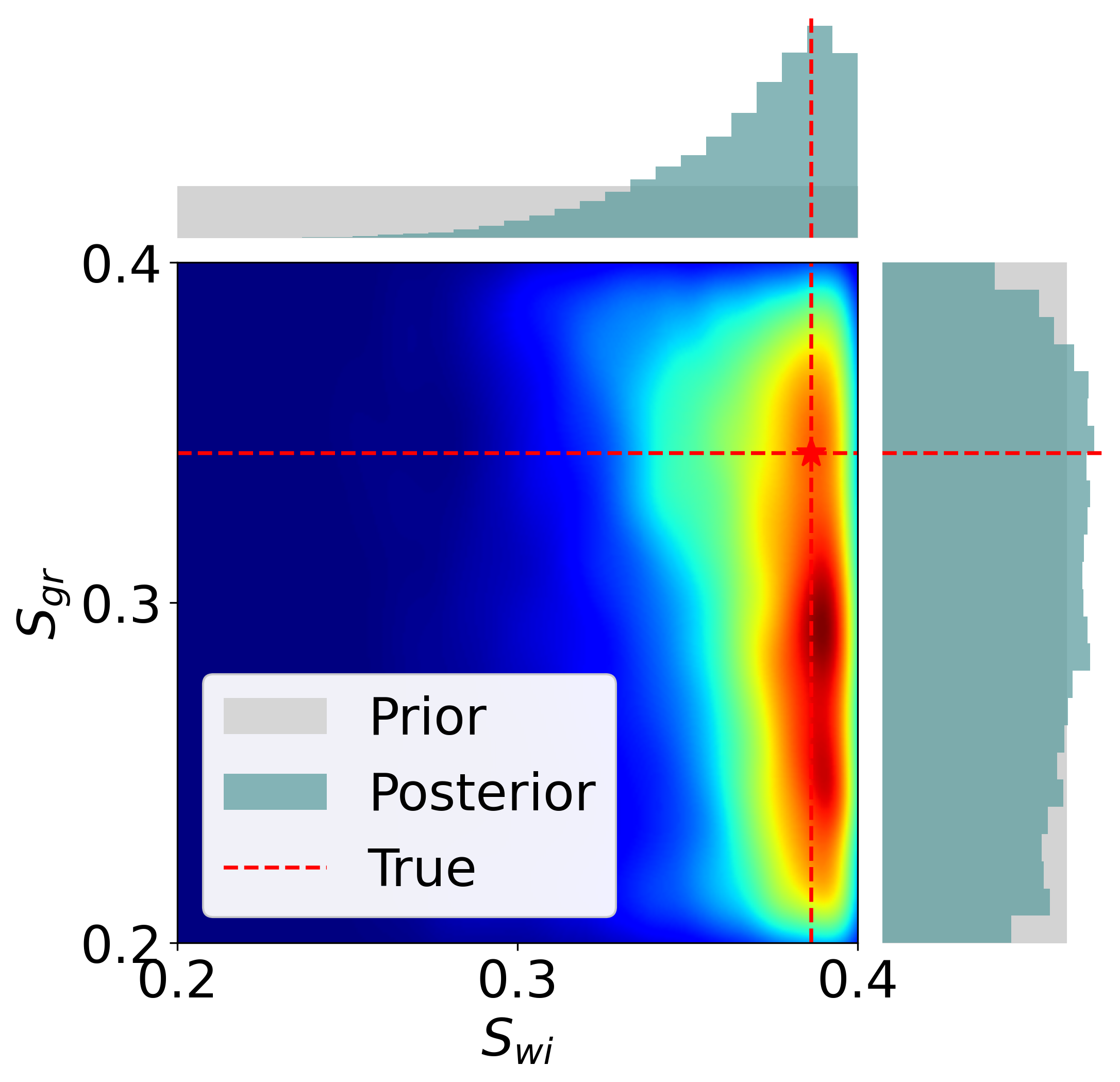}}
\hspace{2mm}
\subfloat[$S_{wi}$, $S_{gr}$ (strategy~2)]{\includegraphics[width=56mm]{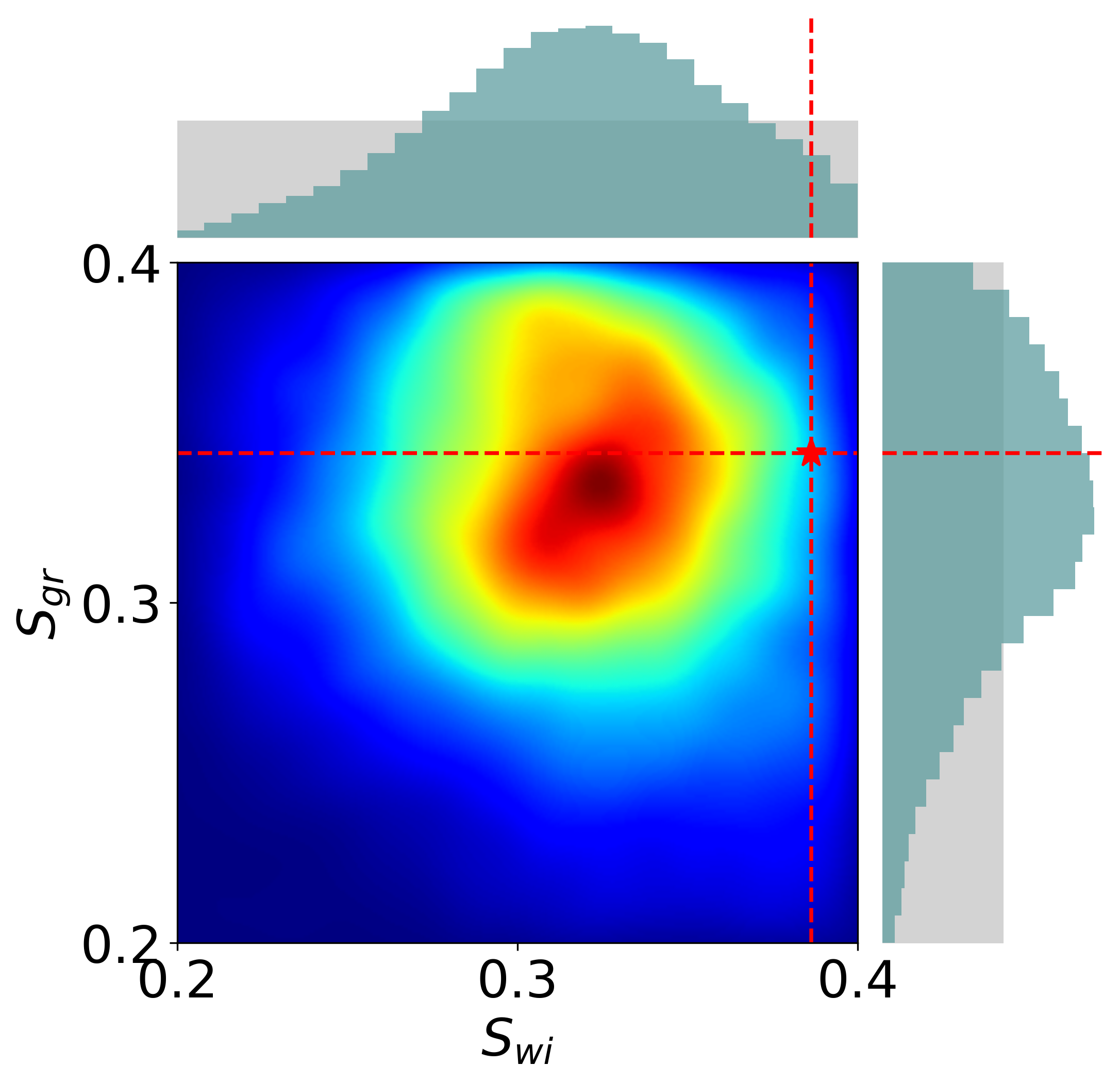}}
\hspace{2mm}
\subfloat[$S_{wi}$, $S_{gr}$ (strategy~3)]{\includegraphics[width=56mm]{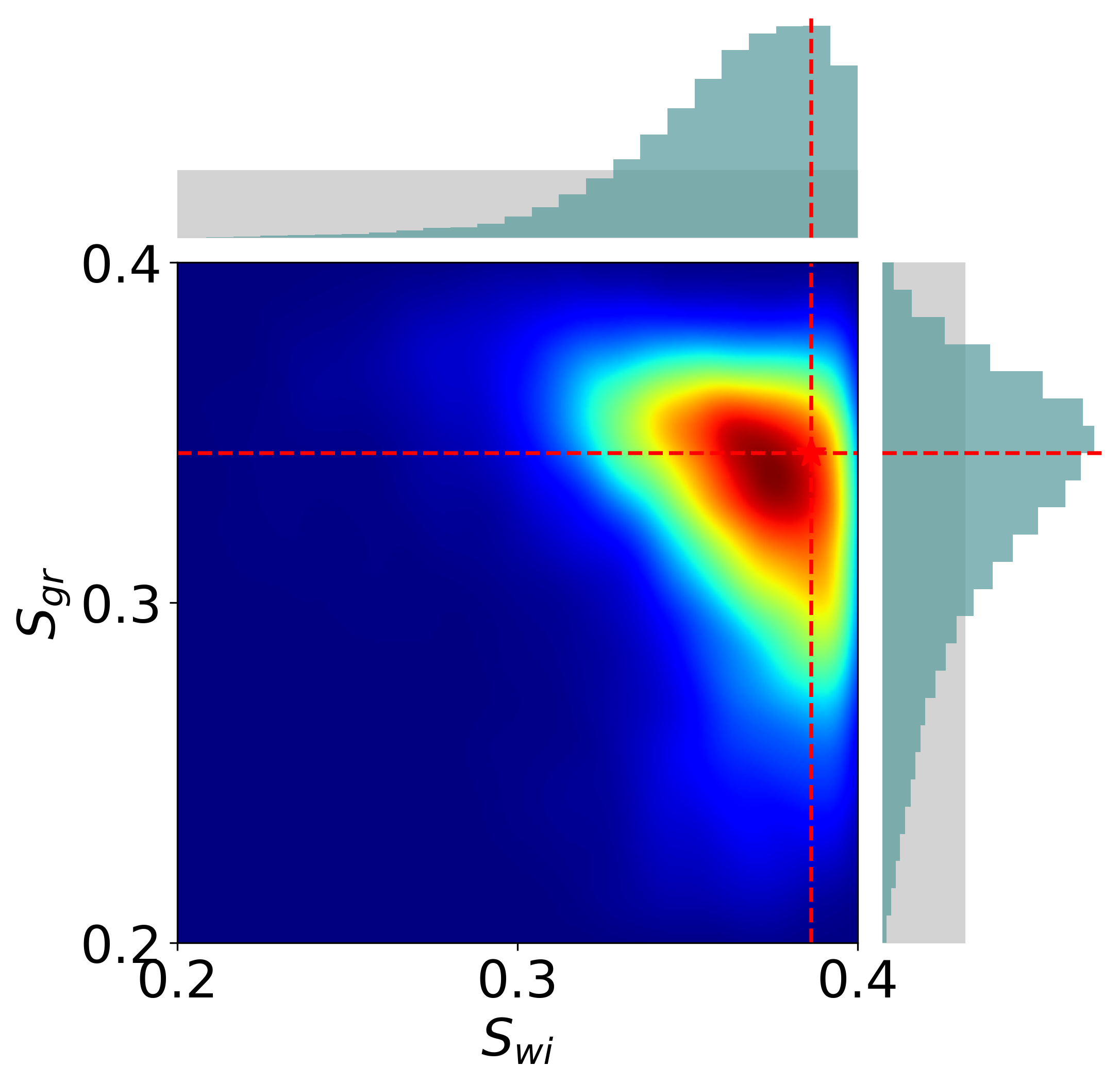}}
\\[1ex]
\subfloat[$a_r$, $k_{rg}^0$ (strategy~1)]{\includegraphics[width=56mm]{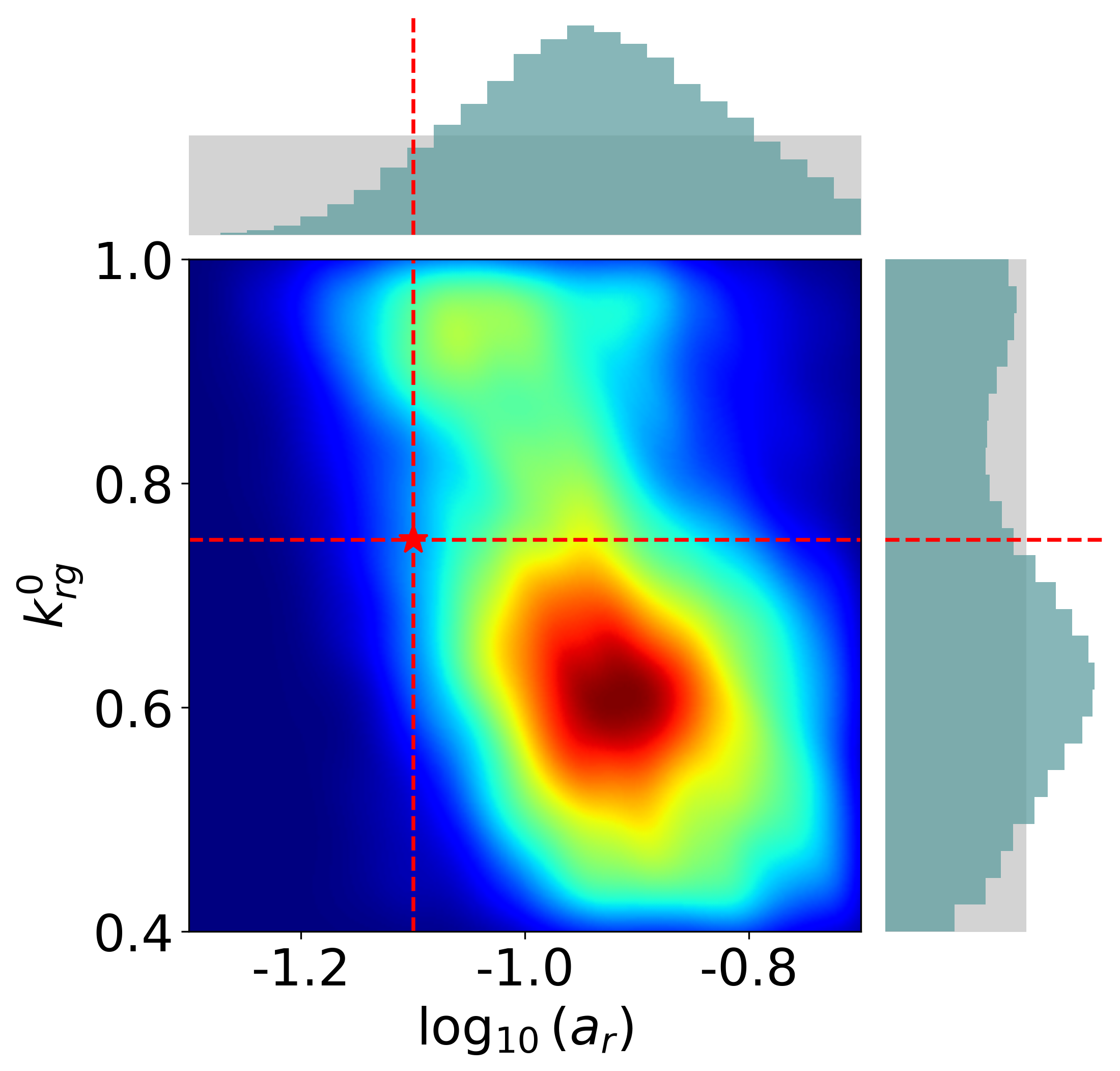}}
\hspace{2mm}
\subfloat[$a_r$, $k_{rg}^0$ (strategy~2)]{\includegraphics[width=56mm]{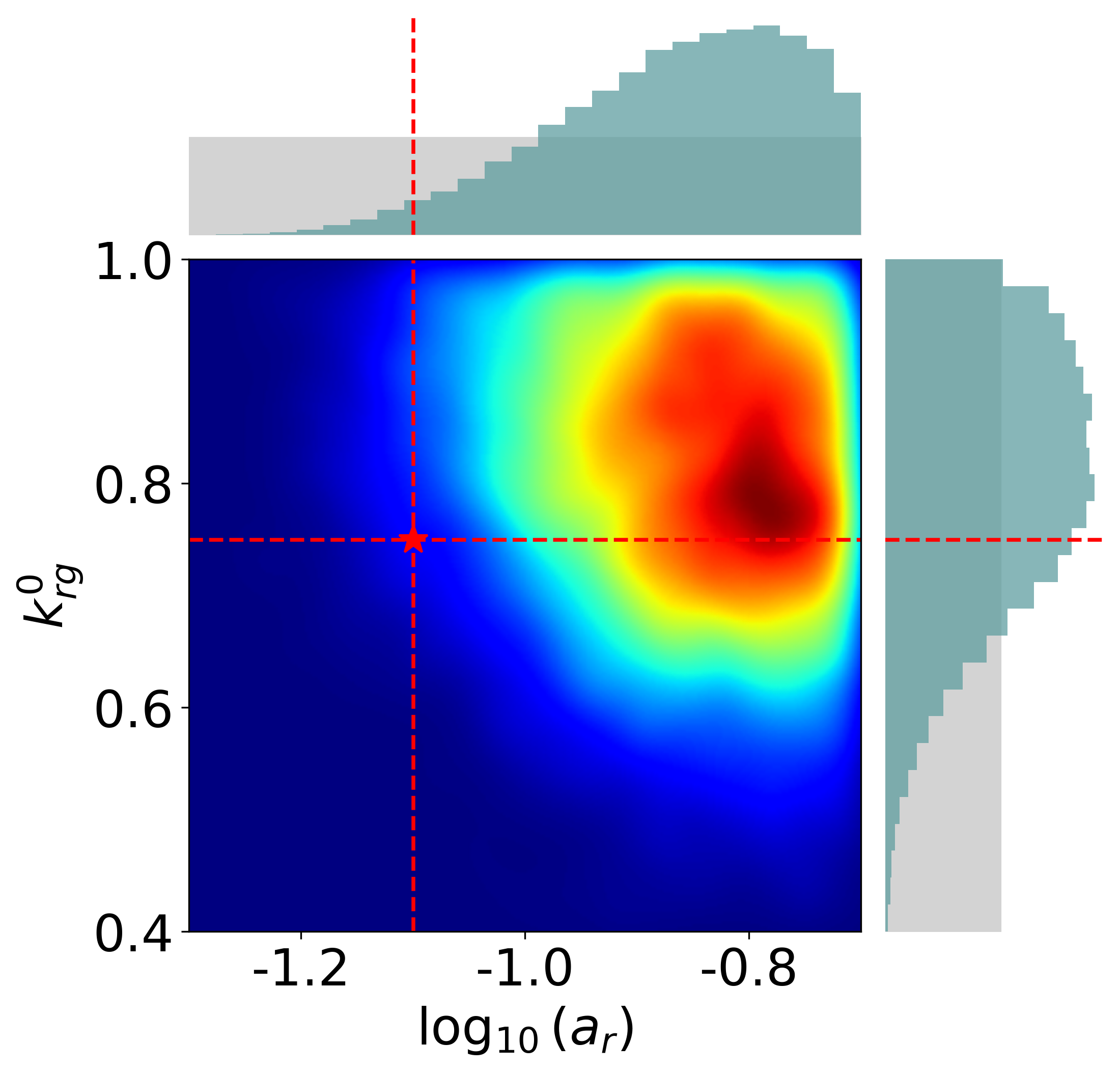}}
\hspace{2mm}
\subfloat[$a_r$, $k_{rg}^0$ (strategy~3)]{\includegraphics[width=56mm]{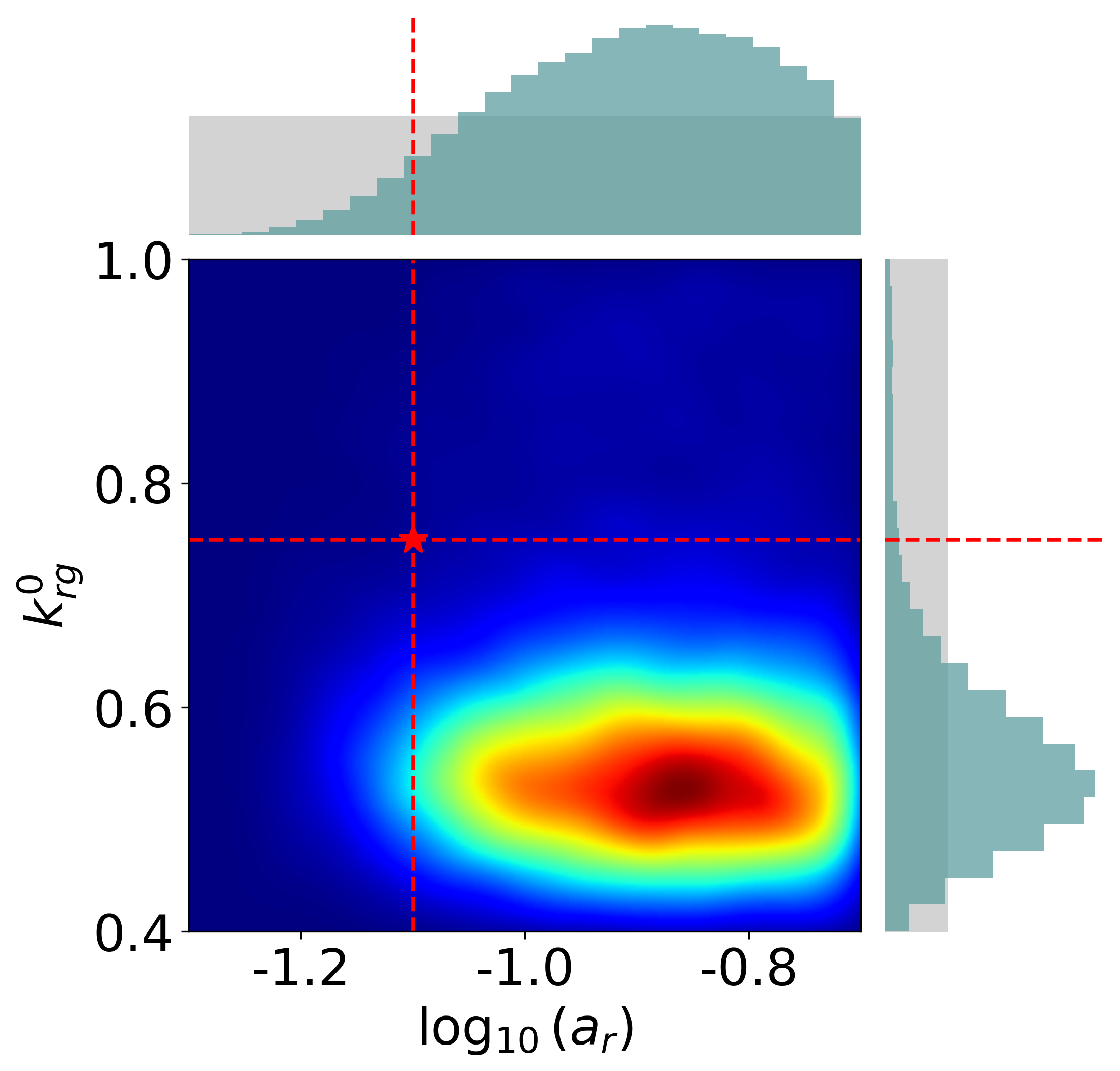}}
\caption{Data assimilation results for the joint and marginal distributions of the metaparameters $\log_{10}(k_{f1})$, $\log_{10}(k_{f2})$, $S_{wi}$, $S_{gr}$, $k_{rg}^0$, and $\log_{10}(a_r)$ for the three perforation and injection strategies. Gray regions (histograms) represent prior distributions, blue histograms are posterior marginal distributions, and red lines and stars denote true values. Heat maps are the joint posterior densities for pairs of related metaparameters. Legend in (d) applies to all subplots.}
\label{meta_2_true}
\end{figure}

Data assimilation results for the metaparameters, for the three perforation and injection strategies, are presented in Figs.~\ref{meta_1_true} and~\ref{meta_2_true}. The heat maps display the joint posterior densities for pairs of related metaparameters, with dark red indicating regions of high posterior density. The gray and blue histograms along the axes show the prior and posterior marginal distributions, and the red dashed lines and stars denote the true values. The metaparameters $d$ and $e$, which relate porosity to log-permeability, are not presented. Rather, the mean and standard deviation of porosity in the target aquifer ($\mu_{\phi}$ and $\sigma_{\phi}$), which are computed from $d$ and $e$, are displayed in Fig.~\ref{meta_1_true}. A large amount of uncertainty reduction is achieved for $\mu_{\log k}$ under all three strategies (Fig.~\ref{meta_1_true}a--c), with the true value falling well within the posterior in all cases. Similarly, substantial uncertainty reduction is observed for $\mu_{\phi}$ and $\sigma_{\phi}$ (Fig.~\ref{meta_1_true}d--f). The uncertainty reduction for $\sigma_{\log k}$ is more limited, particularly for strategy~3.

Posterior results for the fault permeabilities, permeability anisotropy ratio in the target aquifer, and relative permeability parameters are presented in Fig.~\ref{meta_2_true}. The posterior distributions clearly identify that fault~1 is essentially sealing and fault~2 is highly permeable (Fig.~\ref{meta_2_true}a--c). Note that the axis ranges in these subplots cover only a small portion of the $U(-2,2)$ prior ranges for $\log_{10}(k_{f1})$ and $\log_{10}(k_{f2})$. These quantities are well constrained because the pressure and saturation observations are sensitive to the degree of fault leakage.

Differences between the three strategies are observed for $S_{wi}$ and $S_{gr}$ (Fig.~\ref{meta_2_true}d--f). Strategy~3 provides significant uncertainty reduction for both parameters, with the true values located near the peak of the joint posterior distribution. With strategy~1, the posterior distribution for $S_{wi}$ peaks near its true value, though almost no uncertainty reduction is achieved for $S_{gr}$. Strategy~2 provides the least uncertainty reduction for $S_{wi}$ and an intermediate amount for $S_{gr}$. For $\log_{10}(a_r)$ and $k_{rg}^0$ (Fig.~\ref{meta_2_true}g--i), some uncertainty reduction is achieved, though the true values are near the edges of the distributions in some cases.

The anomalous behavior observed in Fig.~\ref{meta_2_true}i (and to some extent in the results for other metaparameters), where the true values of the metaparameters are outside the high-density portion of the posterior distribution, may be due to our treatment of model error. As described earlier, $C_{\mathrm{surr}}$ is computed from surrogate model errors over the full test set. This covariance is then fixed throughout the MCMC run. The actual model error, however, is likely to depend on the metaparameters and operational strategy. This dependence could be treated by updating $C_{\mathrm{surr}}$ during the MCMC procedure (a related strategy was suggested by \citet{oliver2018calibration} in the context of ensemble-based data assimilation). This could be achieved by performing high-fidelity simulations for a set of accepted post burn-in samples and then constructing an updated $C_{\mathrm{surr}}$ from the corresponding pairs of simulation and surrogate results. Alternatively, a subset of the previously simulated training or test cases that are ``close'' to the posterior samples could be identified and then used for the construction of $C_{\mathrm{surr}}$. Approaches along these lines should be investigated in future work.

\begin{figure}[!ht]
\centering
\subfloat[I1 saturation footprint (strategy~1)]{\includegraphics[width=85mm]{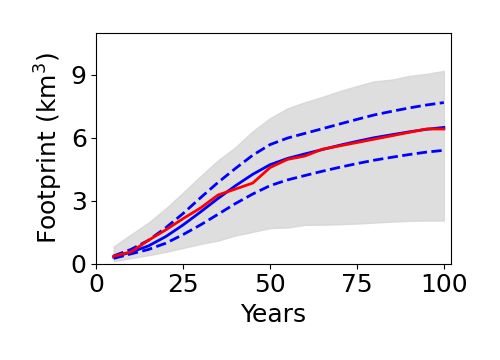}}
\hspace{2mm}
\subfloat[I2 saturation footprint (strategy~1)]{\includegraphics[width=85mm]{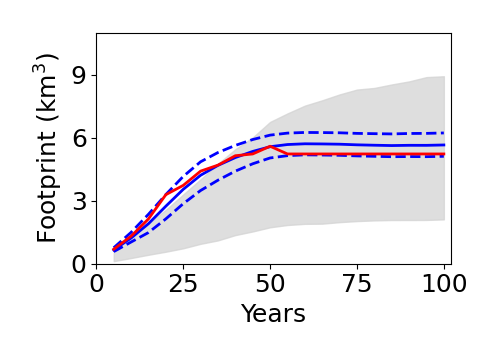}}
\\[1ex]
\subfloat[Total mass of CO$_2$ (strategy~1)]{\includegraphics[width=85mm]{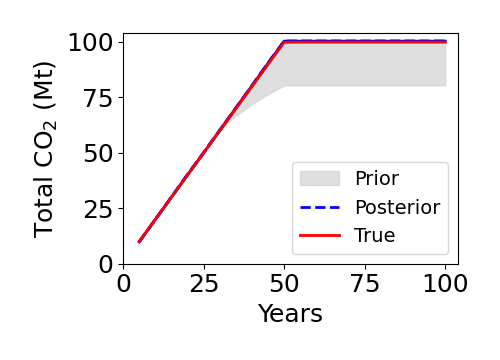}}
\hspace{2mm}
\subfloat[Mobile mass of CO$_2$ (strategy~1)]{\includegraphics[width=85mm]{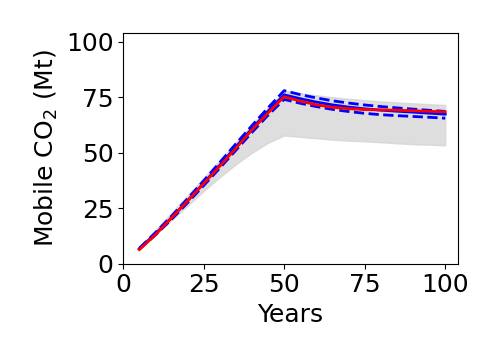}}
\caption{Data assimilation results for the saturation footprint associated with each injector, total injected CO$_2$, and mobile CO$_2$ mass for strategy~1. Gray regions represent the prior P$_{5}$--P$_{95}$ range, red curves denote true values, and blue curves show the posterior P$_{5}$, P$_{50}$ and P$_{95}$ predictions. All four curves overlap in (c). Legend in (c) applies to all subplots.}
\label{posterior_results_scenario_1}
\end{figure}

\begin{figure}[!ht]
\centering
\subfloat[I1 saturation footprint (strategy~2)]{\includegraphics[width=85mm]{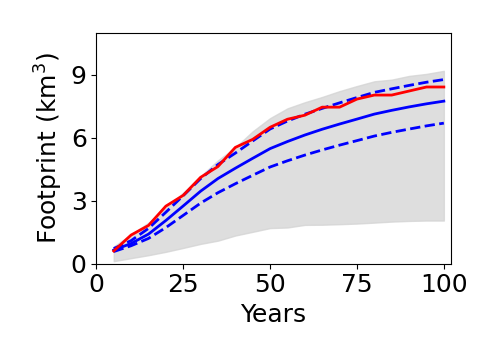}}
\hspace{2mm}
\subfloat[I2 saturation footprint (strategy~2)]{\includegraphics[width=85mm]{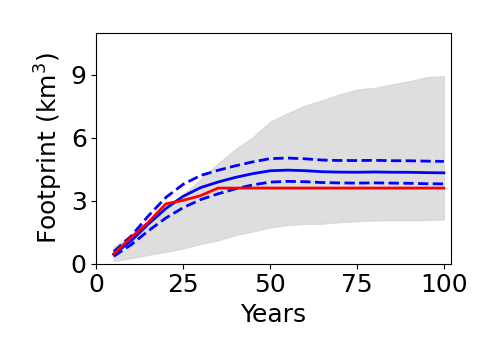}}
\\[1ex]
\subfloat[Total mass of CO$_2$ (strategy~2)]{\includegraphics[width=85mm]{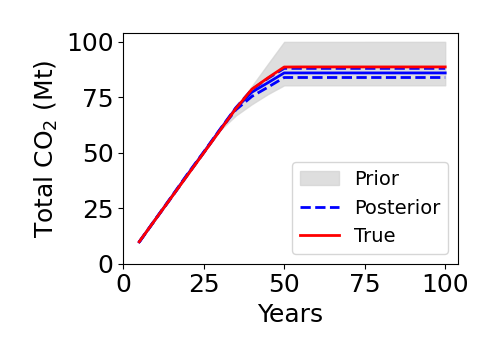}}
\hspace{2mm}
\subfloat[Mobile mass of CO$_2$ (strategy~2)]{\includegraphics[width=85mm]{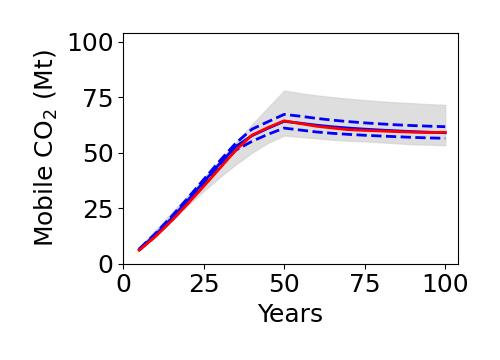}}
\caption{Data assimilation results for the saturation footprint associated with each injector, total injected CO$_2$, and mobile CO$_2$ mass for strategy~2. Gray regions represent the prior P$_{5}$--P$_{95}$ range, red curves denote true values, and blue curves show the posterior P$_{5}$, P$_{50}$ and P$_{95}$ predictions. Legend in (c) applies to all subplots.}
\label{posterior_results_scenario_2}
\end{figure}

\begin{figure}[!ht]
\centering
\subfloat[I1 saturation footprint (strategy~3)]{\includegraphics[width=85mm]{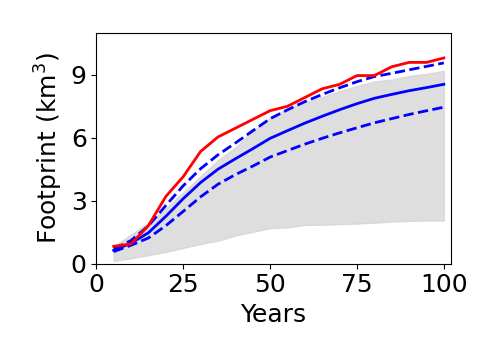}}
\hspace{2mm}
\subfloat[I2 saturation footprint (strategy~3)]{\includegraphics[width=85mm]{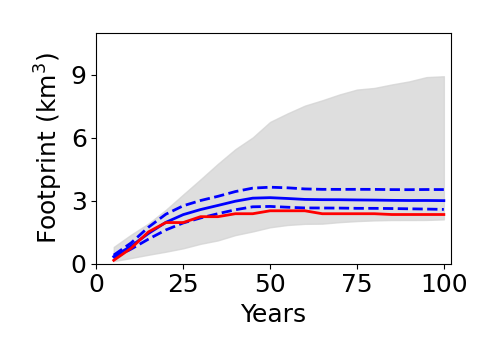}}
\\[1ex]
\subfloat[Total mass of CO$_2$ (strategy~3)]{\includegraphics[width=85mm]{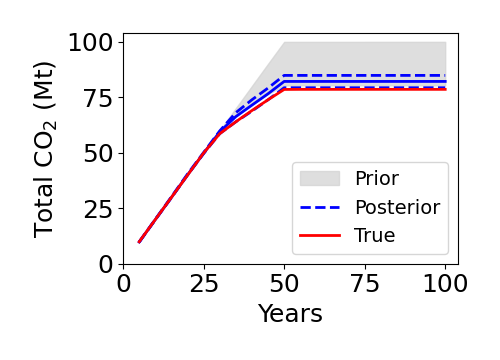}}
\hspace{2mm}
\subfloat[Mobile mass of CO$_2$ (strategy~3)]{\includegraphics[width=85mm]{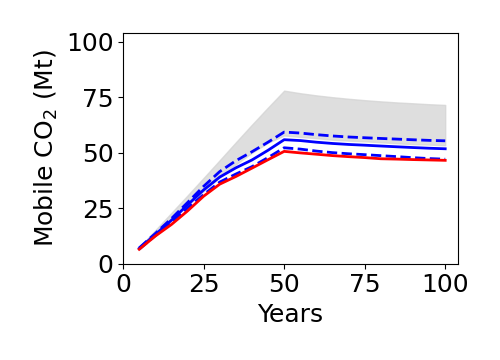}}
\caption{Data assimilation results for the saturation footprint associated with each injector, total injected CO$_2$, and mobile CO$_2$ mass for strategy~3. Gray regions represent the prior P$_{5}$--P$_{95}$ range, red curves denote true values, and blue curves show the posterior P$_{5}$, P$_{50}$ and P$_{95}$ predictions. Legend in (c) applies to all subplots.}
\label{posterior_results_scenario_3}
\end{figure}

We now present posterior predictions for key quantities of interest. The saturation footprint associated with each injector and the total injected and mobile CO$_2$ mass in the overall domain, for each of the three perforation and injection strategies, are shown in Figs.~\ref{posterior_results_scenario_1}, \ref{posterior_results_scenario_2}, and~\ref{posterior_results_scenario_3}. The gray regions show prior P$_5$--P$_{95}$ results computed from the full surrogate model test set, which correspond to randomly sampled geomodel realizations and perforation and injection strategies. The red curves show the true model responses, while the blue curves display the posterior P$_5$, P$_{50}$, and P$_{95}$ predictions. We display results for the P$_5$--P$_{95}$ range rather than the P$_{10}$--P$_{90}$ range because many of the true responses are near the edges of the prior distribution.

Substantial uncertainty reduction is achieved for all quantities under all three strategies. For strategy~1 (Fig.~\ref{posterior_results_scenario_1}), the posterior interval for total CO$_2$ mass nearly collapses onto the true response. Although many of the prior models switch to BHP control and therefore inject less than the full 100~Mt, the posterior results correctly show that the true model achieves the target injection. The posterior P$_{50}$ predictions for the saturation footprints and mobile CO$_2$ mass closely match the true responses, which remain within the posterior P$_5$--P$_{95}$ intervals throughout the 100-year period.

Results for strategies~2 and~3 (Figs.~\ref{posterior_results_scenario_2} and~\ref{posterior_results_scenario_3}) show analogous overall behavior. The posterior intervals are much narrower than the prior ranges, and the results are generally within or very near the P$_5$--P$_{95}$ intervals. There are some discrepancies, however, for strategy~3, e.g., Fig.~\ref{posterior_results_scenario_3}a, where the true result is somewhat above the P$_{95}$ posterior result. This may again be related to our treatment of model error. Specifically, the data assimilation procedure accounts for surrogate model errors in pressure and saturation through $C_{\mathrm{surr}}$, but surrogate error is not included in the predictions for the quantities of interest.

It is evident from the results in Figs.~\ref{meta_1_true} to~\ref{posterior_results_scenario_3} that different perforation and injection strategies lead to different degrees of uncertainty reduction in the metaparameters and quantities of interest. The observed pressure and saturation data are collected at just a few locations, and these pressures and saturations may be sensitive to global characteristics (metaparameters) and to local properties like the detailed permeability field. The amount of uncertainty reduction is thus a complicated function of the true model, the operational strategy, and the observation locations. We note finally that the locations of the observation wells could be optimized, either for a given operational strategy or over all possible strategies, with the goal of maximizing uncertainty reduction in particular metaparameters or quantities of interest. The approach used by \citet{xiaowen_he_2026} could be adapted for this purpose.

\clearpage

\section{Concluding Remarks} 
\label{Conclusions}
In this study, we developed a multimodal auto-regressive transformer surrogate model to treat variable perforation and injection operations, under geological uncertainty, in geological carbon storage problems. A modified version of the SEAM CO$_2$ geomodel -- a system with three stacked aquifers intersected by two large faults -- was considered. Uncertainty in this system is hierarchical and characterized by 10 metaparameters and PCA latent variables. The metaparameters define permeability and porosity mean and standard deviation, permeability anisotropy, fault permeabilities, and relative permeability functions, while the PCA variables inform detailed cell-by-cell properties. Supercritical CO$_2$ is injected through two vertical wells, with a target of 100~Mt over a 50-year injection period, subject to a maximum bottom-hole pressure constraint. In the operational strategies treated here, the wells are perforated in three stages, proceeding from the bottom of the target aquifer to the top, with the stage durations and well injection rates as the control variables. Using high-fidelity GEOS simulations, we demonstrated that the variable perforation and injection strategy can achieve the full 100~Mt injection target while reducing the amount of mobile CO$_2$ relative to a base case involving fully perforated wells and constant injection rates.

The surrogate model introduced in this study processes the 3D geomodel, the relative permeability parameters, and the control variables through separate encoders. The resulting embedded tokens are fused and integrated through multihead self-attention in a transformer encoder. A temporal decoder generates predictions auto-regressively, with encoder-decoder cross-attention correlating the decoder to the encoded multimodal representation. Networks of this type were trained, using results from 4000 GEOS flow simulations, to predict saturation and pressure at the monitoring locations, the total injected and mobile CO$_2$ mass in the overall domain, and the CO$_2$ saturation footprint associated with each injector, all at a series of time steps.

Surrogate model performance was evaluated on a test set of 500 new geomodel realizations, each involving a randomly sampled geomodel, relative permeability parameters, and control variables (which define the perforation and injection strategy). Median errors of 0.028 (MAE) for saturation, 0.21\% for pressure, 2.3\% for the total injected CO$_2$, 3.3\% for the mobile CO$_2$ mass, and 3.9\% and 5.1\% for the saturation footprints of the two injectors were achieved. The surrogate model was shown to reproduce the P$_{10}$, P$_{50}$, and P$_{90}$ ensemble statistics from GEOS for all quantities considered. Importantly, the model captures the switch from rate control to BHP control, which strongly impacts the total amount of CO$_2$ injected, along with the effect of different operational strategies for a specific geomodel realization.

The surrogate model was then applied within a hierarchical MCMC-based data assimilation procedure. A synthetic true model, with observed data comprising noisy saturation and pressure at two observation wells, was considered under three different perforation and injection strategies. The three surrogate-based MCMC runs required around 1--3~million function evaluations to achieve convergence, though the elapsed times for these runs were only about 4.3--10.4~hours on a single GPU. A procedure of this type is infeasible using high-fidelity simulation, since a single GEOS run requires about 13~minutes on 32 CPU cores. Substantial uncertainty reduction was achieved for the fault permeabilities and porosity statistics under all three strategies, with the posterior distribution peaks close to the true values. The degree of uncertainty reduction for other metaparameters was found to vary with the operational strategy. Posterior predictions for the saturation footprint, total injected mass, and mobile CO$ _2$ mass were much narrower than the corresponding prior ranges, and were generally consistent with the true model responses.

There are a number of useful directions for future research in this area. It will be of interest to apply the multimodal auto-regressive transformer surrogate model for optimization of the perforation and injection strategy under geological uncertainty, or within a closed-loop framework that combines data assimilation and optimization, repeatedly, in sequence. As noted in Section~\ref{sec:da_results}, the treatment of surrogate model error, currently based on a fixed error covariance computed from the test set, could be improved by updating this covariance during the MCMC procedure. The observation well locations could be optimized to maximize uncertainty reduction for particular metaparameters or quantities of interest. Finally, the application of the surrogate model to other geological settings (e.g., channelized systems), and to real field operations with actual observed data, should also be pursued.

\section*{CRediT authorship contribution statement}
\textbf{Yifu Han}: Conceptualization, Methodology, Software, Visualization, Formal analysis, Writing -- original draft. \textbf{Louis J. Durlofsky}: Supervision, Conceptualization, Resources, Formal analysis, Writing -- review \& editing.

\section*{Declaration of competing interest}
The authors declare that they have no known competing financial interests or personal relationships that could have appeared to influence the work reported in this paper.

\section*{Data availability}
The code used in this study will be made available on github when this paper is published. Please contact Yifu Han (yifu@stanford.edu) for earlier access.

\section*{Acknowledgments} 
We thank the Stanford Center for Carbon Storage and Stanford Smart Fields Consortium for funding. We are grateful to the SDSS Center for Computation for HPC resources, and to developers at Lawrence Livermore National Laboratory, Stanford University, and TotalEnergies for assistance with GEOS and the SEAM CO$_2$ geomodel.

\bibliographystyle{Main} 
\bibliography{ref}

\end{document}